\documentclass[dvipsnames]{article}

\usepackage[nonatbib, preprint]{neurips_2025}

\usepackage[utf8]{inputenc} 
\usepackage[T1]{fontenc}    
\usepackage{url}            
\usepackage{booktabs}       
\usepackage{amsfonts}       
\usepackage{nicefrac}       
\usepackage{microtype}      
\usepackage[table]{xcolor}
\usepackage[numbers]{natbib}

\usepackage{tabularx}
\usepackage{tabularray}
\usepackage{listings}
\usepackage[inline]{enumitem}
\usepackage{caption}
\usepackage{subcaption}
\usepackage{graphicx}
\usepackage{glossaries}
\usepackage{adjustbox}
\usepackage{pgfplots}
\usepackage{mleftright}
\usepackage{derivative}
\usepackage{pifont}
\usepackage{titlesec}
\usepackage{tablefootnote}
\usepackage{algorithm,algorithmicx,algpseudocode}
\usepackage{soul}
\usepackage[frozencache]{minted}
\usepackage{wrapfig}
\usepackage{bm}
\usepackage[bottom]{footmisc}
\usepackage{afterpage}
\usepackage{multirow}
\usepackage{makecell}
\usepackage{siunitx}
\usepackage{hyperref}       
\UseTblrLibrary{booktabs}
\UseTblrLibrary{siunitx}

\definecolor{myBlue}{HTML}{0F1A5F}
\definecolor{myRed}{HTML}{721010}
\hypersetup{
  colorlinks,
  linkcolor={myRed},
  citecolor={myBlue},
  urlcolor={blue!80!black}
  }
\glsdisablehyper
\hypersetup{breaklinks=true}
\usepgfplotslibrary{groupplots}
\usepgfplotslibrary{colormaps}
\usepgfplotslibrary{fillbetween}
\usetikzlibrary{plotmarks}
\usetikzlibrary{patterns}

\pgfplotsset{compat=1.14}	 %
\pgfplotsset{compat/show suggested version=false}
\pgfplotsset{every mark/.append style={solid}}
\mleftright

\usepackage{amsmath}
\usepackage{amssymb}
\usepackage{mathtools}
\usepackage{amsthm}

\usepackage[capitalize,noabbrev]{cleveref}

\theoremstyle{plain}

\theoremstyle{definition}

\theoremstyle{remark}

\definecolor{C0}{rgb}{0.121569, 0.466667, 0.705882}
\definecolor{C1}{rgb}{1.000000, 0.498039, 0.054902}
\definecolor{C2}{rgb}{0.172549, 0.627451, 0.172549}
\definecolor{C3}{rgb}{0.839216, 0.152941, 0.156863}
\definecolor{C4}{rgb}{0.580392, 0.403922, 0.741176}
\definecolor{C5}{rgb}{0.549020, 0.337255, 0.294118}
\definecolor{C6}{rgb}{0.890196, 0.466667, 0.760784}
\definecolor{C7}{rgb}{0.498039, 0.498039, 0.498039}
\definecolor{C8}{rgb}{0.737255, 0.741176, 0.133333}
\definecolor{C9}{rgb}{0.090196, 0.745098, 0.811765}

\newcolumntype{Y}{>{\centering\arraybackslash}X}
\newcolumntype{C}{>{\hsize=.0\hsize\centering\arraybackslash}X}

\colorlet{LightGoldenrod}{White!40!Goldenrod}
\colorlet{LightGray}{White!90!Periwinkle}
\definecolor{LG}{gray}{0.95}

\sethlcolor{LightGoldenrod}

\newcommand{\mycc}{\cellcolor{LightGray}}

\definecolor{codegreen}{rgb}{0,0.6,0}
  \definecolor{codegray}{rgb}{0.5,0.5,0.5}
  \definecolor{codepurple}{rgb}{0.58,0,0.82}
  \definecolor{backcolour}{rgb}{0.95,0.95,0.92}
  \lstdefinestyle{mystyle}{
    backgroundcolor=\color{backcolour},
    commentstyle=\color{codegreen},
    keywordstyle=\color{magenta},
    numberstyle=\tiny\color{codegray},
    stringstyle=\color{codepurple},
    basicstyle=\ttfamily\footnotesize,
    breakatwhitespace=false,
    breaklines=true,
    captionpos=b,
    keepspaces=true,
    numbers=left,
    numbersep=5pt,
    showspaces=false,
    showstringspaces=false,
    showtabs=false,
    tabsize=2
  }
  
  \lstnewenvironment{mylisting}{\lstset{style=mystyle}}{}

\newcommand{\wcfg}{w}
\newcommand{\wcfglow}{w_{\textnormal{low}}}
\newcommand{\wcfghigh}{w_{\textnormal{high}}}

\newcommand{\pred}{D_{\mtheta}(\vz_t, t, \vy)}

\newcommand{\prednull}{D_{\mtheta}(\vz_t, t,\vy_{\textnormal{null}})}

\newcommand{\predcfg}{\hat{D}_{\textrm{CFG}}(\vz_t, t, \vy)}

\newcommand{\predcond}{D_c(\vz_t)}
\newcommand{\preduncond}{D_u(\vz_t)}
\newcommand{\predguided}{\hat{D}_{\textnormal{CFG}}(\vz_t)}
\newcommand{\predguidedours}{\hat{D}_{\textnormal{\gls{method}}}(\vz_t)}

\newcommand{\predguidedlow}{\hat{D}^{\textnormal{low}}_{\textnormal{CFG}}(\vz_t)}
\newcommand{\predguidedhigh}{\hat{D}^{\textnormal{high}}_{\textnormal{CFG}}(\vz_t)}

\newcommand{\predguidedlowfdg}{\hat{D}^{\textnormal{low}}_{\textnormal{\acrshort{method}}}(\vz_t)}
\newcommand{\predguidedhighfdg}{\hat{D}^{\textnormal{high}}_{\textnormal{\acrshort{method}}}(\vz_t)}

\DeclareMathOperator{\oplow}{\psi_{\textnormal{low}}}
\DeclareMathOperator{\ophigh}{\psi_{\textnormal{high}}}
\DeclareMathOperator{\opfreq}{\psi}
\DeclareMathOperator{\opfreqinv}{\psi^{-1}}

\usepackage{amsmath,amsfonts,bm, mathtools}

\newcommand{\mc}[1]{\mathcal{#1}}

\def\eqref#1{equation~\ref{#1}}

\def\1{\bm{1}}

\newcommand{\dd}{\mathrm{d}}

\def\vzero{{\bm{0}}}

\def\vx{{\bm{x}}}
\def\vy{{\bm{y}}}
\def\vz{{\bm{z}}}

\def\mI{{\pmb{I}}}

\DeclareMathAlphabet{\mathsfit}{\encodingdefault}{\sfdefault}{m}{sl}
\SetMathAlphabet{\mathsfit}{bold}{\encodingdefault}{\sfdefault}{bx}{n}

\def\mepsilon{{\bm{\epsilon}}}

\def\mtheta{{\bm{\theta}}}

\newcommand{\pdata}{p_{\textnormal{data}}}

\newcommand{\norm}[1]{\left\lVert #1 \right\rVert}

\newcommand{\trp}[1]{#1^{\top}}

\newcommand{\normal}[2]{\mc{N}\prn{#1, #2}}

\DeclarePairedDelimiterX{\infdivx}[2]{(}{)}{%
  #1\delimsize\|#2%
}

\newcommand{\brk}[1]{\left[ #1 \right]}

\newcommand{\prn}[1]{\left( #1 \right)}

\newcommand{\zero}{\pmb{0}}

\DeclareDocumentCommand{\ex}{m o}{
   \mathbb{E}\IfValueT{#2}{_{#2}}\left[#1\right]
}

\DeclareMathOperator{\grad}{\nabla}

\DeclarePairedDelimiterX\Set[1]{\lbrace}{\rbrace}%
 {  #1 }

\def\ddefloop#1{\ifx\ddefloop#1\else\ddef{#1}\expandafter\ddefloop\fi}
\def\ddef#1{\expandafter\def\csname #1bb\endcsname{\ensuremath{\mathbb{#1}}}}
\ddefloop ABCDEFGHIJKLMNOPQRSTUVWXYZ\ddefloop
\def\ddefloop#1{\ifx\ddefloop#1\else\ddef{#1}\expandafter\ddefloop\fi}
\def\ddef#1{\expandafter\def\csname #1b\endcsname{\ensuremath{\mathbf{#1}}}}
\ddefloop ABCDEFGHIJKLMNOPQRSTUVWXYZ\ddefloop
\def\ddef#1{\expandafter\def\csname #1c\endcsname{\ensuremath{\mathcal{#1}}}}
\ddefloop ABCDEFGHIJKLMNOPQRSTUVWXYZ\ddefloop
\def\ddef#1{\expandafter\def\csname #1hat\endcsname{\ensuremath{\widehat{#1}}}}
\ddefloop ABCDEFGHIJKLMNOPQRSTUVWXYZ\ddefloop
\def\ddef#1{\expandafter\def\csname hc#1\endcsname{\ensuremath{\widehat{\mathcal{#1}}}}}
\ddefloop ABCDEFGHIJKLMNOPQRSTUVWXYZ\ddefloop
\def\ddef#1{\expandafter\def\csname #1til\endcsname{\ensuremath{\widetilde{#1}}}}
\ddefloop ABCDEFGHIJKLMNOPQRSTUVWXYZ\ddefloop
\def\ddef#1{\expandafter\def\csname tc#1\endcsname{\ensuremath{\widetilde{\mathcal{#1}}}}}
\ddefloop ABCDEFGHIJKLMNOPQRSTUVWXYZ\ddefloop
\def\ddef#1{\expandafter\def\csname #1Bar\endcsname{\ensuremath{\bar{#1}}}}
\ddefloop ABCDEFGHIJKLMNOPQRSTUVWXYZ\ddefloop

\newacronym{ldm}{LDM}{latent diffusion model}
\newacronym{vae}{VAE}{variational autoencoder}
\newacronym{sdvae}{SD-VAE}{Stable Diffusion VAE}
\newacronym{method}{FDG}{frequency-decoupled guidance}
\newacronym{cfg}{CFG}{classifier-free guidance}

\newacronym{apg}{APG}{adaptive projected guidance}

\newcommand{\figTeaser}{
    \begin{figure}[h!]
        \centering
        \begin{minipage}{\textwidth}
            \begin{subfigure}{0.493\textwidth}
                \begin{tikzpicture}
                    \node[anchor=south west, inner sep=0] (image) at (0,0) {
                            \includegraphics[width=\textwidth]{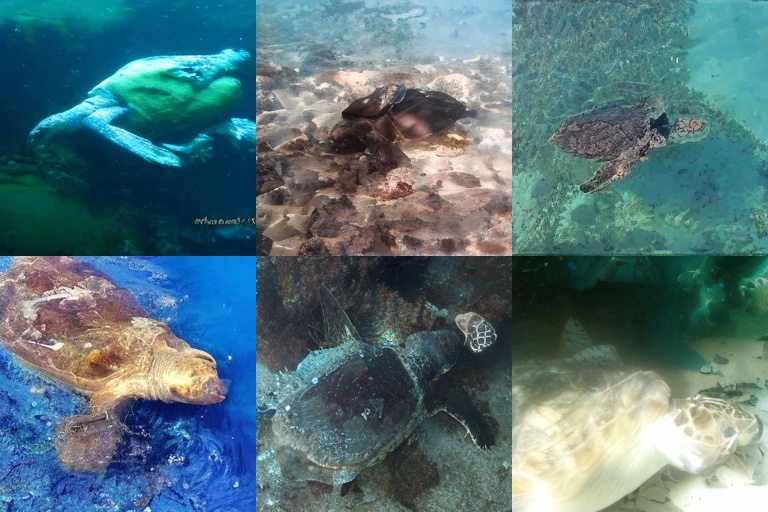}
                    };
                    \node[rotate=0, align=center, font=\normalsize, anchor=base, yshift=4pt] at (image.north) {CFG \strut};
                \end{tikzpicture}
            \end{subfigure}
            \hfill  
            \begin{subfigure}{0.493\textwidth}
                \begin{tikzpicture}
                    \node[anchor=south west, inner sep=0] (image) at (0,0) {
                \includegraphics[width=\textwidth]{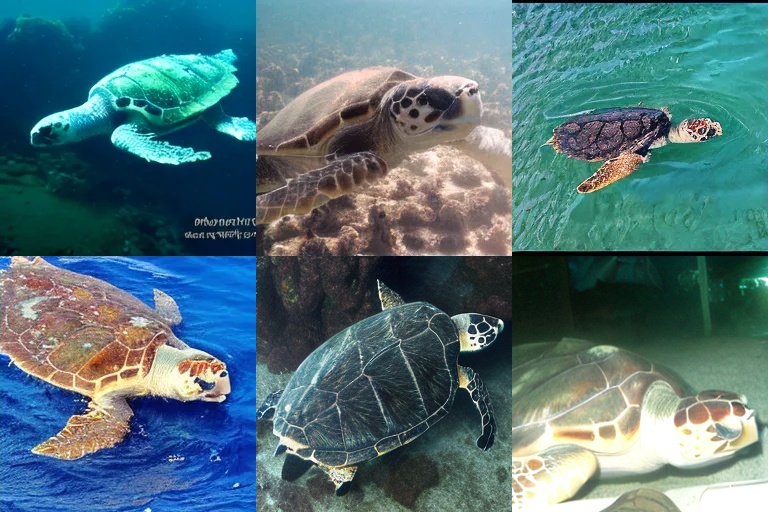}
                    };
                    \node[rotate=0, align=center, font=\normalsize, anchor=base, yshift=4pt] at (image.north) {\acrshort{method} (Ours) \strut};
                \end{tikzpicture}
            \end{subfigure}
        \end{minipage}
        \caption{Low classifier-free guidance produces images with good diversity and color composition but often results in low-quality and blurry generations. We propose \acrfull{method}, a novel modification to the CFG update rule in the frequency domain that significantly improves image quality at low guidance scales while, by design, avoiding common issues at high CFG scales such as reduced diversity. Images are generated using class-conditional DiT-XL/2 \citep{peeblesScalableDiffusionModels2022} with a guidance scale of $1.2$.}
        \label{fig:teaser}

    \end{figure}
}

\newcommand{\CFGDecompFig}{
\begin{figure}[t]
    \centering
        \begin{subfigure}[b]{0.245\textwidth}
            \begin{tikzpicture}
                        \node[anchor=south west, inner sep=0] (image) at (0,0) {
                            \includegraphics[width=\linewidth]{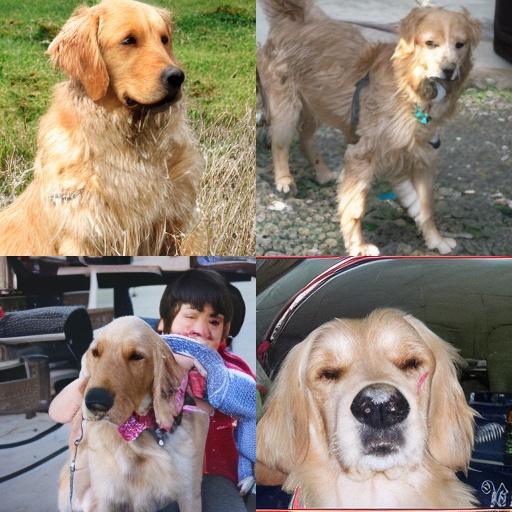}
                        };
                        \node[rotate=0, align=center, font=\small, anchor=base, yshift=4pt] at (image.north) {$\wcfglow=1.5$, $\wcfghigh=1.5$ \\ \fontsize{8pt}{14pt}\selectfont{\textcolor{C3}{low quality}, \textcolor{C2}{high diversity}} \strut};
                    \end{tikzpicture}
        \end{subfigure}
        \hfill
        \begin{subfigure}[b]{0.245\textwidth}
            \begin{tikzpicture}
                \node[anchor=south west, inner sep=0] (image) at (0,0) {
                    \includegraphics[width=\linewidth]{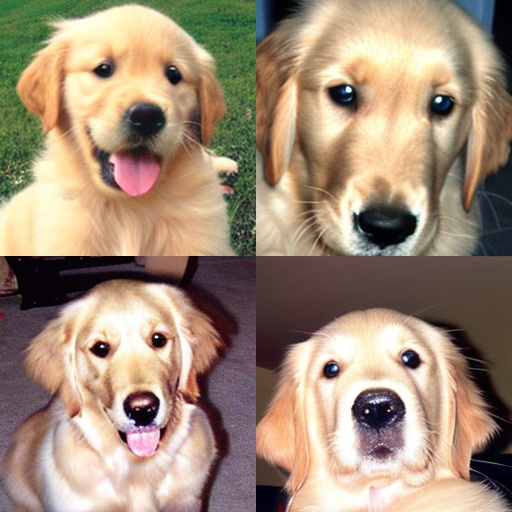}
                };
                \node[rotate=0, align=center, font=\small, anchor=base, yshift=4pt] at (image.north) {$\wcfglow=7$, $\wcfghigh=7$  \\ \fontsize{8pt}{14pt}\selectfont{\textcolor{C2}{high quality}, \textcolor{C3}{low diversity}} \strut};
            \end{tikzpicture}
            
        \end{subfigure}
        \hfill
        \begin{subfigure}[b]{0.245\textwidth}
            \begin{tikzpicture}
                \node[anchor=south west, inner sep=0] (image) at (0,0) {
                    \includegraphics[width=\linewidth]{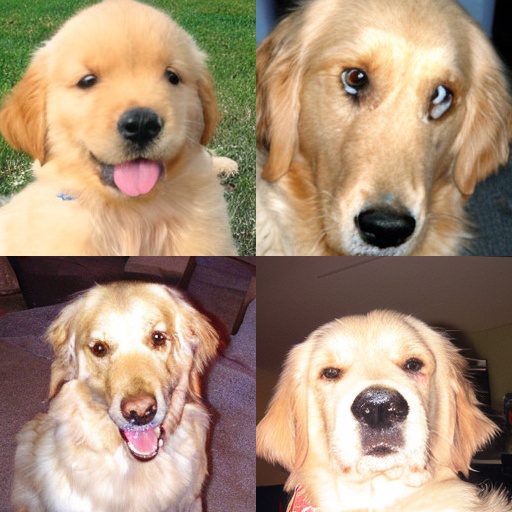}
                };
                \node[rotate=0, align=center, font=\small, anchor=base, yshift=4pt] at (image.north) {$\wcfglow=7$, $\wcfghigh=1.5$  \\ \fontsize{8pt}{14pt}\selectfont{\textcolor{C3}{low quality}, \textcolor{C3}{low diversity}} \strut};
            \end{tikzpicture}
        \end{subfigure}
        \hfill
        \begin{subfigure}[b]{0.245\textwidth}
            \begin{tikzpicture}
                \node[anchor=south west, inner sep=0] (image) at (0,0) {
                    \includegraphics[width=\linewidth]{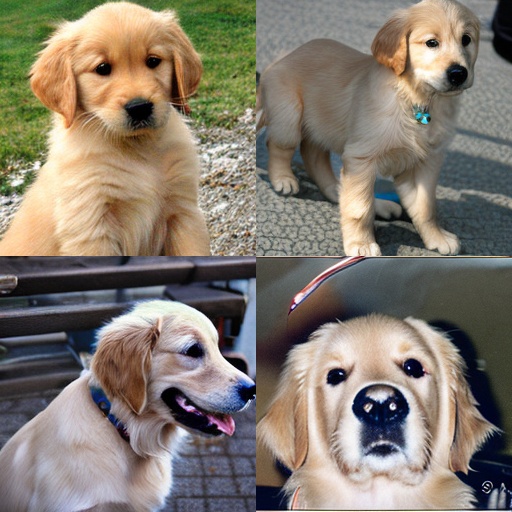}
                };
                \node[rotate=0, align=center, font=\small, anchor=base, yshift=4pt] at (image.north) {$\wcfglow=1.5$, $\wcfghigh=7$ \\ \fontsize{8pt}{14pt}\selectfont{\textcolor{C2}{high quality}, \textcolor{C2}{high diversity}} \strut};
            \end{tikzpicture}
        \end{subfigure}

    \caption{Illustration of how different frequency components of CFG affect generation. Sampling with low CFG scales results in diverse generations but lower overall quality. Increasing the CFG scale improves quality but reduces diversity. We show that the low-frequency component of the CFG signal primarily drives the reduction in diversity, while the high-frequency component contributes to quality enhancement without affecting diversity.}
    \label{fig:method}
\end{figure}

}

\newcommand{\figMain}{
    \begin{figure}[t]
        \vspace*{-0.75cm}
        \centering
        \begin{minipage}{\textwidth}
            \begin{subfigure}{0.245\textwidth}
                \begin{tikzpicture}
                    \node[anchor=south west, inner sep=0] (image) at (0,0) {
                        \includegraphics[width=\textwidth]{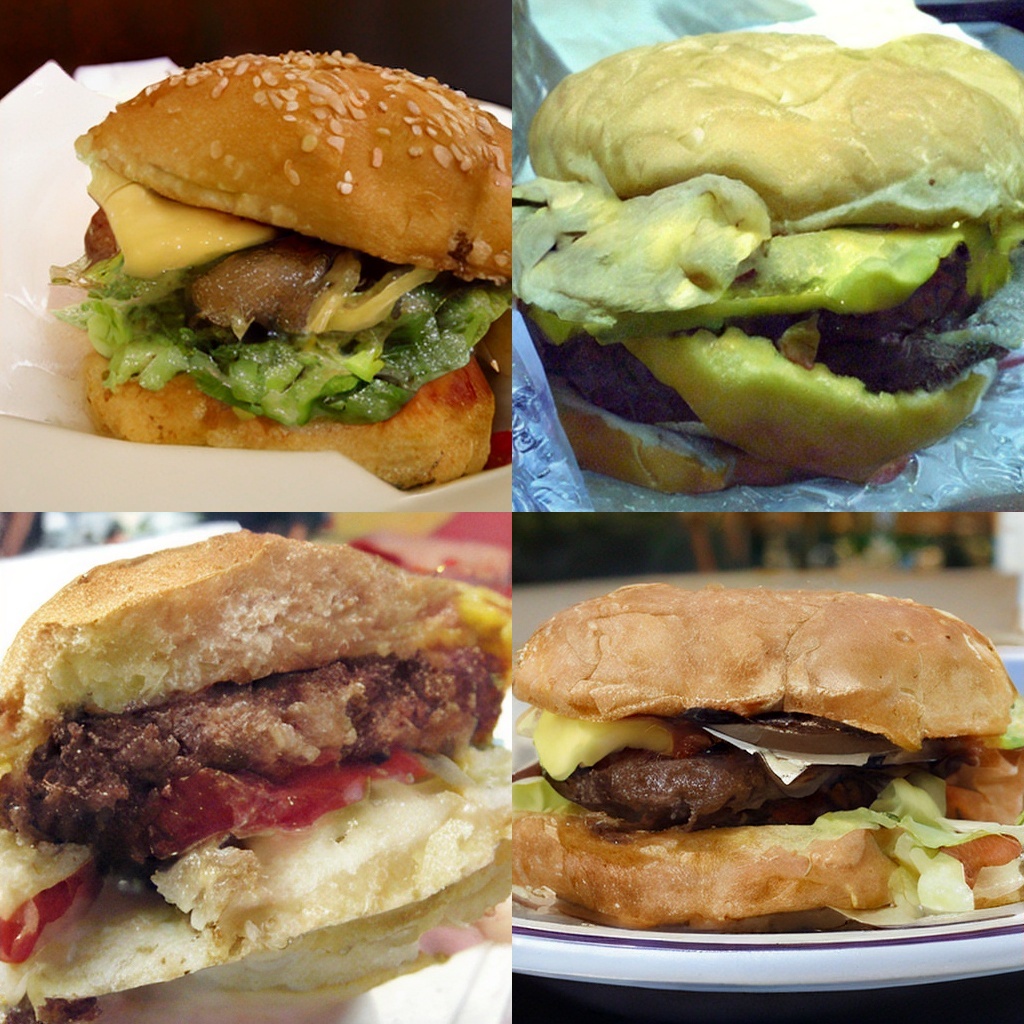}
                    };
                    \node[rotate=0, align=center, font=\normalsize, anchor=base, yshift=4pt] at (image.north) {CFG \strut};
                \end{tikzpicture}
            \end{subfigure}
            \begin{subfigure}{0.245\textwidth}
                \begin{tikzpicture}
                    \node[anchor=south west, inner sep=0] (image) at (0,0) {
                        \includegraphics[width=\textwidth]{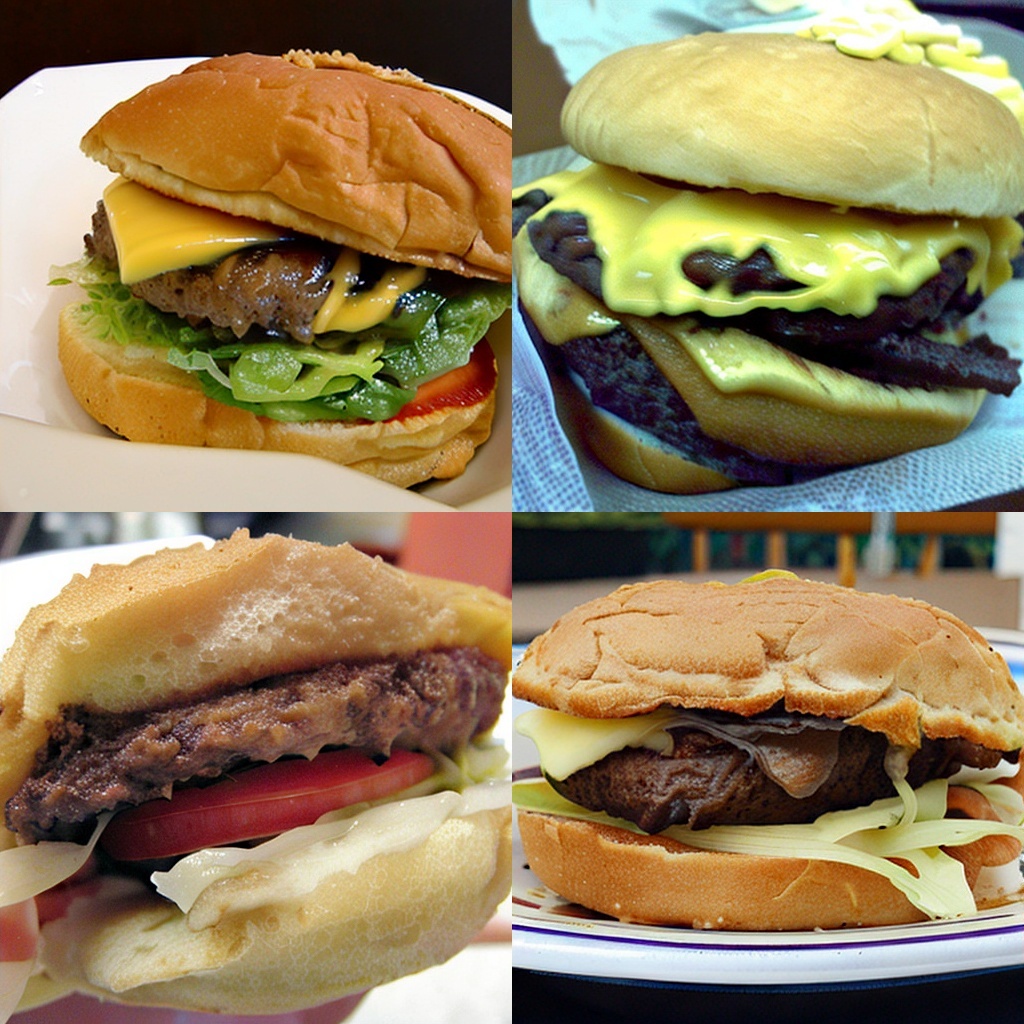}
                    };
                    \node[rotate=0, align=center, font=\normalsize, anchor=base, yshift=4pt] at (image.north) {\gls{method} (Ours) \strut};
                \end{tikzpicture}
            \end{subfigure}
            \begin{subfigure}{0.245\textwidth}
                \begin{tikzpicture}
                    \node[anchor=south west, inner sep=0] (image) at (0,0) {
                        \includegraphics[width=\textwidth]{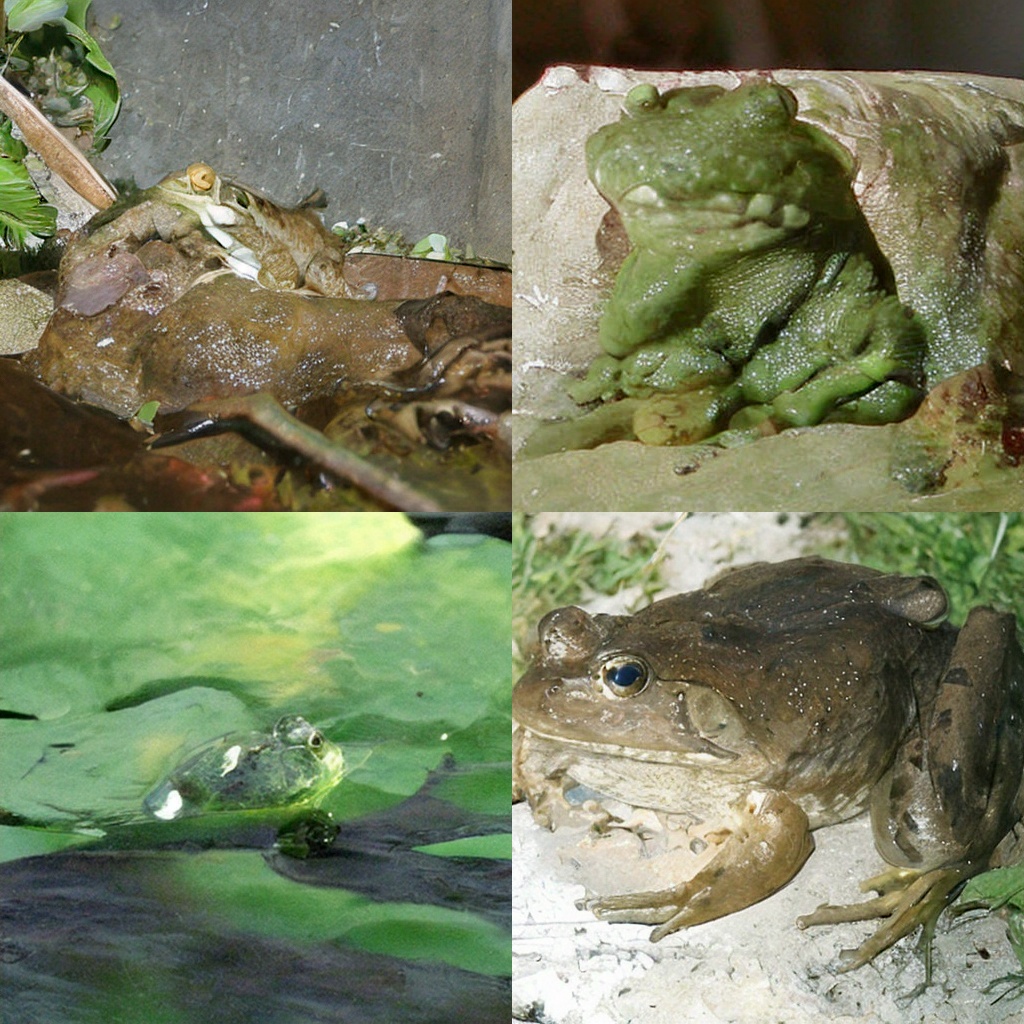}
                    };
                    \node[rotate=0, align=center, font=\normalsize, anchor=base, yshift=4pt] at (image.north) {CFG \strut};
                \end{tikzpicture}
            \end{subfigure}
            \begin{subfigure}{0.245\textwidth}
                \begin{tikzpicture}
                    \node[anchor=south west, inner sep=0] (image) at (0,0) {
                        \includegraphics[width=\textwidth]{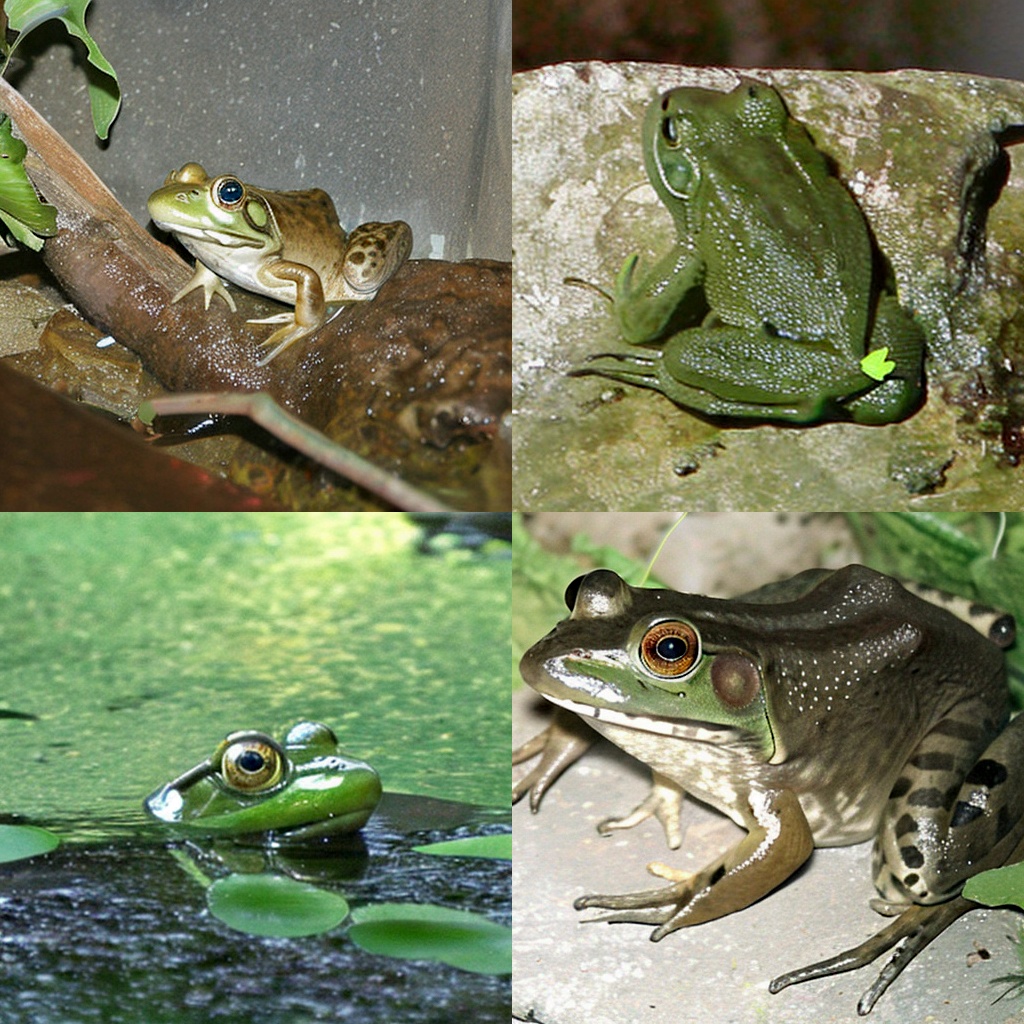}
                    };
                    \node[rotate=0, align=center, font=\normalsize, anchor=base, yshift=4pt] at (image.north) {\gls{method} (Ours) \strut};
                \end{tikzpicture}
            \end{subfigure}
            \begin{subfigure}{0.245\textwidth}
                \begin{tikzpicture}
                    \node[anchor=south west, inner sep=0] (image) at (0,0) {
                        \includegraphics[width=\textwidth]{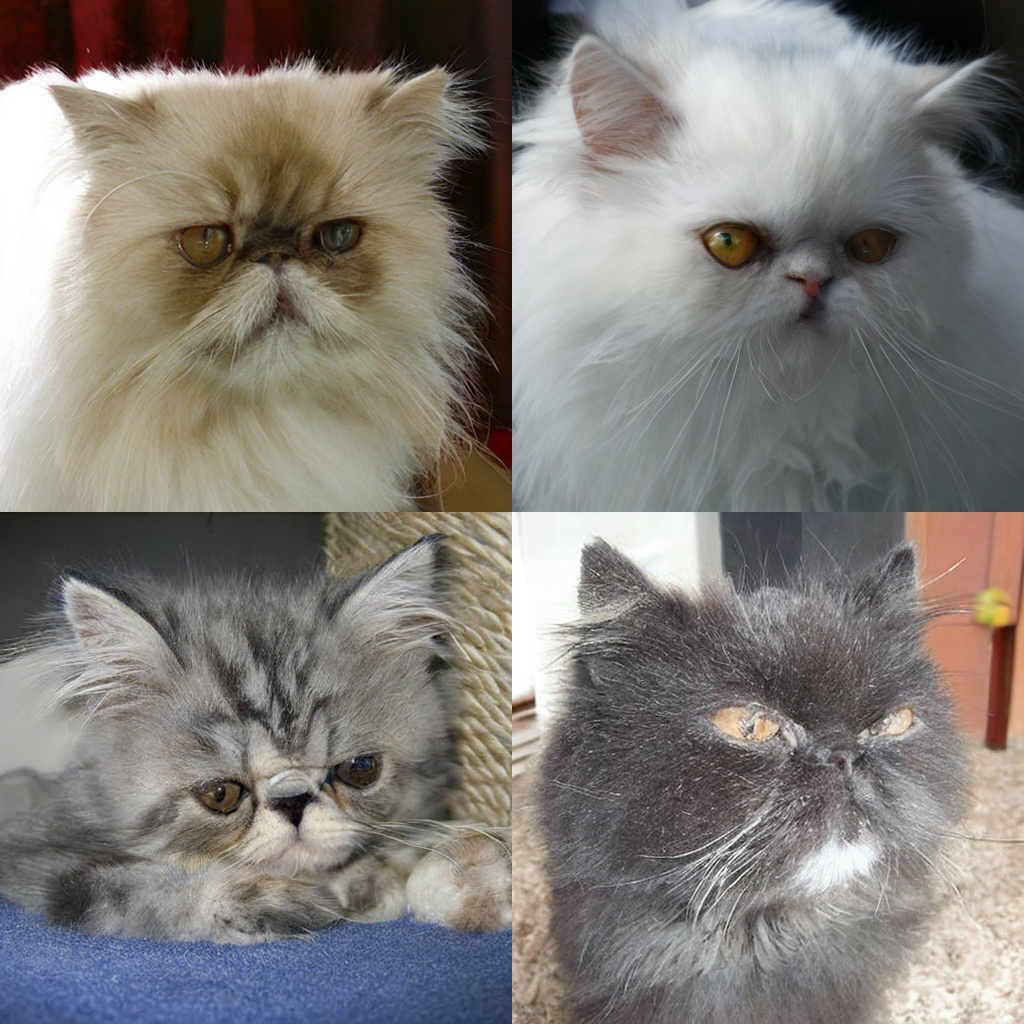}
                    };
                    \node[rotate=0, align=center, font=\normalsize, anchor=base, yshift=4pt] at (image.north) {CFG \strut};
                \end{tikzpicture}
            \end{subfigure}
            \begin{subfigure}{0.245\textwidth}
                \begin{tikzpicture}
                    \node[anchor=south west, inner sep=0] (image) at (0,0) {
                        \includegraphics[width=\textwidth]{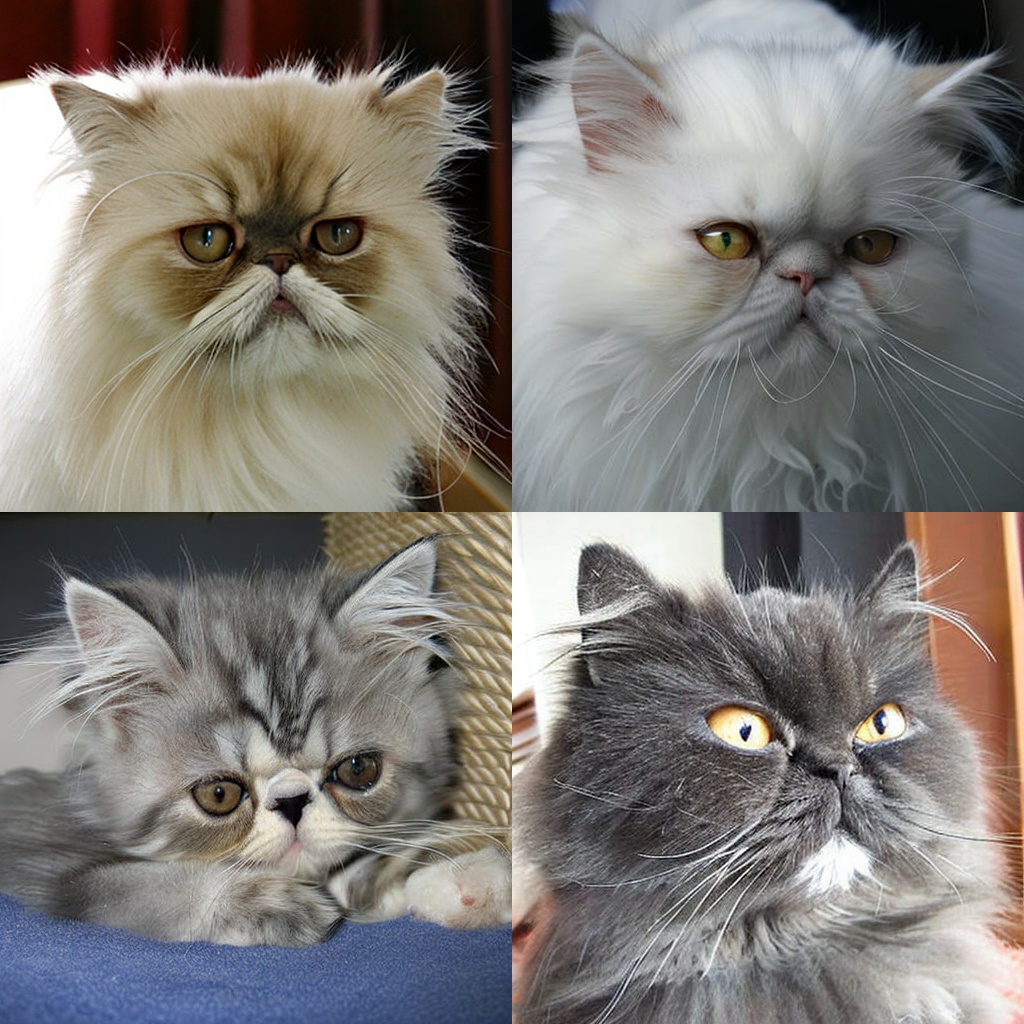}
                    };
                    \node[rotate=0, align=center, font=\normalsize, anchor=base, yshift=4pt] at (image.north) {\gls{method} (Ours) \strut};
                \end{tikzpicture}
            \end{subfigure}
            \begin{subfigure}{0.245\textwidth}
                \begin{tikzpicture}
                    \node[anchor=south west, inner sep=0] (image) at (0,0) {
                        \includegraphics[width=\textwidth]{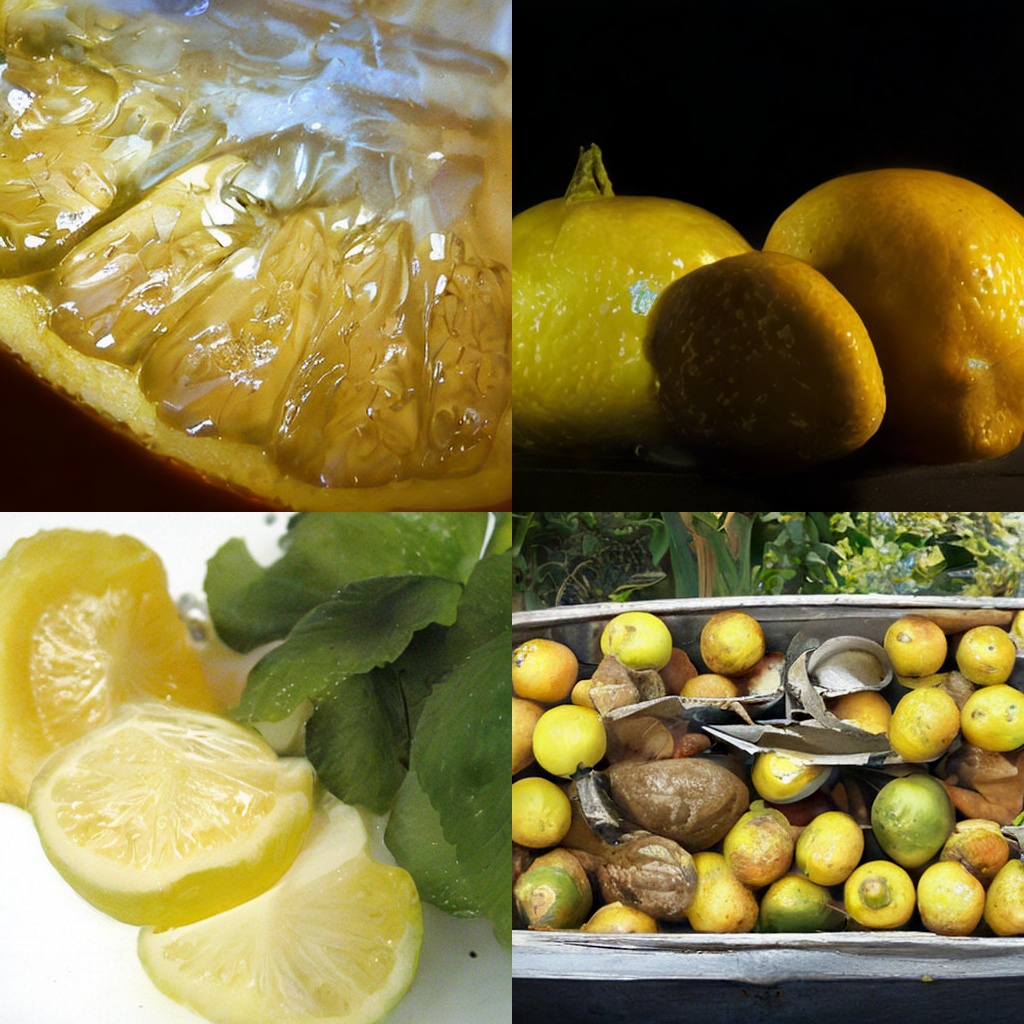}
                    };
                    \node[rotate=0, align=center, font=\normalsize, anchor=base, yshift=4pt] at (image.north) {CFG \strut};
                \end{tikzpicture}
            \end{subfigure}
            \begin{subfigure}{0.245\textwidth}
                \begin{tikzpicture}
                    \node[anchor=south west, inner sep=0] (image) at (0,0) {
                        \includegraphics[width=\textwidth]{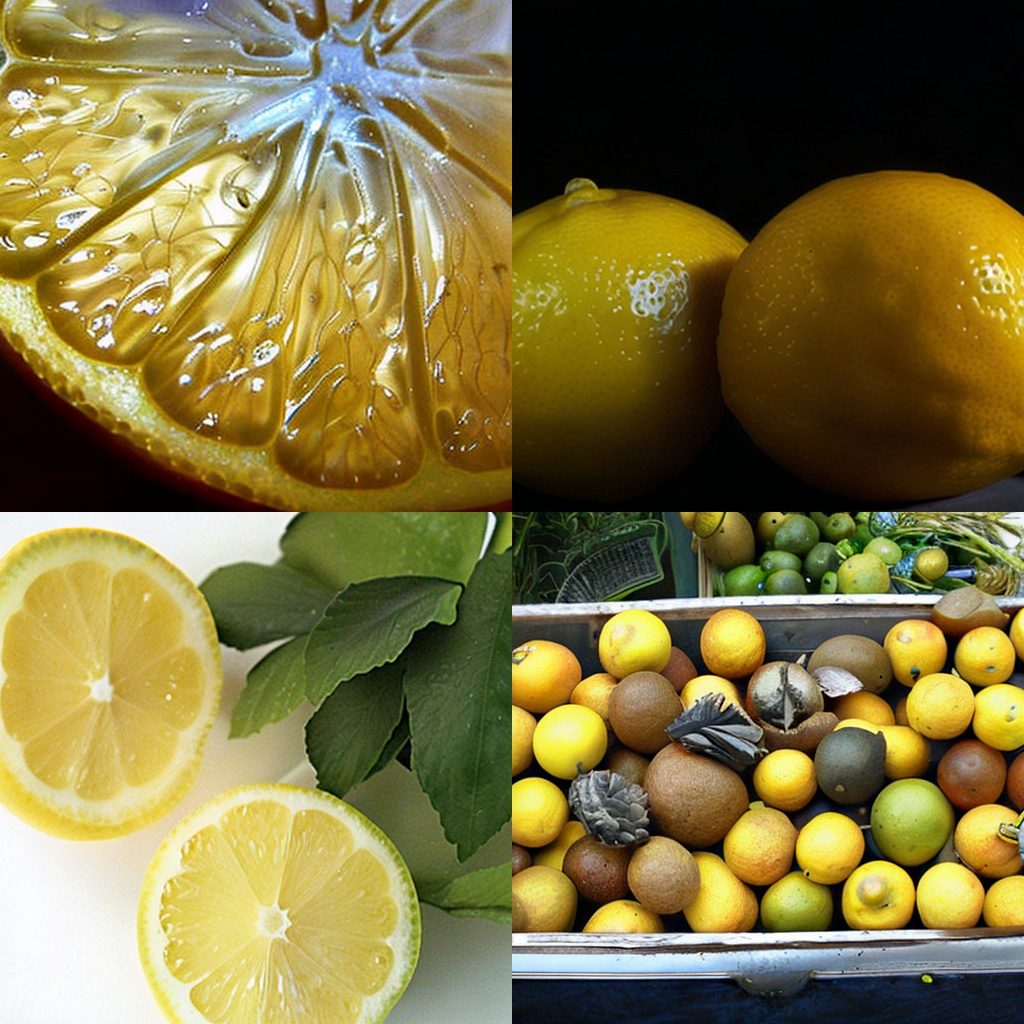}
                    };
                    \node[rotate=0, align=center, font=\normalsize, anchor=base, yshift=4pt] at (image.north) {\gls{method} (Ours) \strut};
                \end{tikzpicture}
            \end{subfigure}
        \end{minipage}
        \caption{Class-conditional generation results using EDM2 with $\wcfg=\wcfglow=1.25$. \gls{method} enhances image quality while maintaining the diversity of low CFG scales.}
        \label{fig:edm-main}

        \begin{minipage}{\textwidth}
            \begin{subfigure}{0.245\textwidth}
                \begin{tikzpicture}
                    \node[anchor=south west, inner sep=0] (image) at (0,0) {
                        \includegraphics[width=\textwidth]{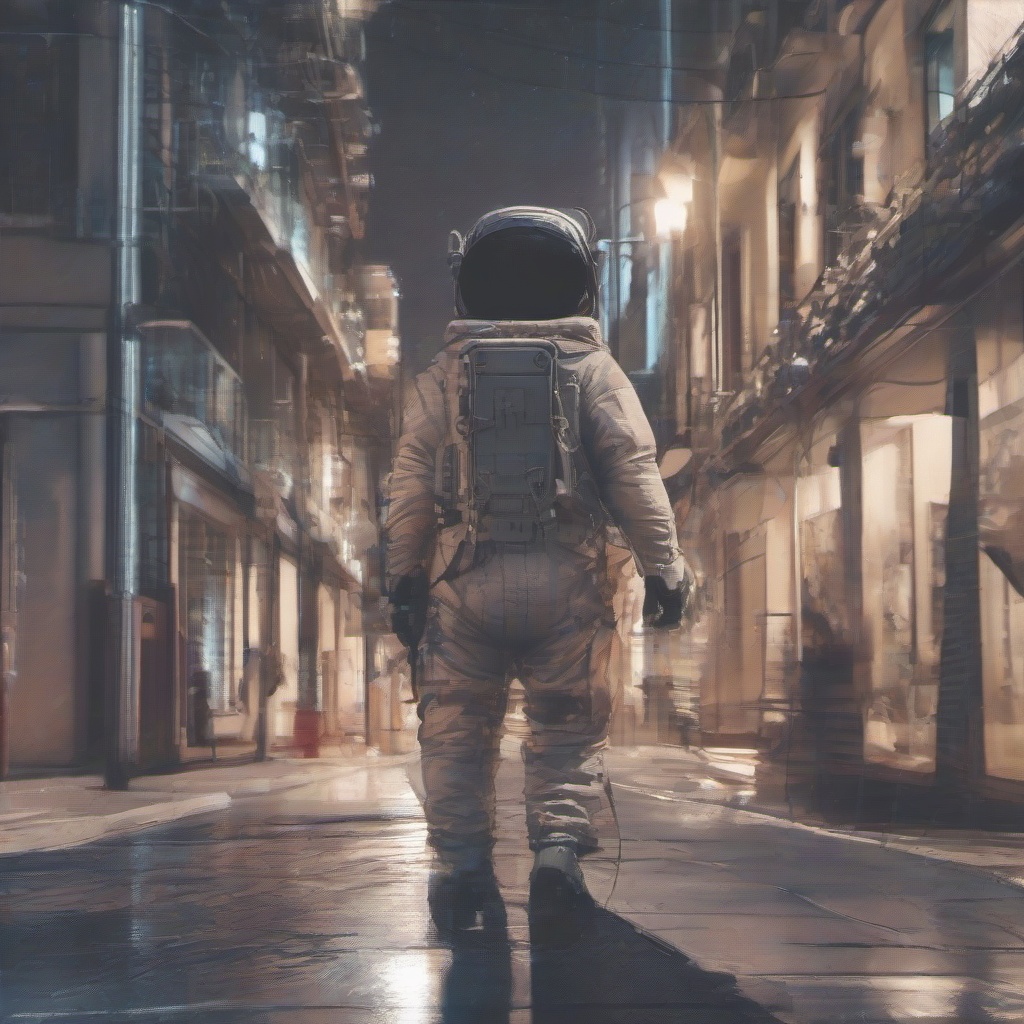}
                    };
                    \node[rotate=0, align=center, font=\normalsize, anchor=base, yshift=4pt] at (image.north) {CFG \strut};
                \end{tikzpicture}
            \end{subfigure}
            \begin{subfigure}{0.245\textwidth}
                \begin{tikzpicture}
                    \node[anchor=south west, inner sep=0] (image) at (0,0) {
                        \includegraphics[width=\textwidth]{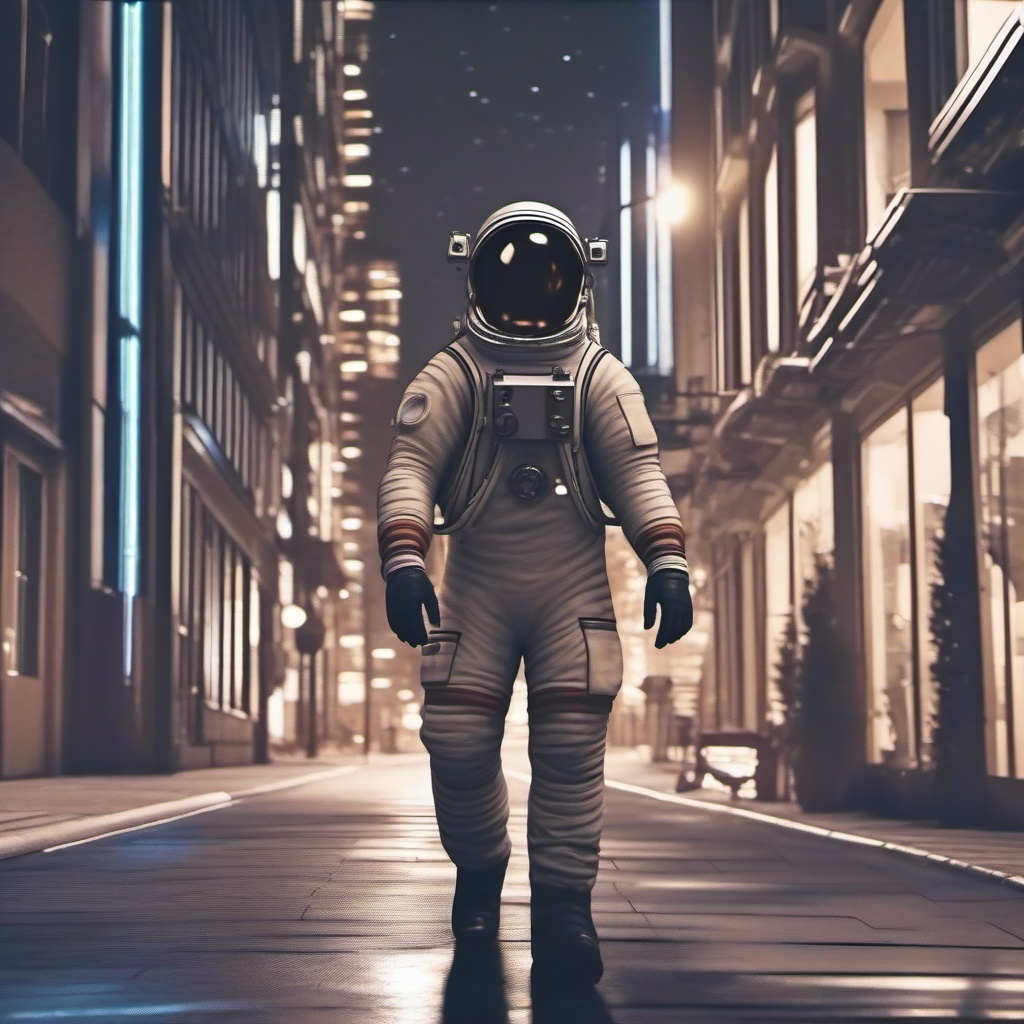}
                    };
                    \node[rotate=0, align=center, font=\normalsize, anchor=base, yshift=4pt] at (image.north) {\gls{method} (Ours) \strut};
                \end{tikzpicture}
            \end{subfigure}
            \begin{subfigure}{0.245\textwidth}
                \begin{tikzpicture}
                    \node[anchor=south west, inner sep=0] (image) at (0,0) {
                        \includegraphics[width=\textwidth]{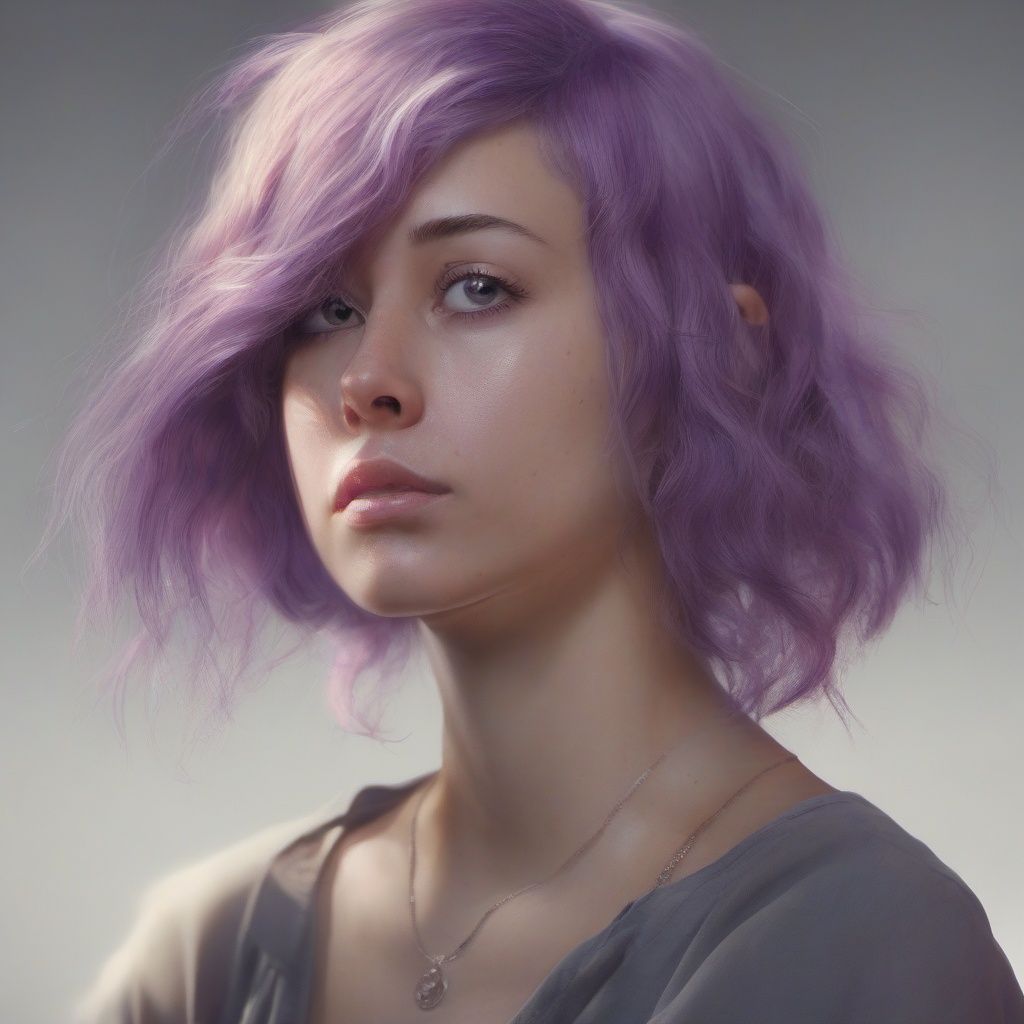}
                    };
                    \node[rotate=0, align=center, font=\normalsize, anchor=base, yshift=4pt] at (image.north) {CFG \strut};
                \end{tikzpicture}
            \end{subfigure}
            \begin{subfigure}{0.245\textwidth}
                \begin{tikzpicture}
                    \node[anchor=south west, inner sep=0] (image) at (0,0) {
                        \includegraphics[width=\textwidth]{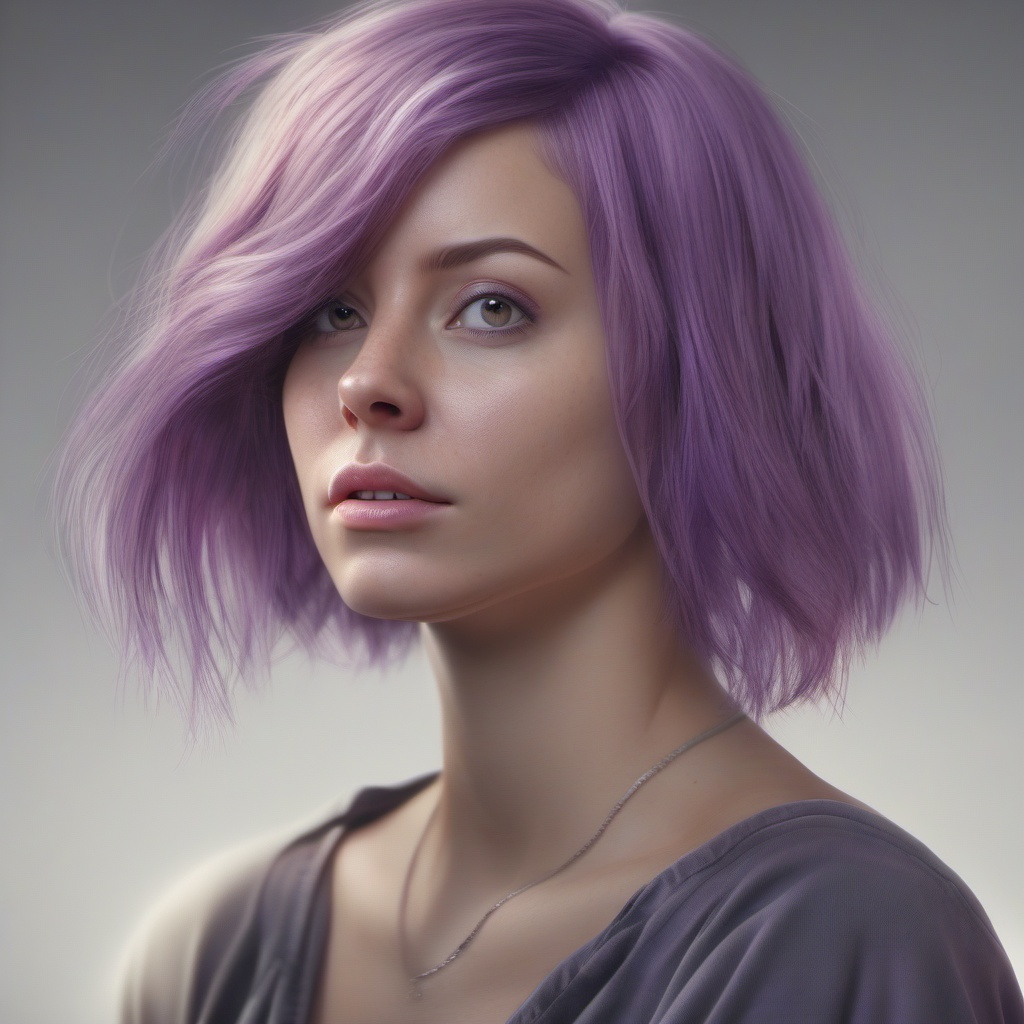}
                    };
                    \node[rotate=0, align=center, font=\normalsize, anchor=base, yshift=4pt] at (image.north) {\gls{method} (Ours) \strut};
                \end{tikzpicture}
            \end{subfigure}
            \begin{subfigure}{0.245\textwidth}
                \begin{tikzpicture}
                    \node[anchor=south west, inner sep=0] (image) at (0,0) {
                        \includegraphics[width=\textwidth]{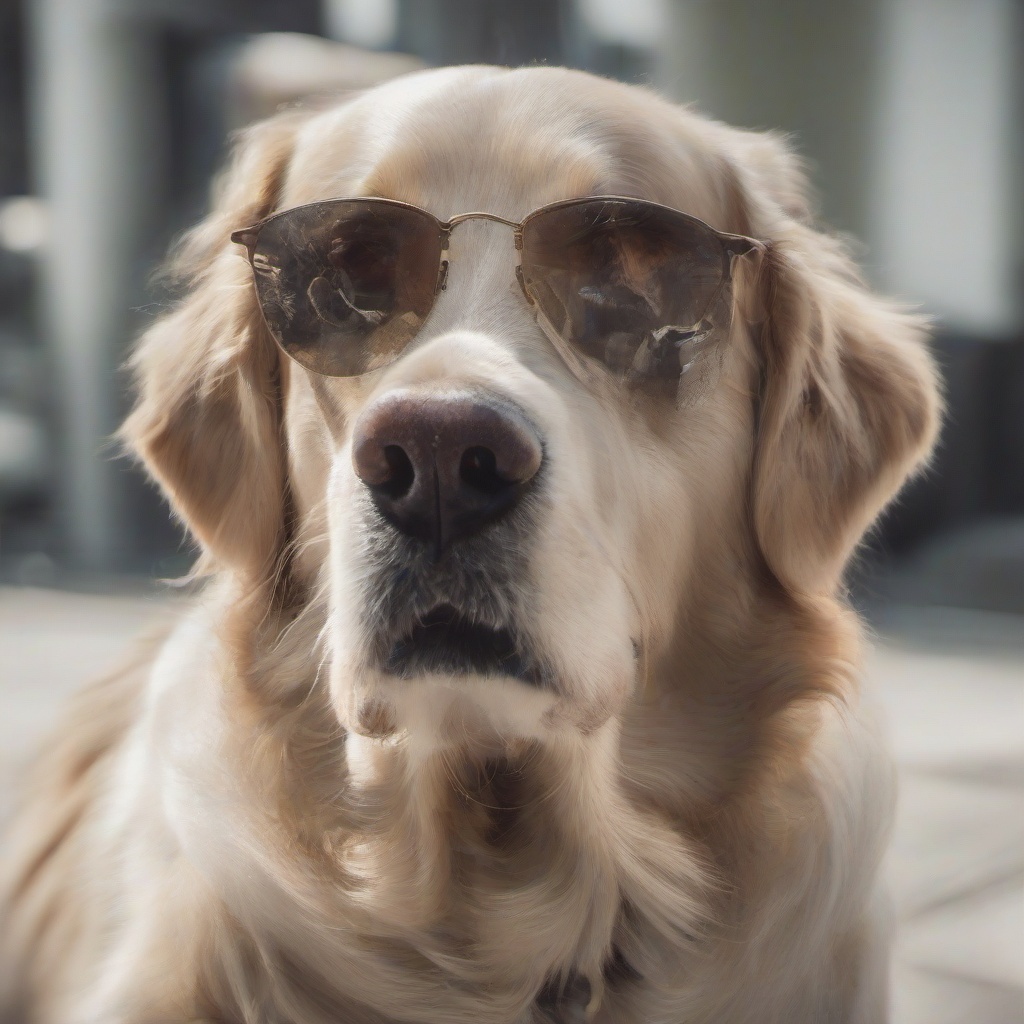}
                    };
                    \node[rotate=0, align=center, font=\normalsize, anchor=base, yshift=4pt] at (image.north) {CFG \strut};
                \end{tikzpicture}
            \end{subfigure}
            \begin{subfigure}{0.245\textwidth}
                \begin{tikzpicture}
                    \node[anchor=south west, inner sep=0] (image) at (0,0) {
                        \includegraphics[width=\textwidth]{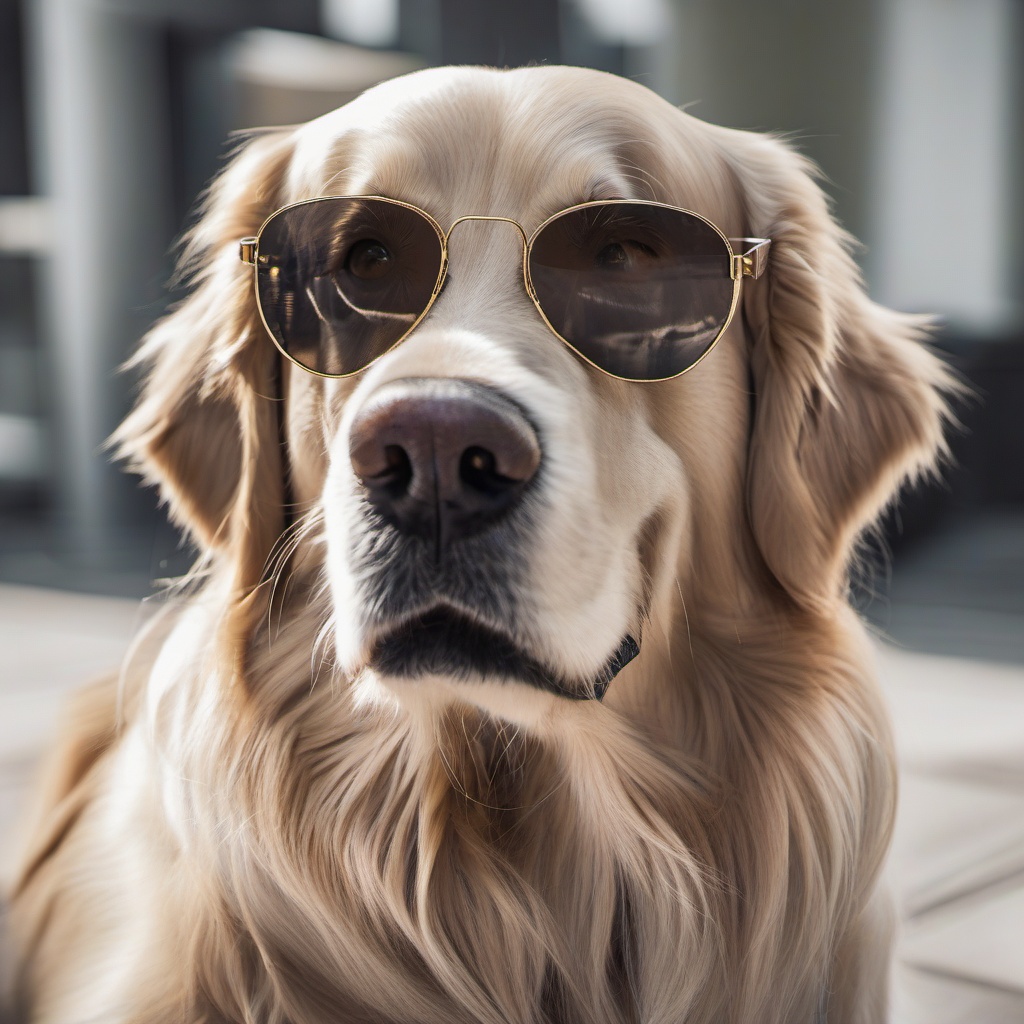}
                    };
                    \node[rotate=0, align=center, font=\normalsize, anchor=base, yshift=4pt] at (image.north) {\gls{method} (Ours) \strut};
                \end{tikzpicture}
            \end{subfigure}
            \begin{subfigure}{0.245\textwidth}
                \begin{tikzpicture}
                    \node[anchor=south west, inner sep=0] (image) at (0,0) {
                        \includegraphics[width=\textwidth]{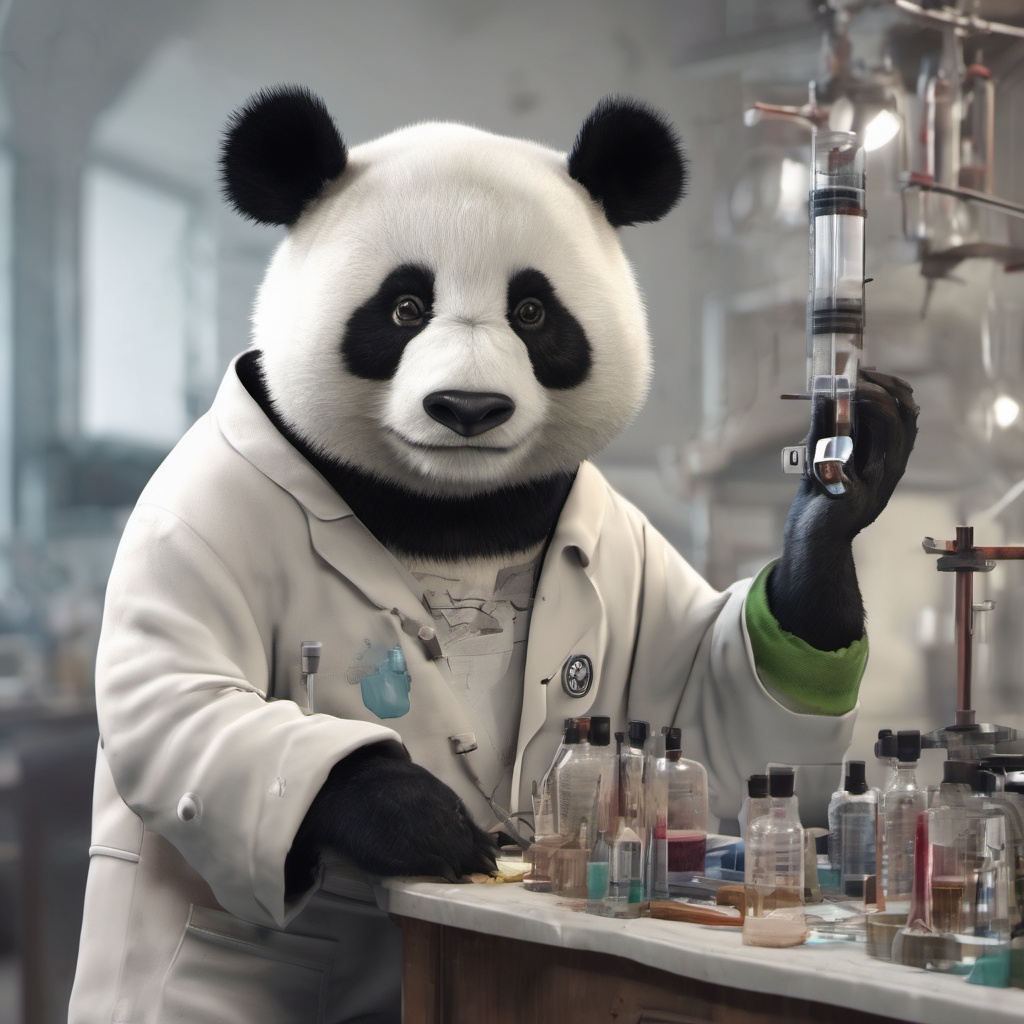}
                    };
                    \node[rotate=0, align=center, font=\normalsize, anchor=base, yshift=4pt] at (image.north) {CFG \strut};
                \end{tikzpicture}
            \end{subfigure}
            \begin{subfigure}{0.245\textwidth}
                \begin{tikzpicture}
                    \node[anchor=south west, inner sep=0] (image) at (0,0) {
                        \includegraphics[width=\textwidth]{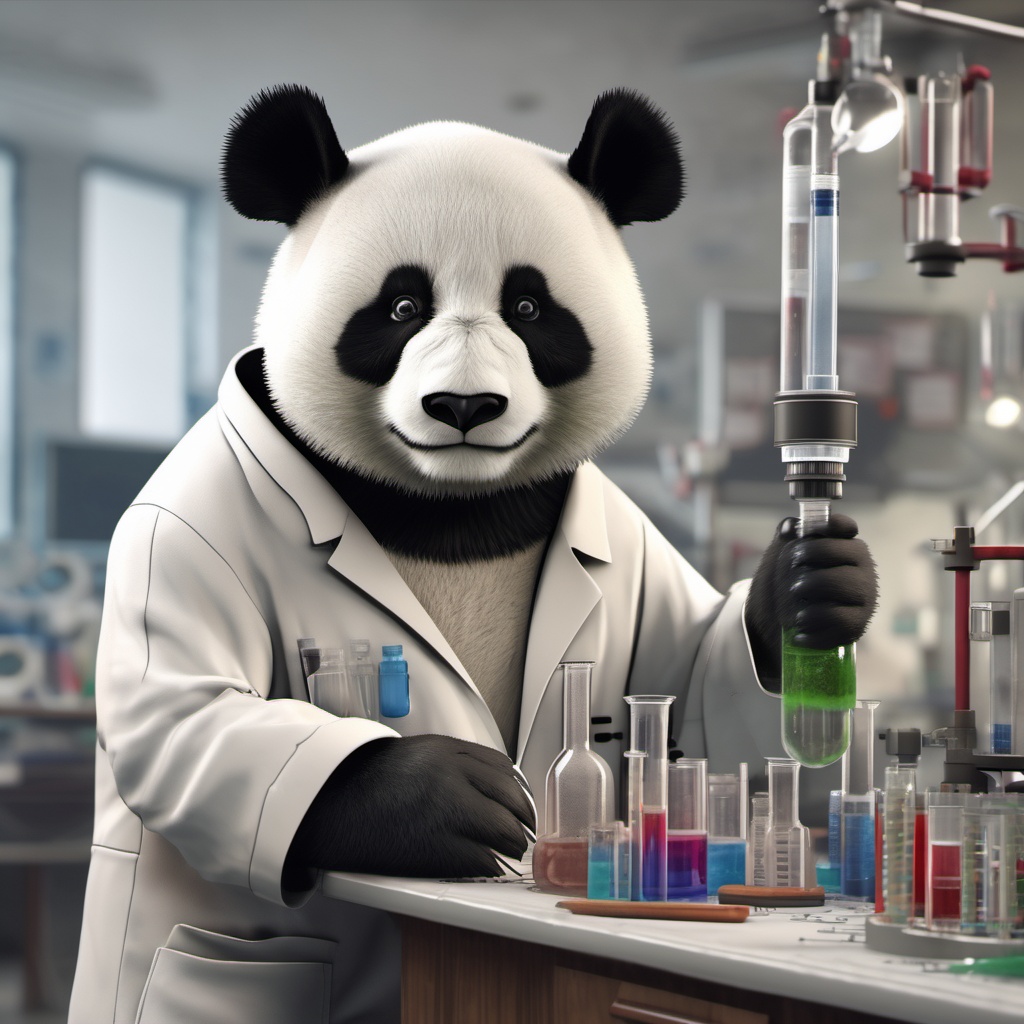}
                    };
                    \node[rotate=0, align=center, font=\normalsize, anchor=base, yshift=4pt] at (image.north) {\gls{method} (Ours) \strut};
                \end{tikzpicture}
            \end{subfigure}
        \end{minipage}
        \caption{Text-to-image generation results using Stable Diffusion XL with $\wcfg = \wcfglow = 2$. \gls{method} enhances the details of the CFG image while maintaining the overall structure and color palette.}
        \label{fig:sd-main}
        \vspace*{-0.4cm}

    \end{figure}
}

\newcommand{\figSDthreeMain}{
    \begin{figure}[t!]
        \centering
        \begin{minipage}{\textwidth}
            \begin{subfigure}{0.49\textwidth}
                \begin{subfigure}{0.495\textwidth}
                    \begin{tikzpicture}
                        \node[anchor=south west, inner sep=0] (image) at (0,0) {
                            \includegraphics[width=\textwidth]{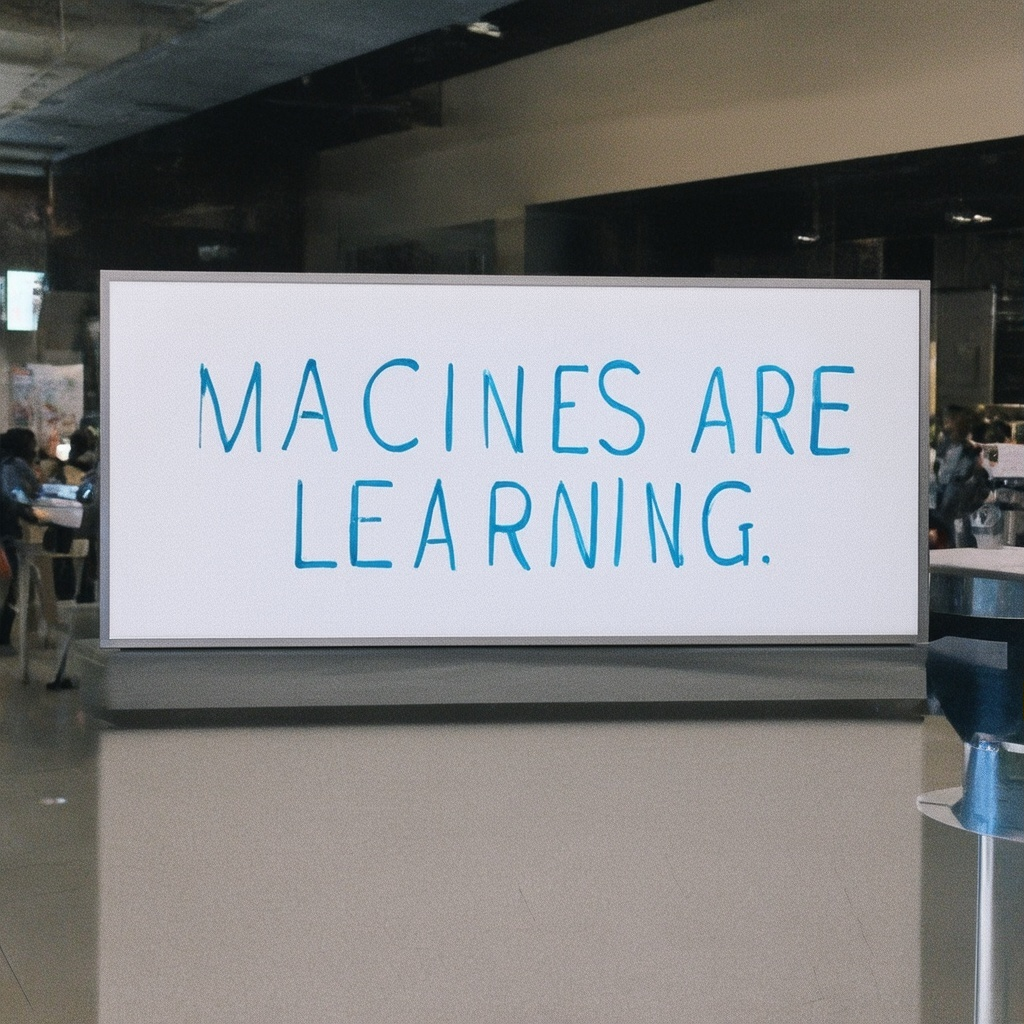}
                        };
                        \node[rotate=0, align=center, font=\normalsize, anchor=base, yshift=4pt] at (image.north) {CFG \strut};
                    \end{tikzpicture}
                \end{subfigure}
                \begin{subfigure}{0.495\textwidth}
                    \begin{tikzpicture}
                        \node[anchor=south west, inner sep=0] (image) at (0,0) {
                            \includegraphics[width=\textwidth]{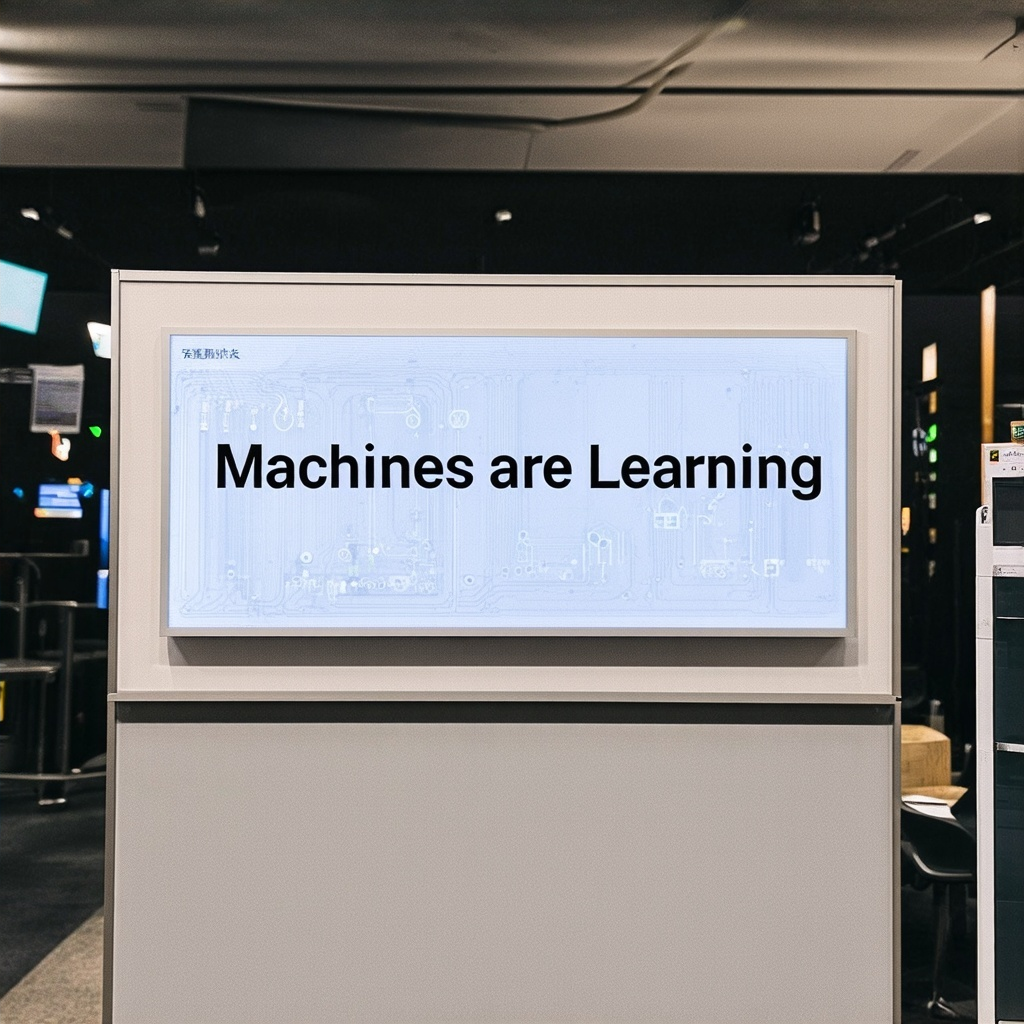}
                        };
                        \node[rotate=0, align=center, font=\normalsize, anchor=base, yshift=4pt] at (image.north) {\gls{method} (Ours) \strut};
                    \end{tikzpicture}
                \end{subfigure}
                \caption*{A sign that says ``Machines are Learning''}
            \end{subfigure}
            \begin{subfigure}{0.49\textwidth}
                \begin{subfigure}{0.495\textwidth}
                    \begin{tikzpicture}
                        \node[anchor=south west, inner sep=0] (image) at (0,0) {
                            \includegraphics[width=\textwidth]{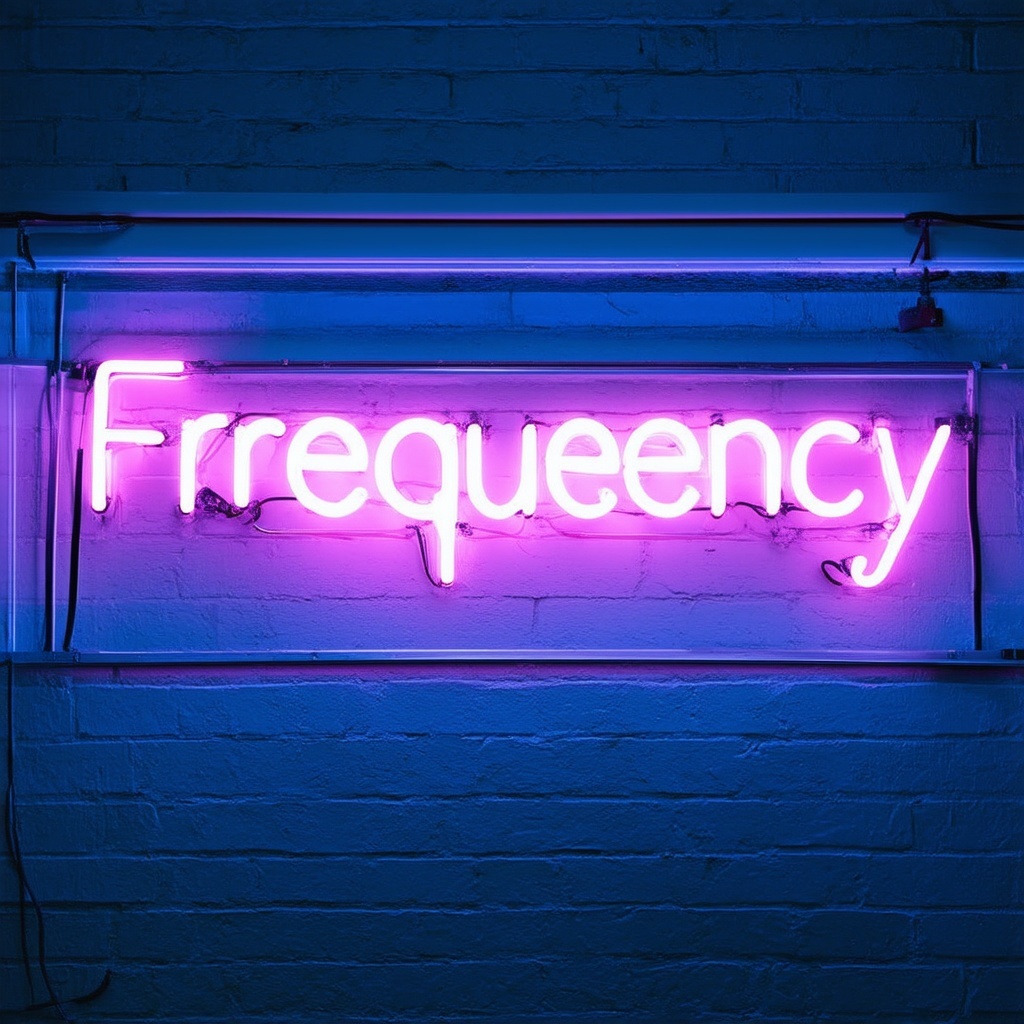}
                        };
                        \node[rotate=0, align=center, font=\normalsize, anchor=base, yshift=4pt] at (image.north) {CFG \strut};
                    \end{tikzpicture}
                \end{subfigure}
                \begin{subfigure}{0.495\textwidth}
                    \begin{tikzpicture}
                        \node[anchor=south west, inner sep=0] (image) at (0,0) {
                            \includegraphics[width=\textwidth]{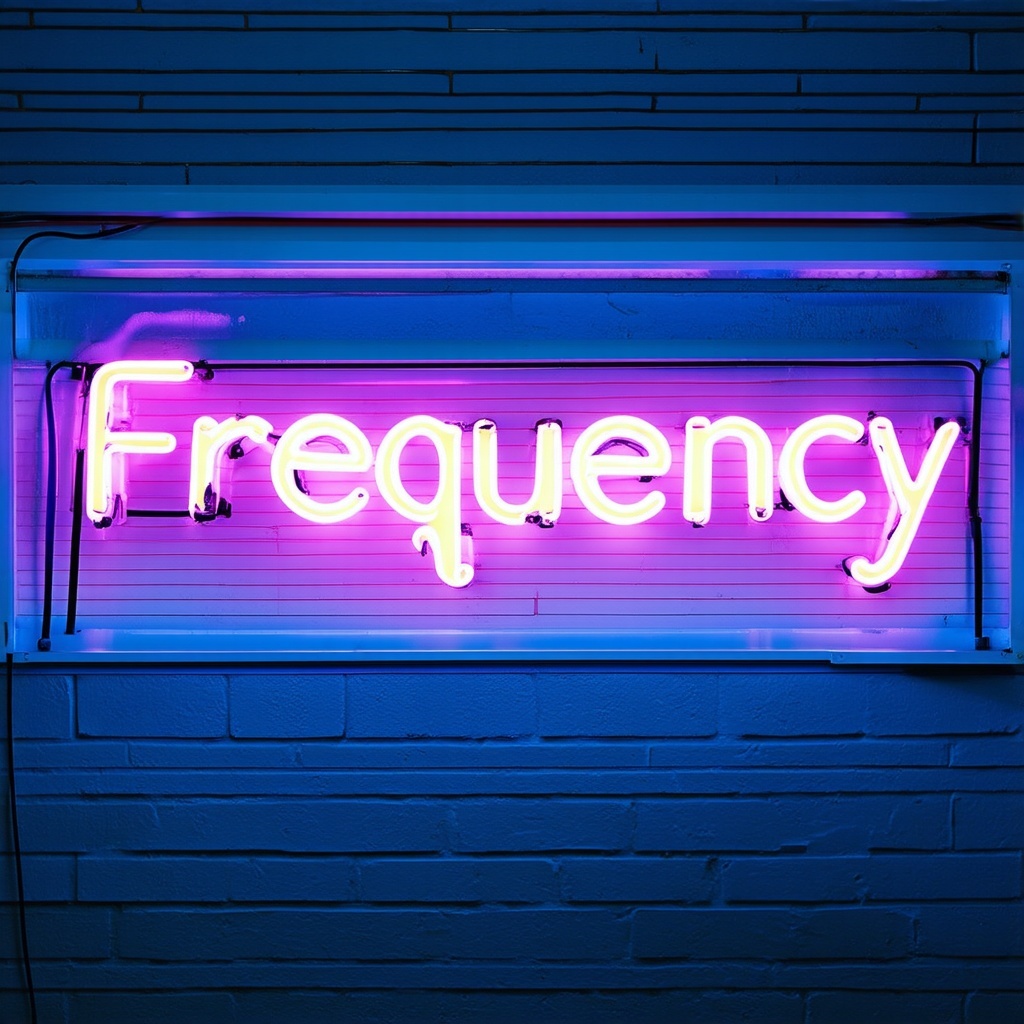}
                        };
                        \node[rotate=0, align=center, font=\normalsize, anchor=base, yshift=4pt] at (image.north) {\gls{method} (Ours) \strut};
                    \end{tikzpicture}
                \end{subfigure}
                \caption*{A neon sign that says ``Frequency''}
            \end{subfigure}
            
            \begin{adjustbox}{valign=t}
            \begin{subfigure}{0.49\textwidth}
                \begin{subfigure}{0.495\textwidth}
                    \begin{tikzpicture}
                        \node[anchor=south west, inner sep=0] (image) at (0,0) {
                            \includegraphics[width=\textwidth]{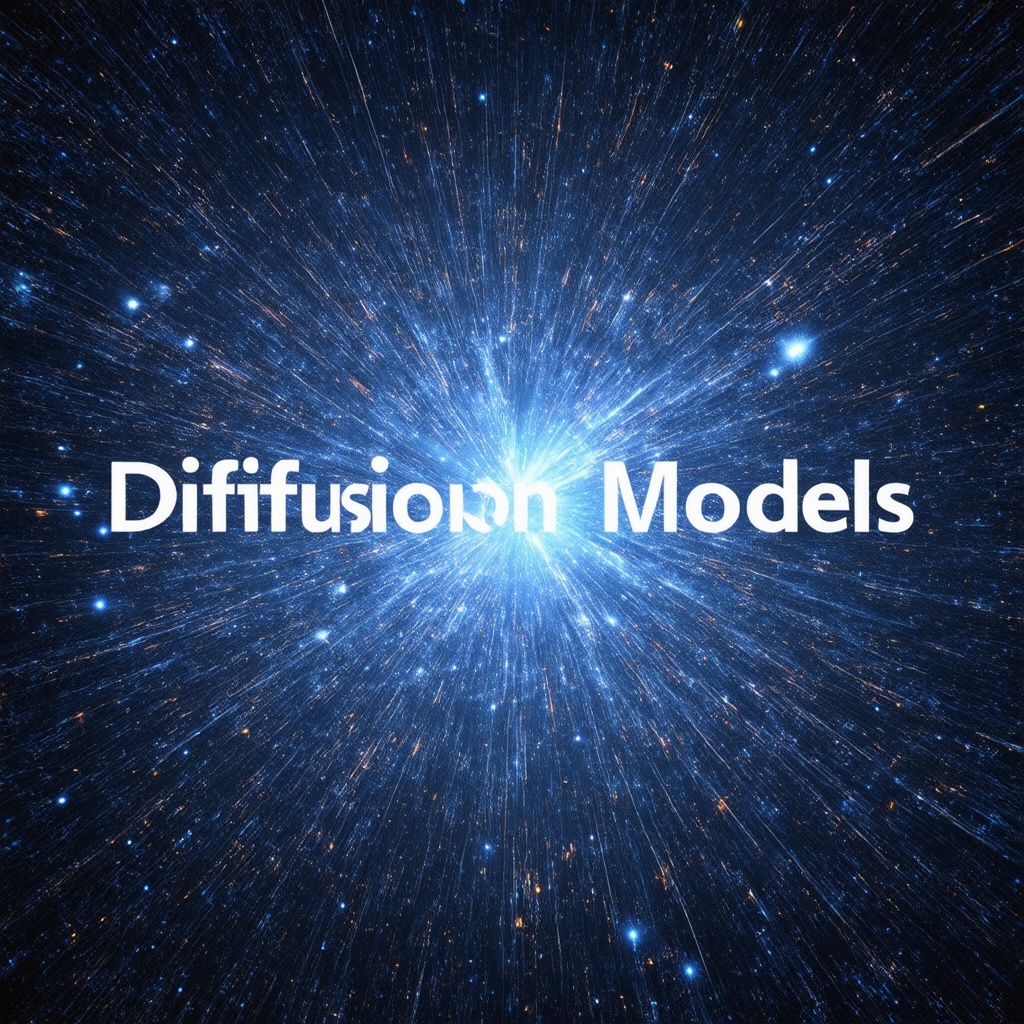}
                        };
                        \node[rotate=0, align=center, font=\normalsize, anchor=base, yshift=4pt] at (image.north) {CFG \strut};
                    \end{tikzpicture}
                \end{subfigure}
                \begin{subfigure}{0.495\textwidth}
                    \begin{tikzpicture}
                        \node[anchor=south west, inner sep=0] (image) at (0,0) {
                            \includegraphics[width=\textwidth]{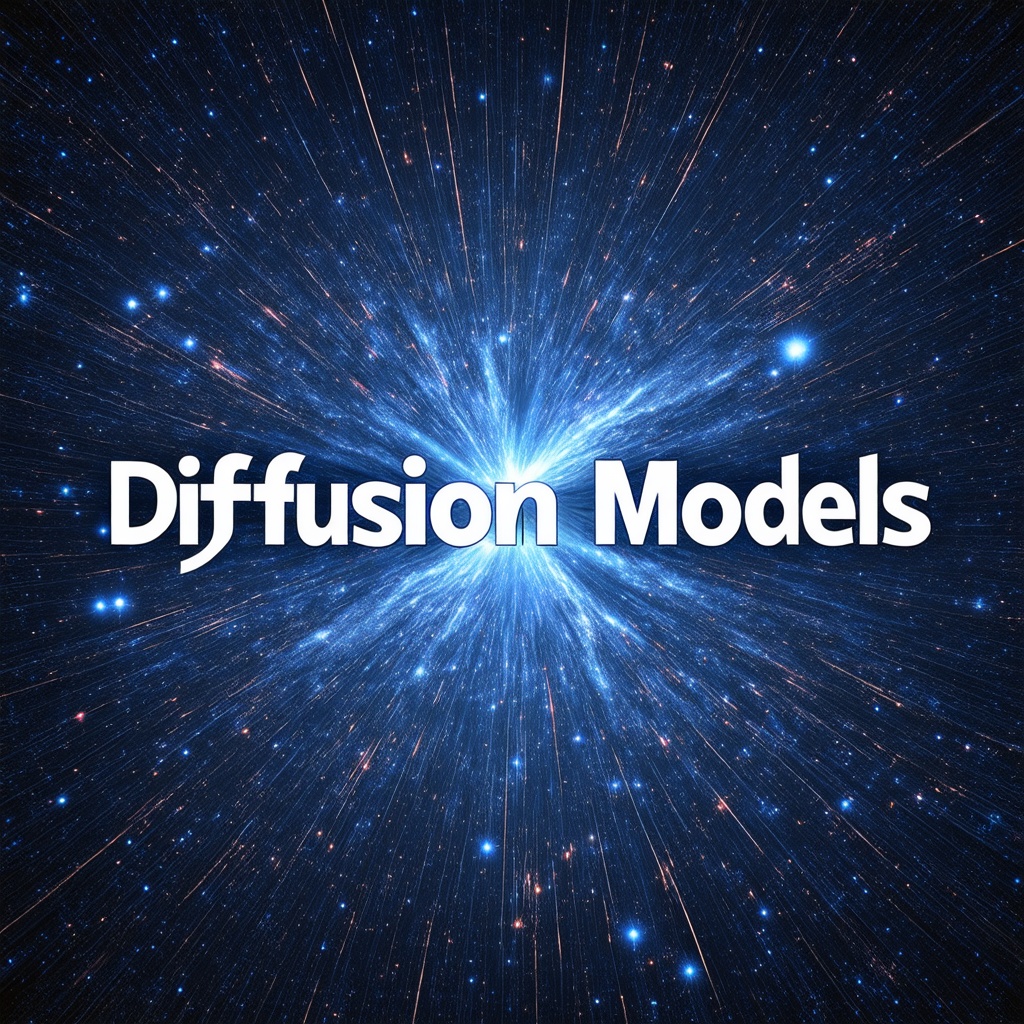}
                        };
                        \node[rotate=0, align=center, font=\normalsize, anchor=base, yshift=4pt] at (image.north) {\gls{method} (Ours) \strut};
                    \end{tikzpicture}
                \end{subfigure}
                \caption*{{A cosmic sign that says ``Diffusion Models''}}
            \end{subfigure}
        \end{adjustbox}
            \begin{adjustbox}{valign=t}
                \begin{subfigure}{0.49\textwidth}
                    \begin{subfigure}{0.495\textwidth}
                        \begin{tikzpicture}
                            \node[anchor=south west, inner sep=0] (image) at (0,0) {
                                \includegraphics[width=\textwidth]{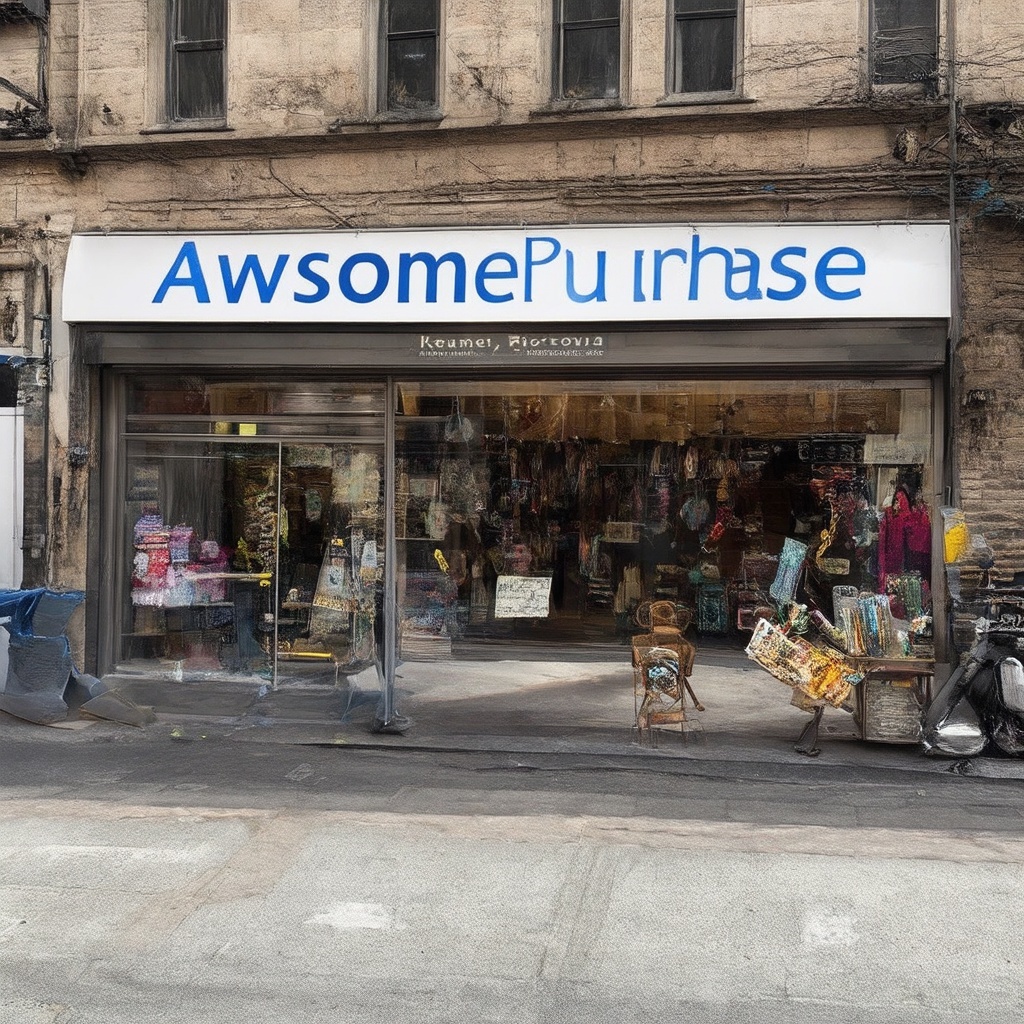}
                            };
                            \node[rotate=0, align=center, font=\normalsize, anchor=base, yshift=4pt] at (image.north) {CFG \strut};
                        \end{tikzpicture}
                    \end{subfigure}
                    \begin{subfigure}{0.495\textwidth}
                        \begin{tikzpicture}
                            \node[anchor=south west, inner sep=0] (image) at (0,0) {
                                \includegraphics[width=\textwidth]{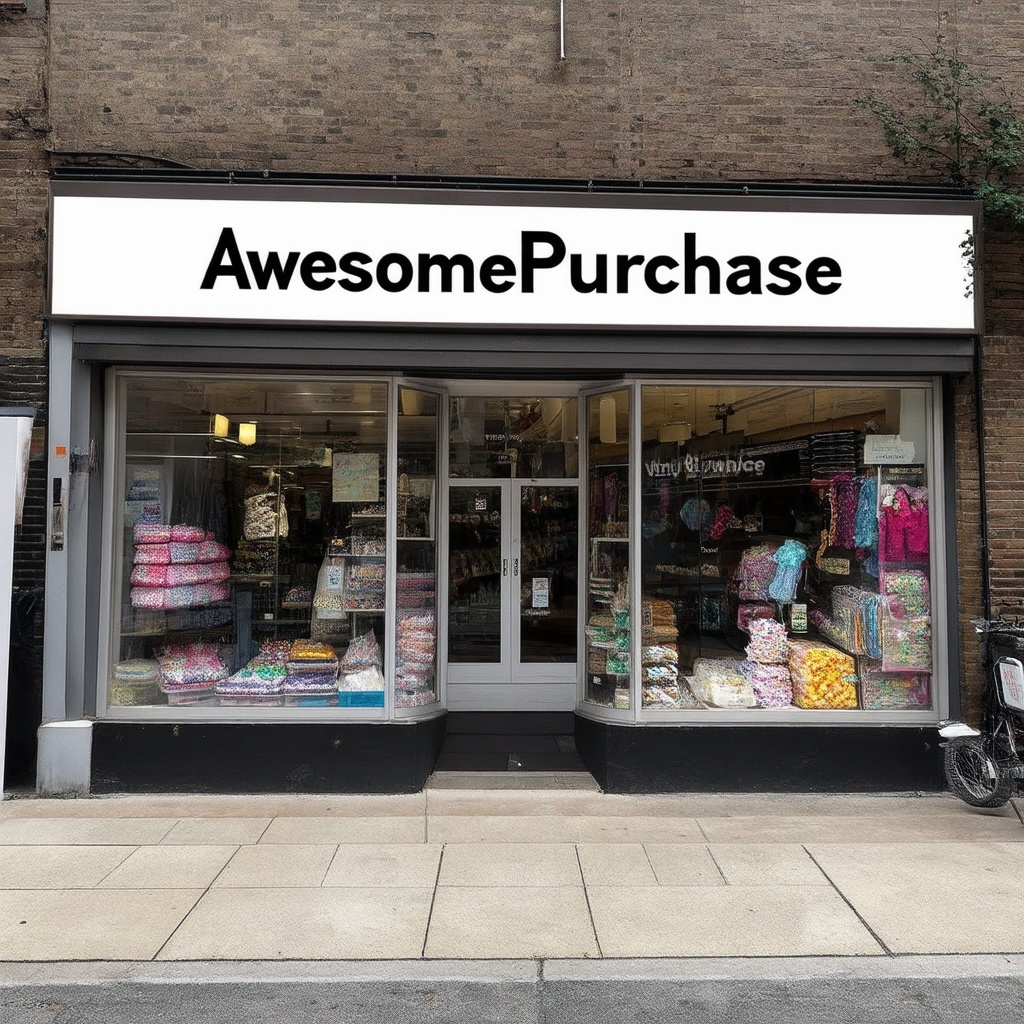}
                            };
                            \node[rotate=0, align=center, font=\normalsize, anchor=base, yshift=4pt] at (image.north) {\gls{method} (Ours) \strut};
                        \end{tikzpicture}
                    \end{subfigure}
                    \caption*{A store front with ``AwesomePurchase'' written on it}
                \end{subfigure}
            \end{adjustbox}
        \end{minipage}
        \caption{Compared with CFG, \gls{method} can improve the quality of text rendering in generated images while also using a low guidance scale to preserve the realism of generated images.}
        \label{fig:sd3-text}
    \end{figure}
}

\newcommand{\figEffectFreq}{
    \begin{figure}[t]
    \centering
    \begin{subfigure}[b]{0.32\textwidth}
        \includegraphics[width=\linewidth]{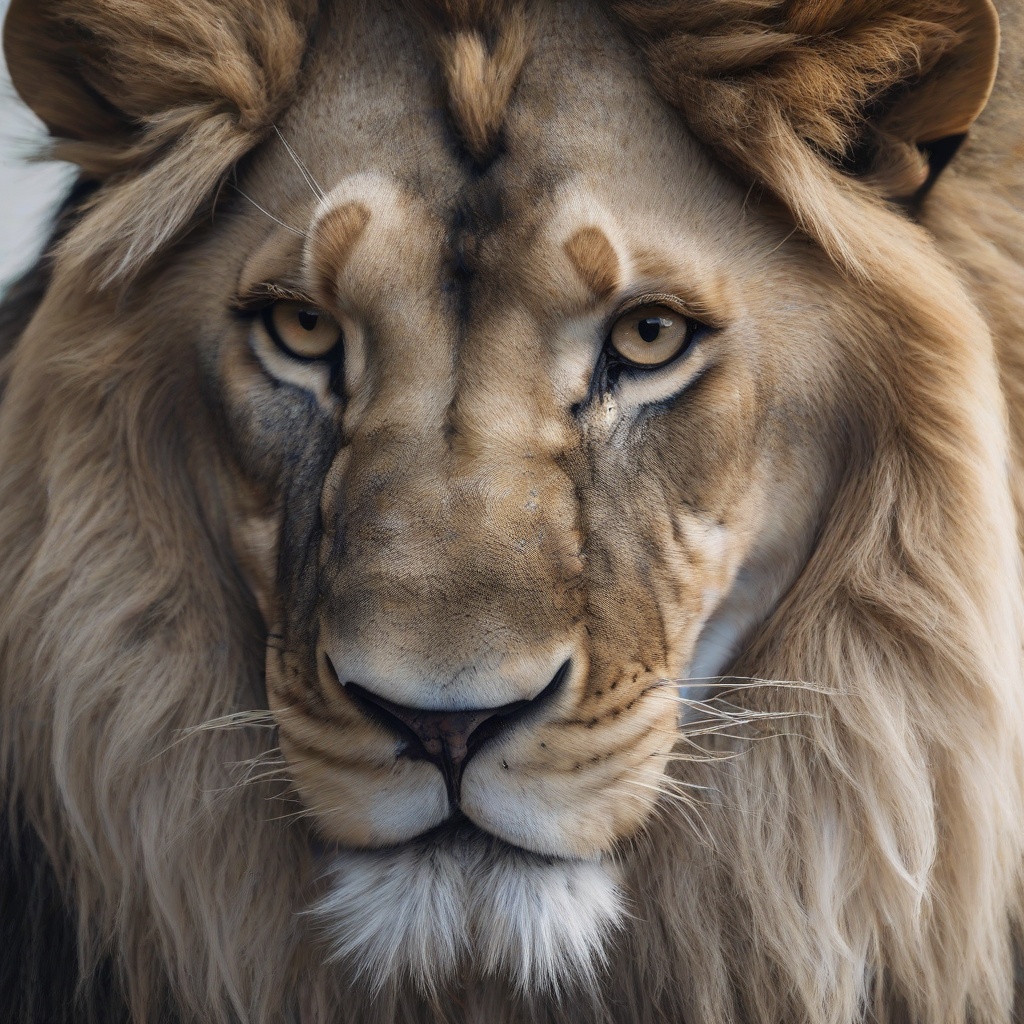}
        \caption{CFG}
    \end{subfigure}
    \hfil
    \begin{subfigure}[b]{0.32\textwidth}
        \includegraphics[width=\linewidth]{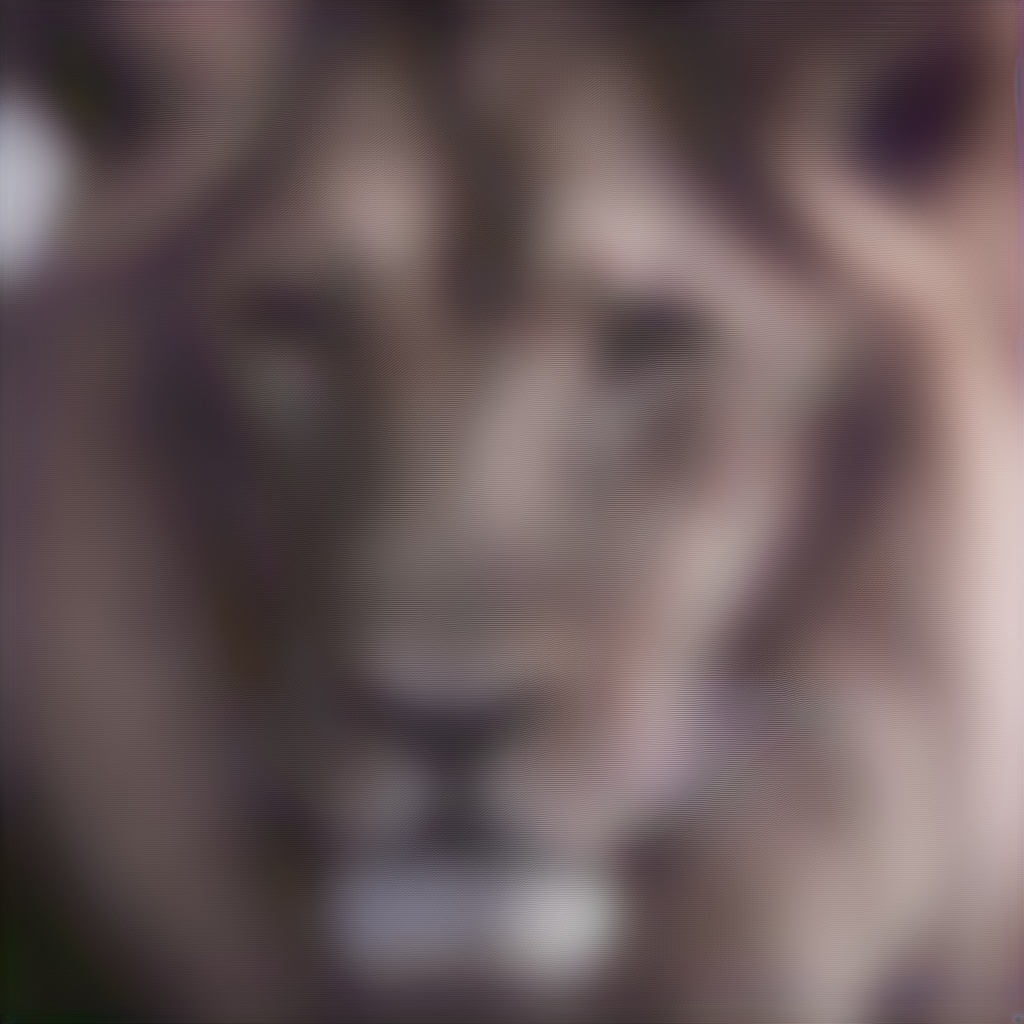}
        \caption{CFG with low frequencies}
    \end{subfigure}
    \hfil
    \begin{subfigure}[b]{0.32\textwidth}
        \includegraphics[width=\linewidth]{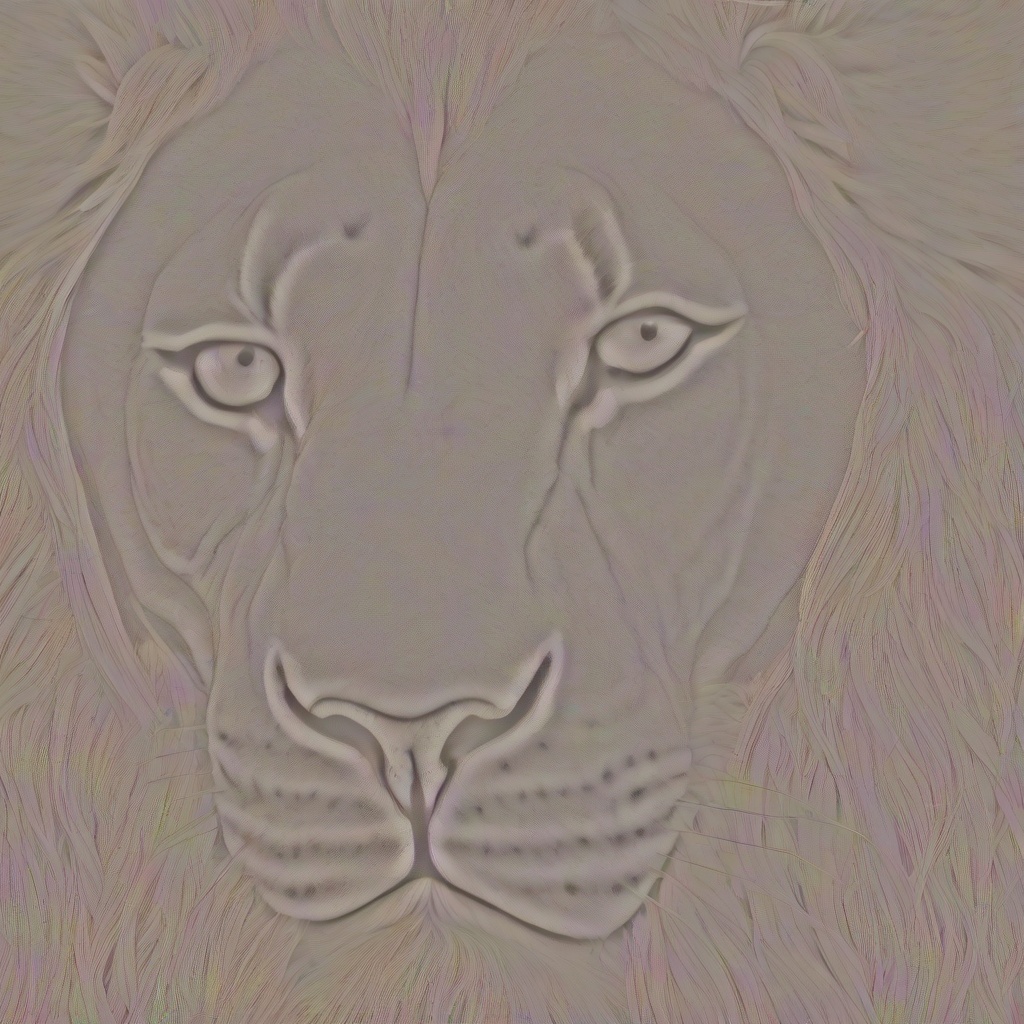}
        \caption{CFG with high frequencies}
    \end{subfigure}

    \caption{Effect of low- and high-frequency components of CFG on generated results. In this experiment, we explicitly set either low or high frequencies to zero to isolate the effect of the other component. Note that low frequencies determine the overall structure of the output, while high frequencies contribute to details.}
    \label{fig:freq-effect}
\end{figure}
}

\newcommand{\figAPG}{
    \begin{figure}[t]
    \centering
        \begin{subfigure}{0.495\textwidth}
            \begin{subfigure}{0.495\textwidth}
                    \begin{tikzpicture}
                        \node[anchor=south west, inner sep=0] (image) at (0,0) {
                            \includegraphics[width=\textwidth]{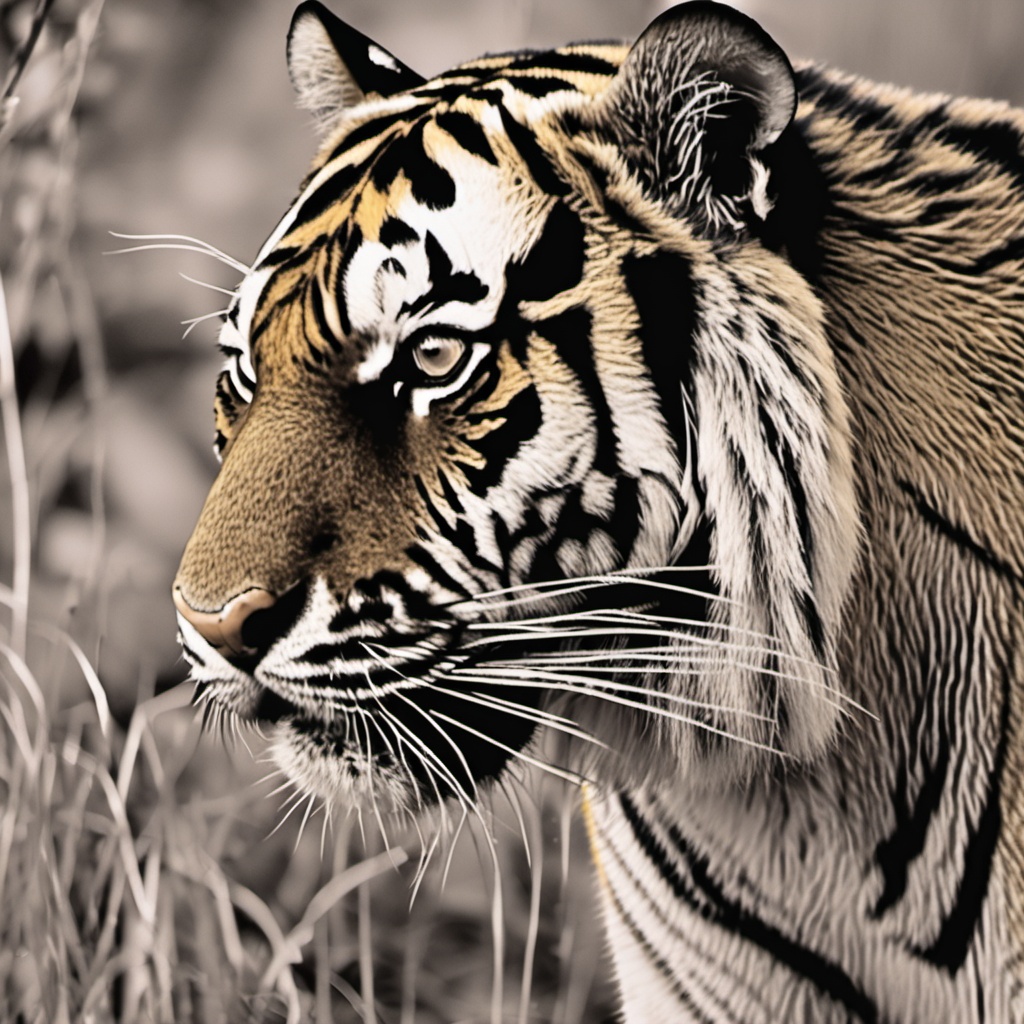}
                        };
                        \node[rotate=0, align=center, font=\normalsize, anchor=base, yshift=4pt] at (image.north) {\gls{method} \strut};
                    \end{tikzpicture}
                    \end{subfigure}
                    \begin{subfigure}{0.495\textwidth}
                    \begin{tikzpicture}
                        \node[anchor=south west, inner sep=0] (image) at (0,0) {
                            \includegraphics[width=\textwidth]{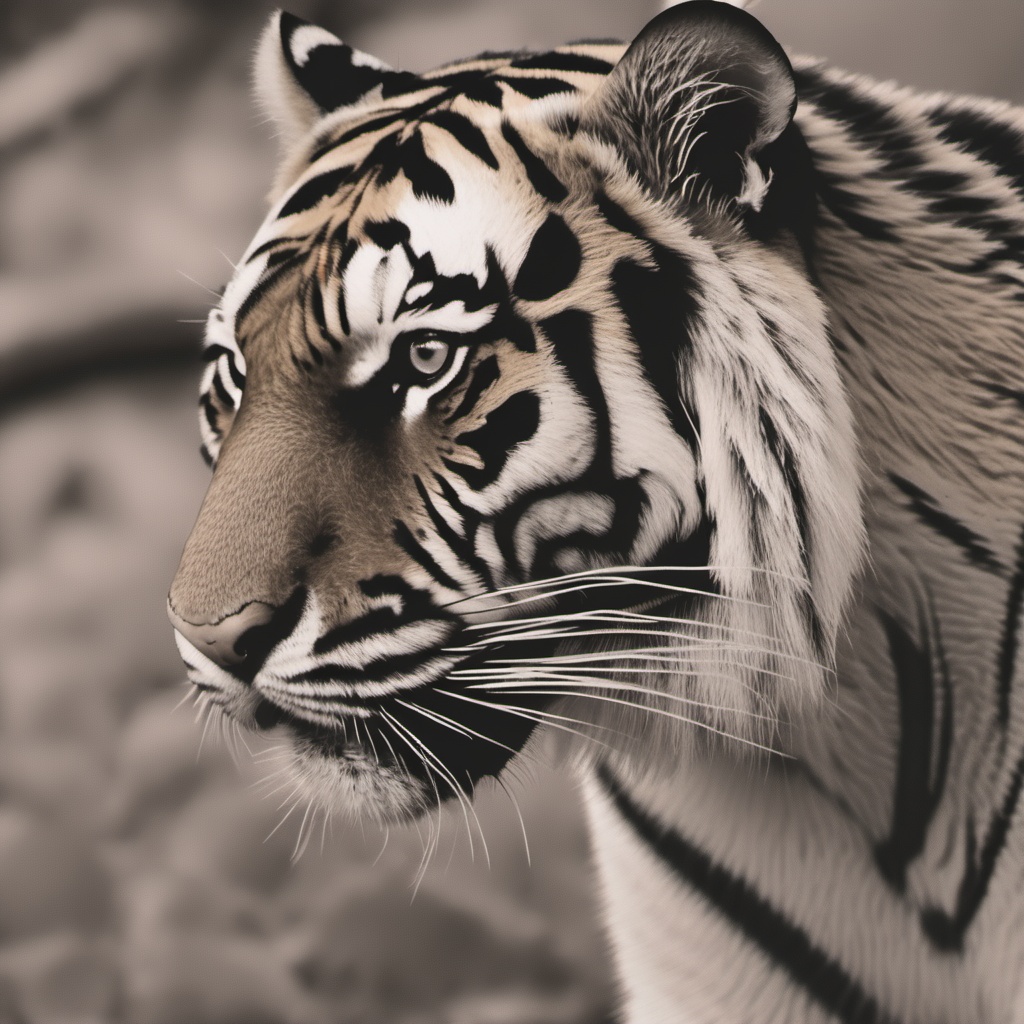}
                        };
                        \node[rotate=0, align=center, font=\normalsize, anchor=base, yshift=4pt] at (image.north) {\gls{method} + \gls{apg} \strut};
                    \end{tikzpicture}
                    \end{subfigure}
                \caption*{Image of a tiger}
                \end{subfigure}
                \hfil
            \begin{subfigure}{0.495\textwidth}
            \begin{subfigure}{0.495\textwidth}
                    \begin{tikzpicture}
                        \node[anchor=south west, inner sep=0] (image) at (0,0) {
                            \includegraphics[width=\textwidth]{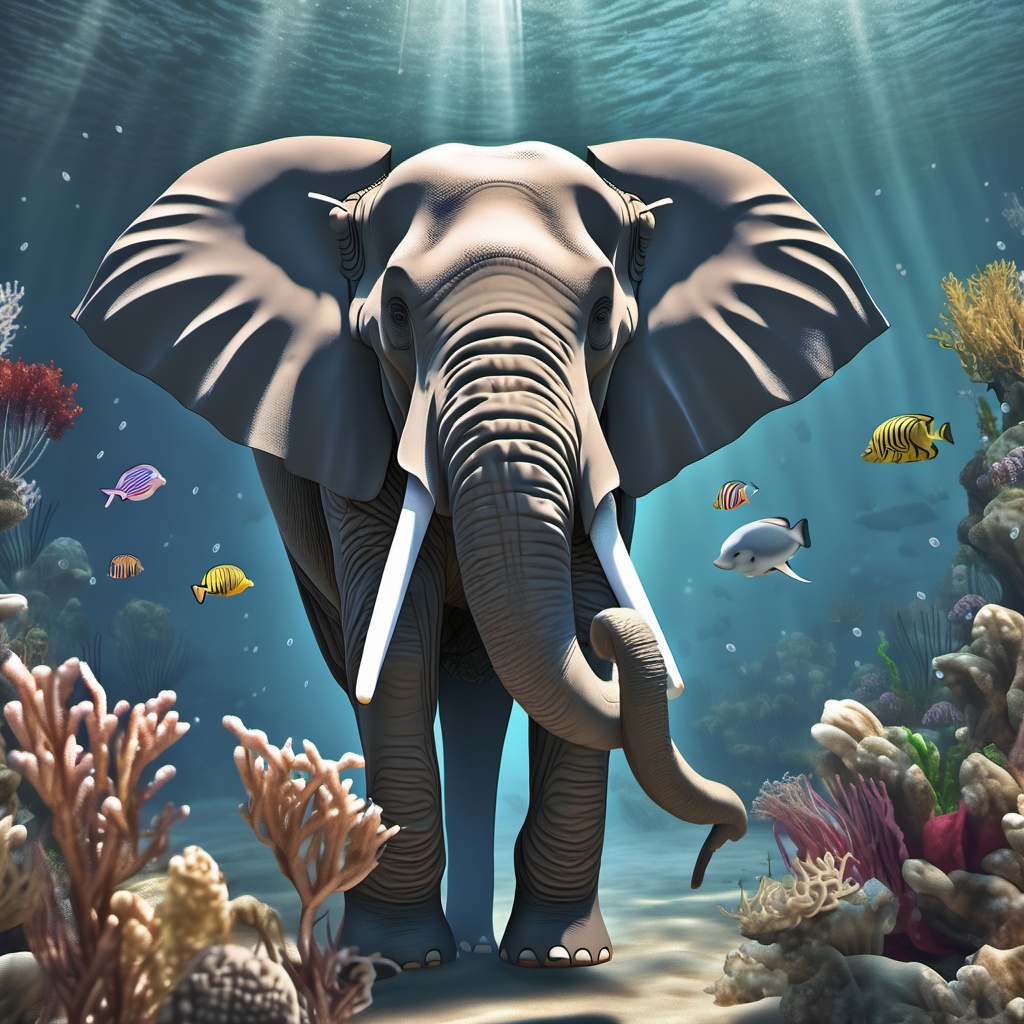}
                        };
                        \node[rotate=0, align=center, font=\normalsize, anchor=base, yshift=4pt] at (image.north) {\gls{method} \strut};
                    \end{tikzpicture}
                    \end{subfigure}
                    \begin{subfigure}{0.495\textwidth}
                    \begin{tikzpicture}
                        \node[anchor=south west, inner sep=0] (image) at (0,0) {
                            \includegraphics[width=\textwidth]{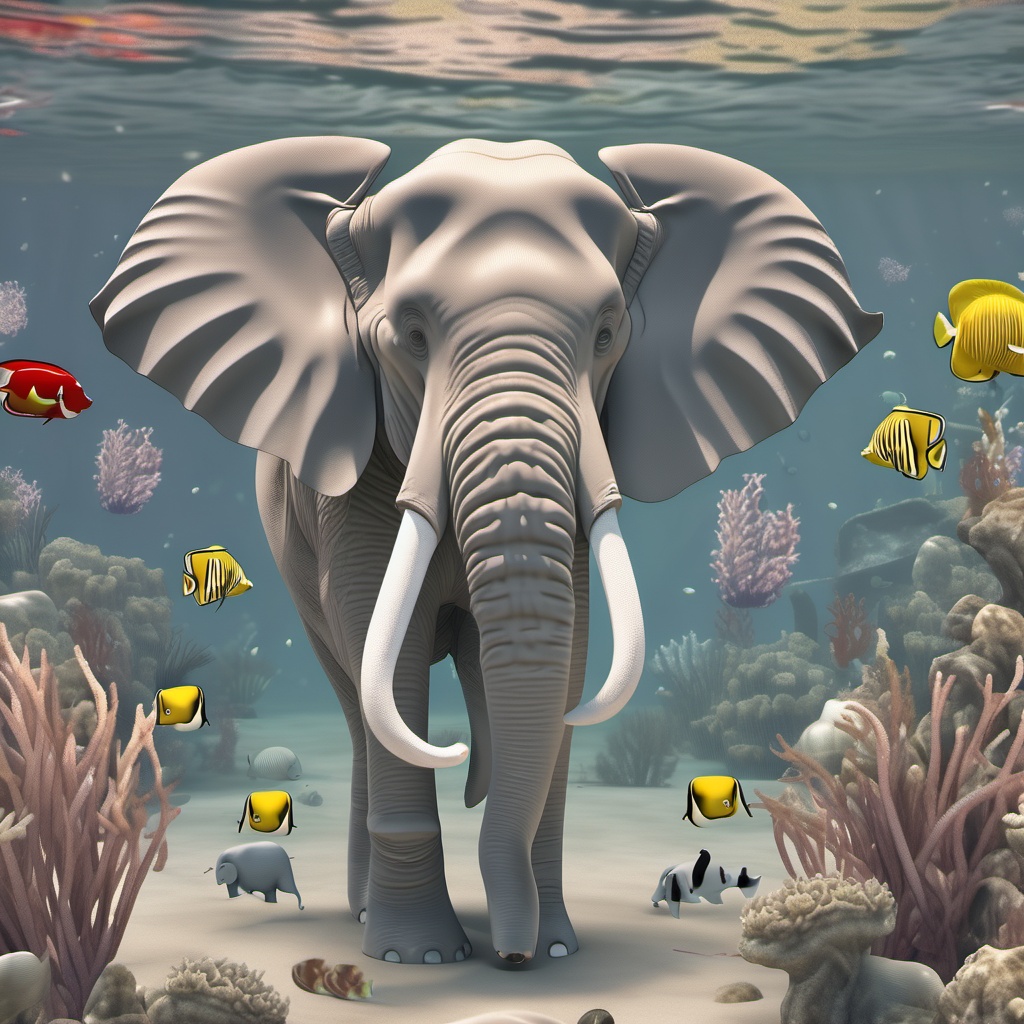}
                        };
                        \node[rotate=0, align=center, font=\normalsize, anchor=base, yshift=4pt] at (image.north) {\gls{method} + \gls{apg} \strut};
                    \end{tikzpicture}
                    \end{subfigure}
                \caption*{Elephant under the see, realistic, 4k}
                \end{subfigure}
    \caption{Combining \gls{method} with \gls{apg}. The \gls{apg} update rule can be seamlessly integrated with \gls{method} to produce generations with more realistic colors. For this experiment, we set $\wcfglow = 5$ and $\wcfghigh = 15$.}
    \label{fig:apg}
\end{figure}
}

\newcommand{\figDiTAppendix}{
    \begin{figure}[t]
        \centering
        \begin{minipage}{\textwidth}
            \begin{subfigure}{0.49\textwidth}
                \begin{tikzpicture}
                    \node[anchor=south west, inner sep=0] (image) at (0,0) {
                        \includegraphics[width=\textwidth]{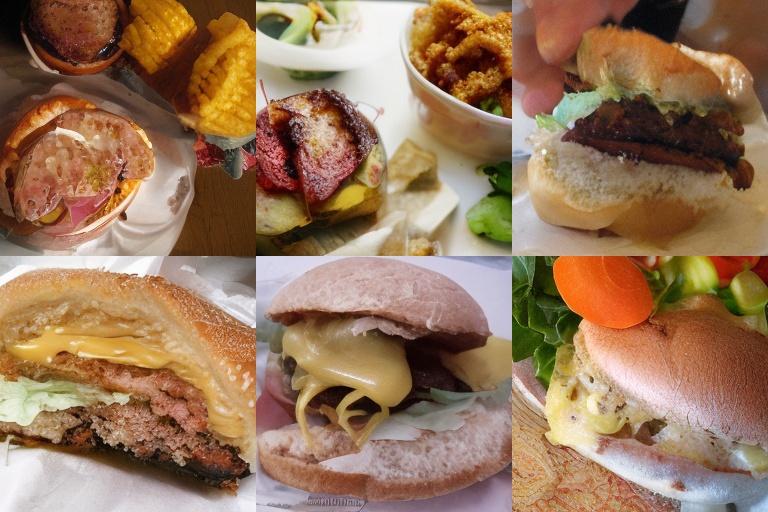}
                    };
                    \node[rotate=0, align=center, font=\normalsize, anchor=base, yshift=4pt] at (image.north) {CFG \strut};
                \end{tikzpicture}
            \end{subfigure}
            \hfill
            \begin{subfigure}{0.49\textwidth}
                \begin{tikzpicture}
                    \node[anchor=south west, inner sep=0] (image) at (0,0) {
                        \includegraphics[width=\textwidth]{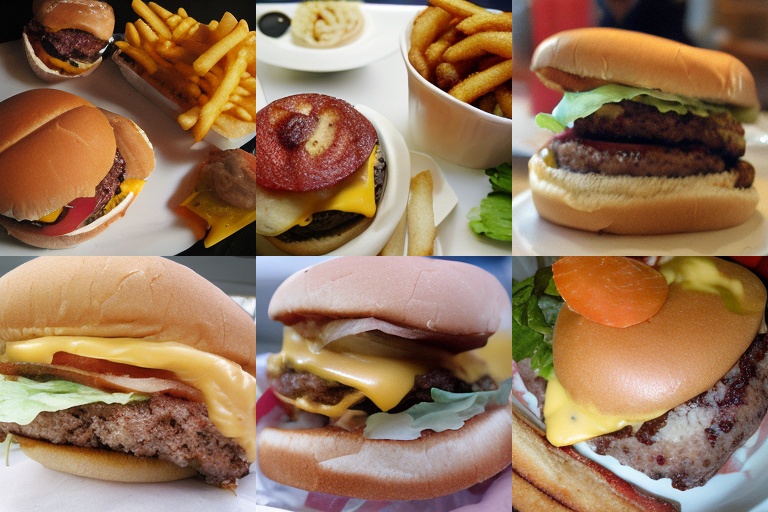}
                    };
                    \node[rotate=0, align=center, font=\normalsize, anchor=base, yshift=4pt] at (image.north) {\gls{method} (Ours) \strut};
                \end{tikzpicture}
            \end{subfigure}
            \begin{subfigure}{0.49\textwidth}
                \begin{tikzpicture}
                    \node[anchor=south west, inner sep=0] (image) at (0,0) {
                        \includegraphics[width=\textwidth]{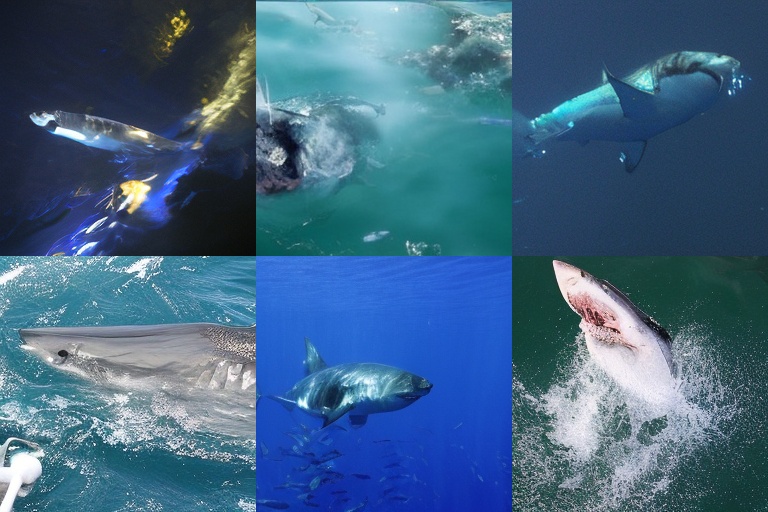}
                    };
                    \node[rotate=0, align=center, font=\normalsize, anchor=base, yshift=4pt] at (image.north) {CFG \strut};
                \end{tikzpicture}
            \end{subfigure}
            \hfill
            \begin{subfigure}{0.49\textwidth}
                \begin{tikzpicture}
                    \node[anchor=south west, inner sep=0] (image) at (0,0) {
                        \includegraphics[width=\textwidth]{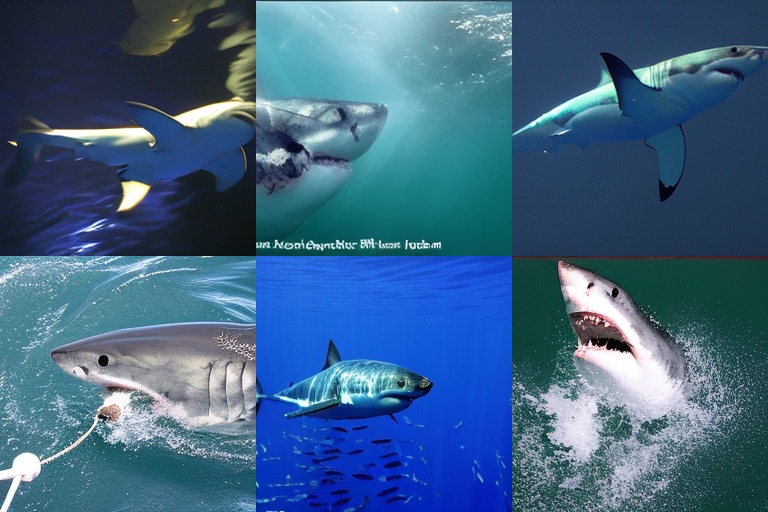}
                    };
                    \node[rotate=0, align=center, font=\normalsize, anchor=base, yshift=4pt] at (image.north) {\gls{method} (Ours) \strut};
                \end{tikzpicture}
            \end{subfigure}
            \begin{subfigure}{0.49\textwidth}
                \begin{tikzpicture}
                    \node[anchor=south west, inner sep=0] (image) at (0,0) {
                        \includegraphics[width=\textwidth]{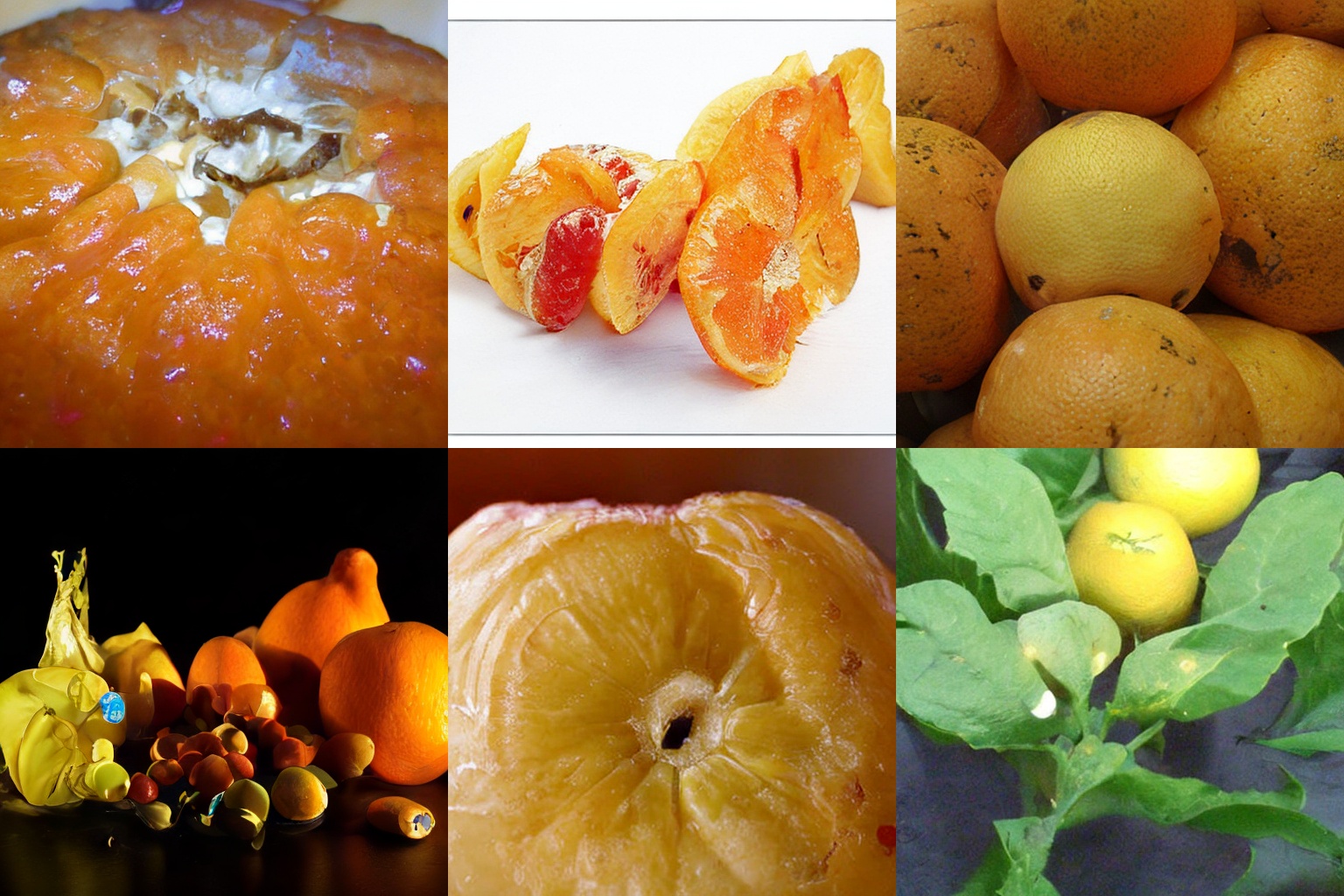}
                    };
                    \node[rotate=0, align=center, font=\normalsize, anchor=base, yshift=4pt] at (image.north) {CFG \strut};
                \end{tikzpicture}
            \end{subfigure}
            \hfill
            \begin{subfigure}{0.49\textwidth}
                \begin{tikzpicture}
                    \node[anchor=south west, inner sep=0] (image) at (0,0) {
                        \includegraphics[width=\textwidth]{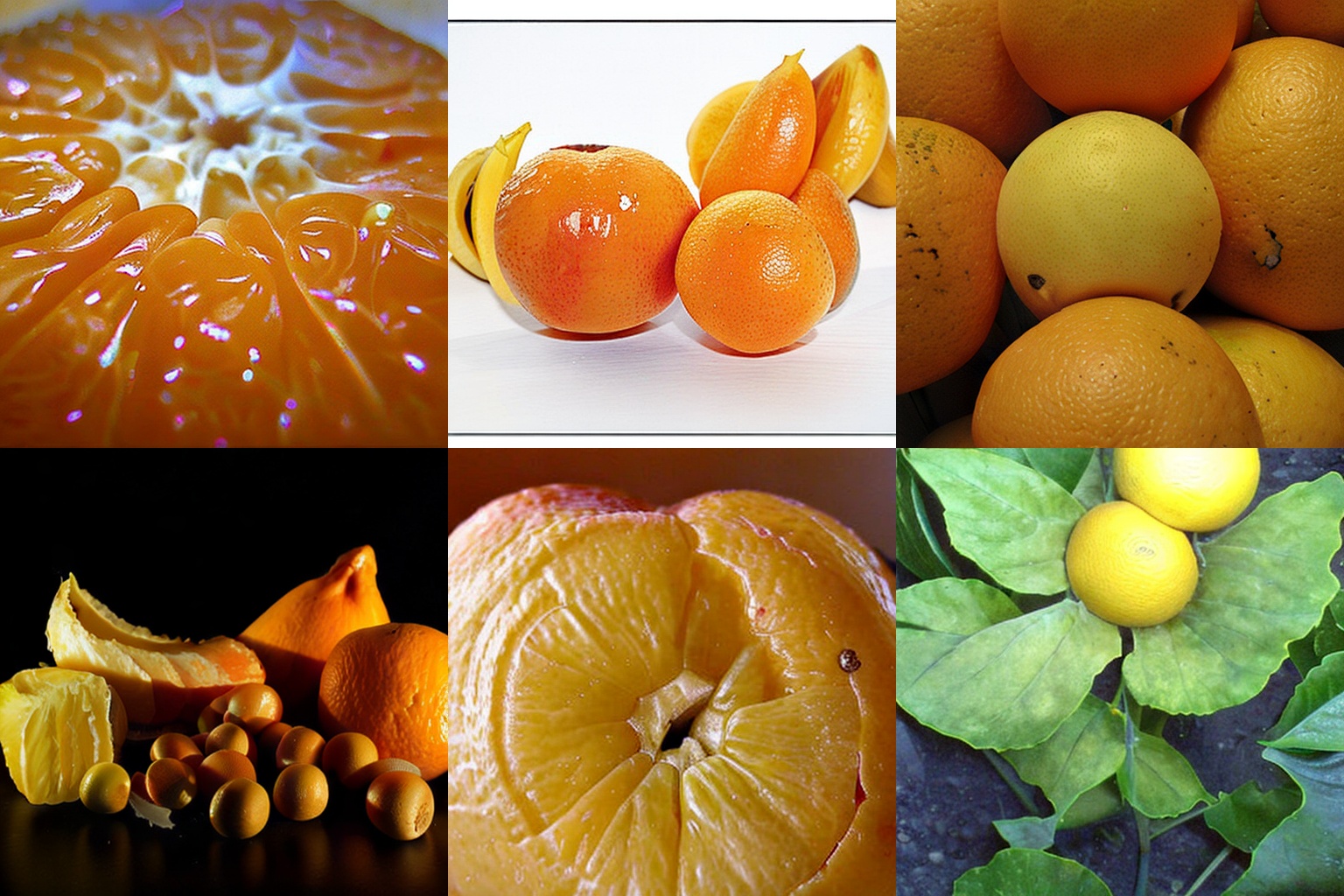}
                    };
                    \node[rotate=0, align=center, font=\normalsize, anchor=base, yshift=4pt] at (image.north) {\gls{method} (Ours) \strut};
                \end{tikzpicture}
            \end{subfigure}
            \begin{subfigure}{0.49\textwidth}
                \begin{tikzpicture}
                    \node[anchor=south west, inner sep=0] (image) at (0,0) {
                        \includegraphics[width=\textwidth]{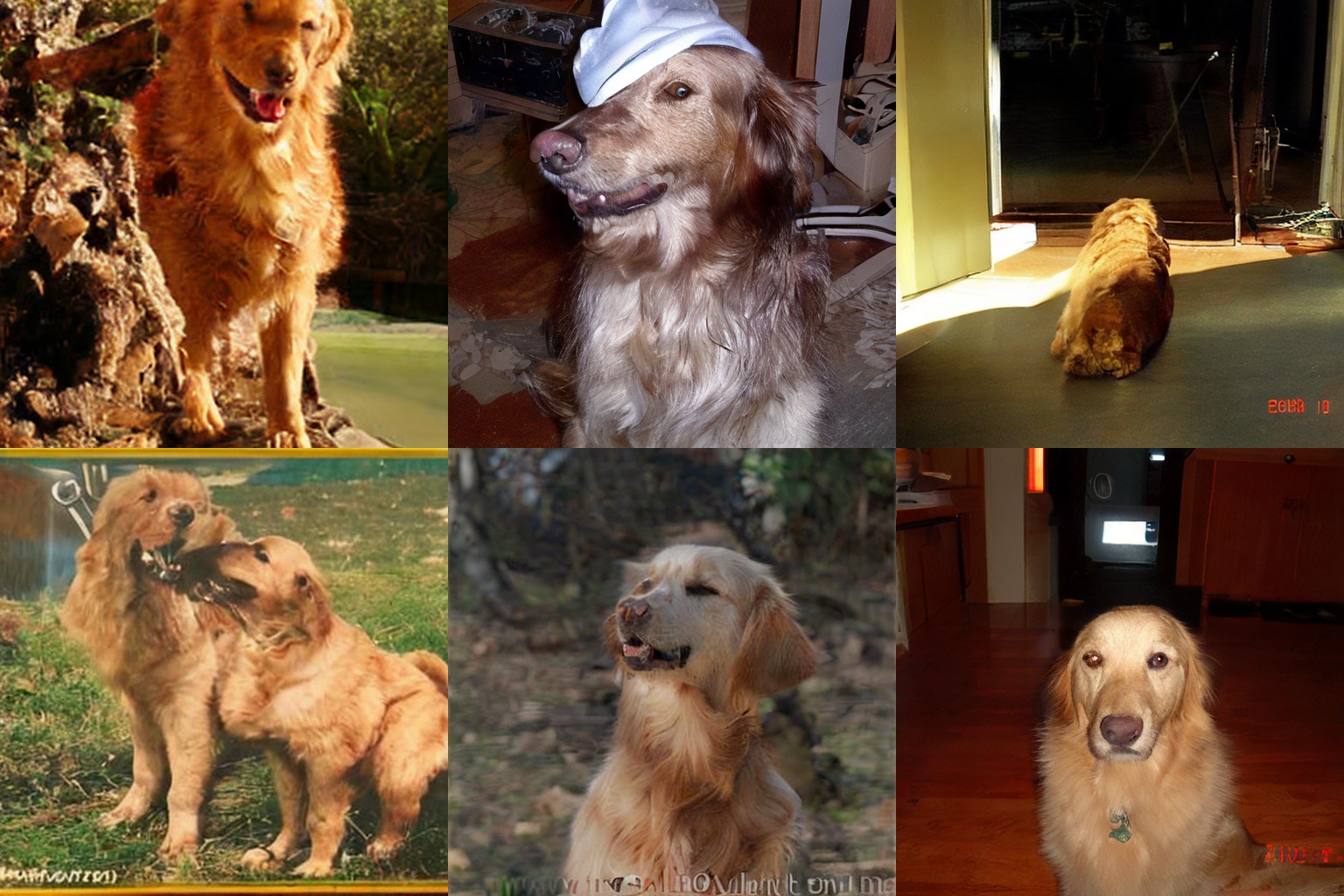}
                    };
                    \node[rotate=0, align=center, font=\normalsize, anchor=base, yshift=4pt] at (image.north) {CFG \strut};
                \end{tikzpicture}
            \end{subfigure}
            \hfill
            \begin{subfigure}{0.49\textwidth}
                \begin{tikzpicture}
                    \node[anchor=south west, inner sep=0] (image) at (0,0) {
                        \includegraphics[width=\textwidth]{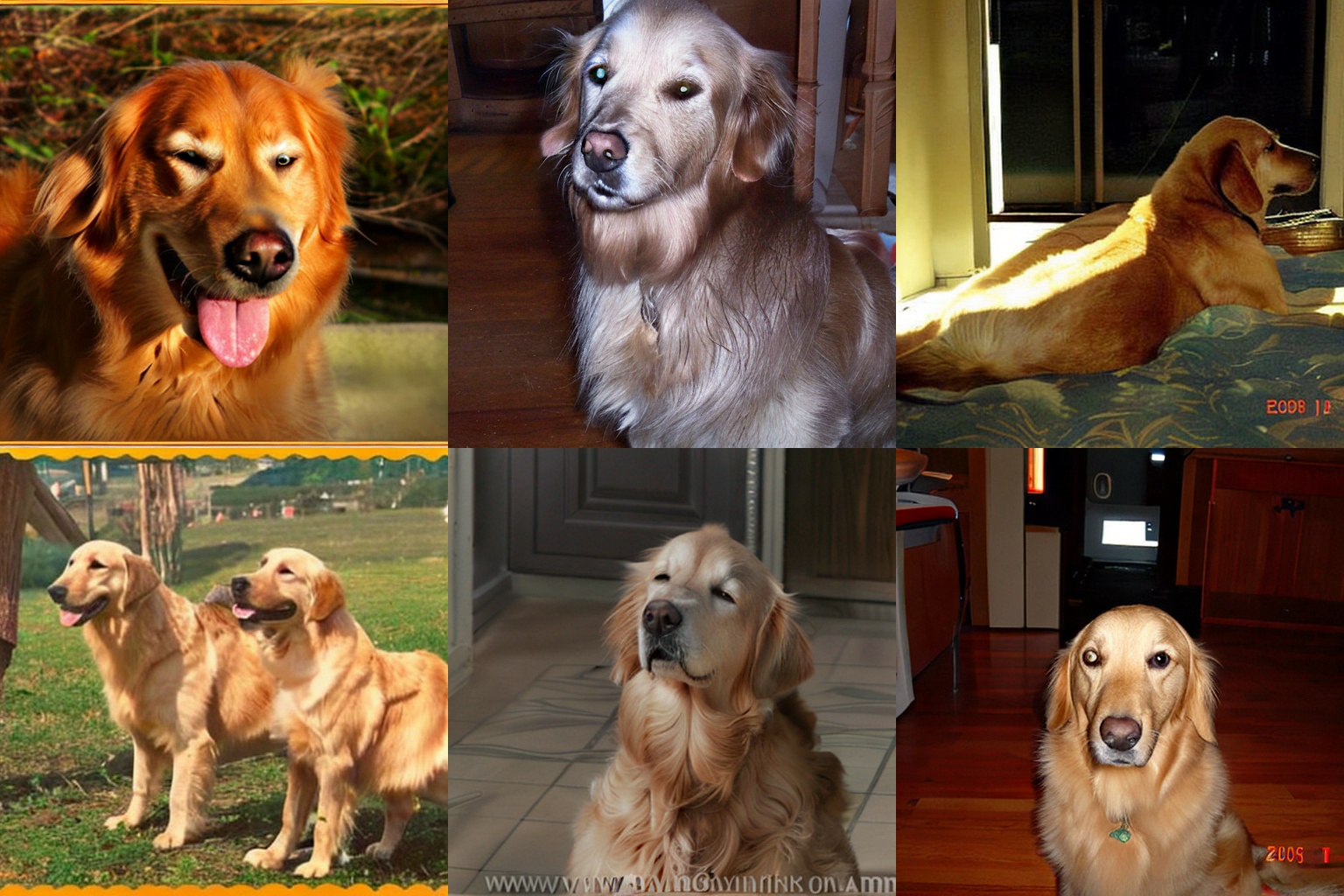}
                    };
                    \node[rotate=0, align=center, font=\normalsize, anchor=base, yshift=4pt] at (image.north) {\gls{method} (Ours) \strut};
                \end{tikzpicture}
            \end{subfigure}
        \end{minipage}
        \caption{More visual results comparing \gls{method} and CFG using class-conditional ImageNet models.}
        \label{fig:imagenet-appendix}
    \end{figure}
}

\newcommand{\figSDXLAppendix}{
    \begin{figure}[t!]
        \centering
        \begin{minipage}{\textwidth}
            \begin{subfigure}{0.49\textwidth}
                \begin{subfigure}{0.495\textwidth}
                    \begin{tikzpicture}
                        \node[anchor=south west, inner sep=0] (image) at (0,0) {
                            \includegraphics[width=\textwidth]{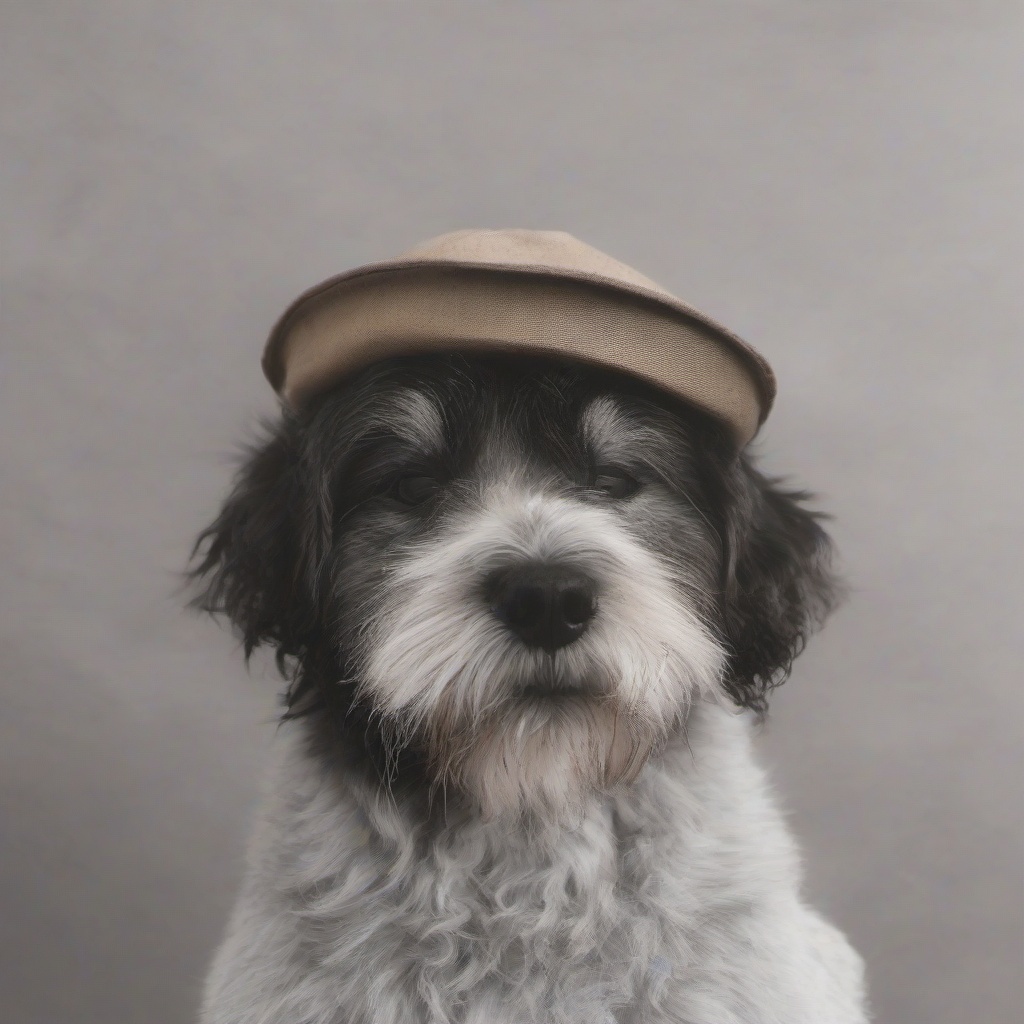}
                        };
                        \node[rotate=0, align=center, font=\normalsize, anchor=base, yshift=4pt] at (image.north) {CFG \strut};
                    \end{tikzpicture}
                \end{subfigure}
                \begin{subfigure}{0.495\textwidth}
                    \begin{tikzpicture}
                        \node[anchor=south west, inner sep=0] (image) at (0,0) {
                            \includegraphics[width=\textwidth]{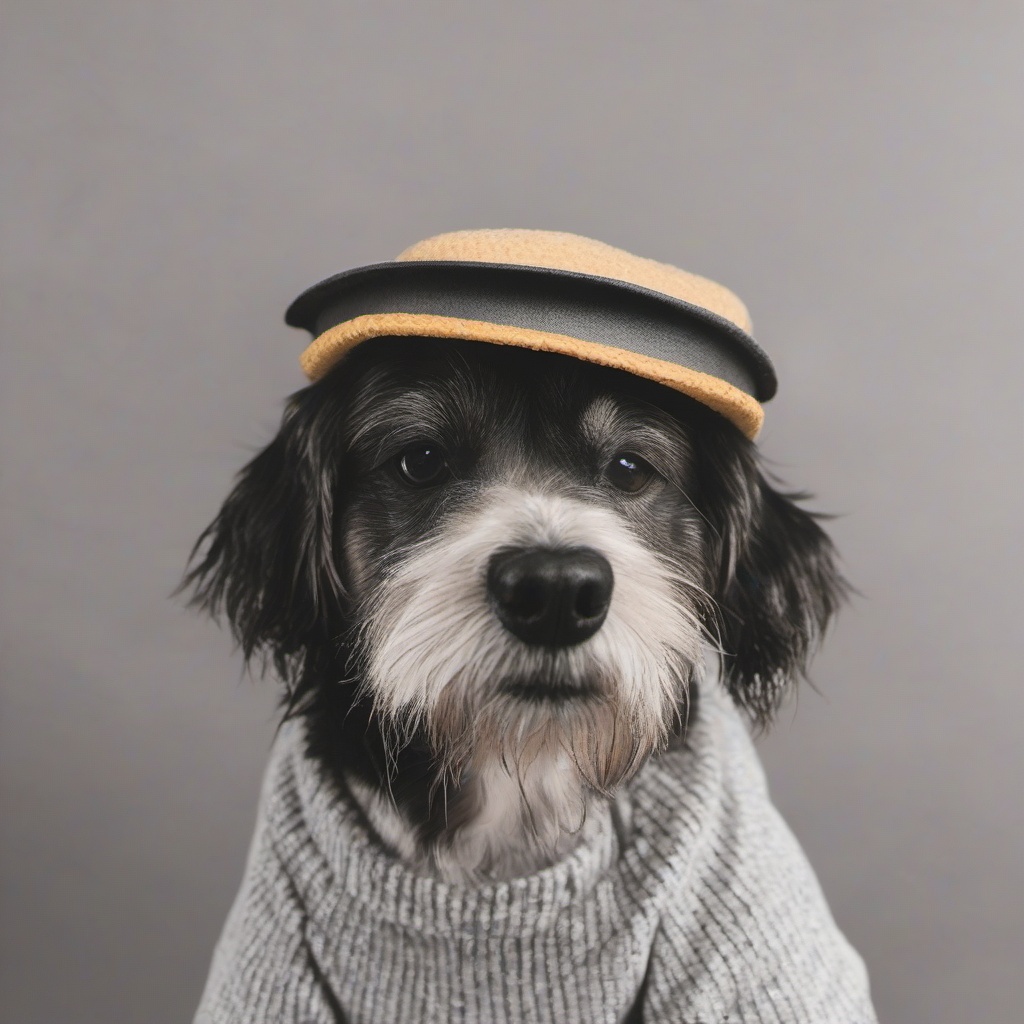}
                        };
                        \node[rotate=0, align=center, font=\normalsize, anchor=base, yshift=4pt] at (image.north) {\gls{method} (Ours) \strut};
                    \end{tikzpicture}
                \end{subfigure}
                \caption*{A dog wearing a hat}
            \end{subfigure}
            \begin{subfigure}{0.49\textwidth}
                \begin{subfigure}{0.495\textwidth}
                    \begin{tikzpicture}
                        \node[anchor=south west, inner sep=0] (image) at (0,0) {
                            \includegraphics[width=\textwidth]{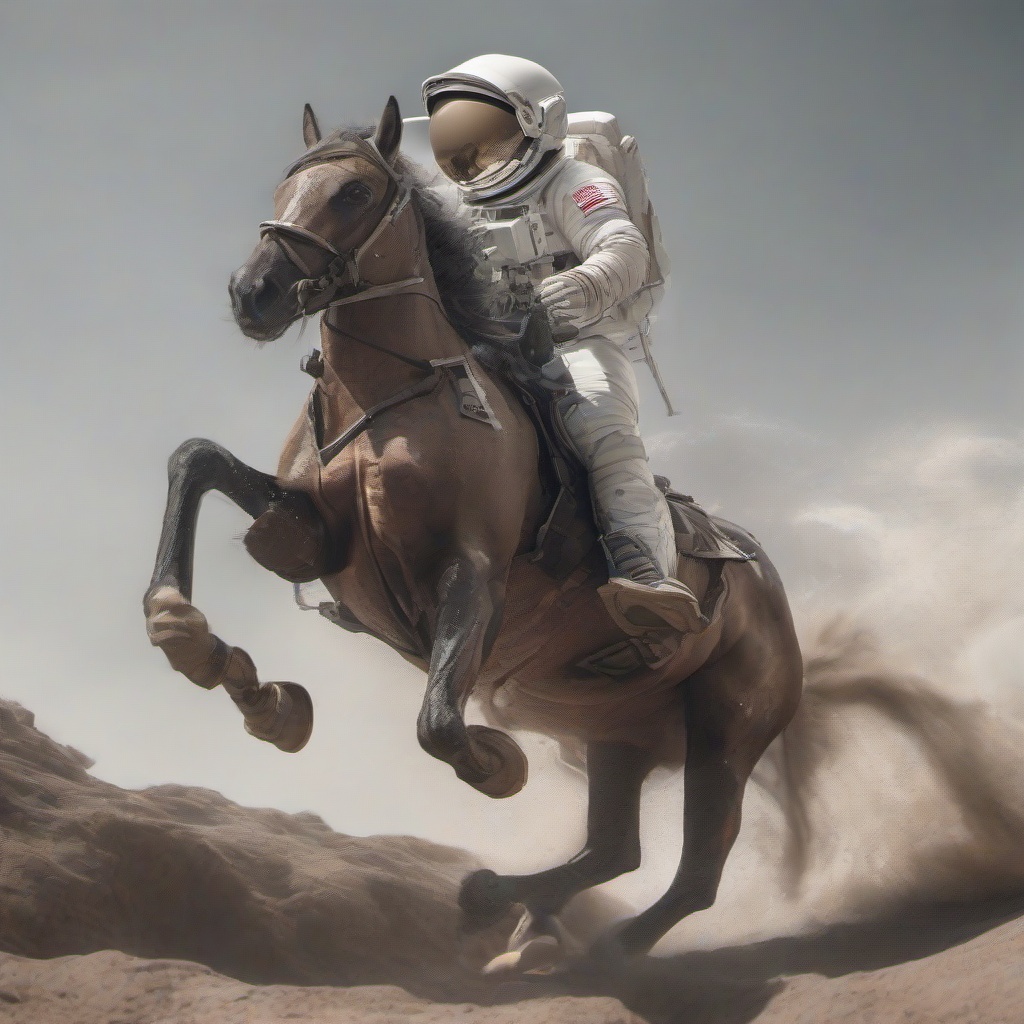}
                        };
                        \node[rotate=0, align=center, font=\normalsize, anchor=base, yshift=4pt] at (image.north) {CFG \strut};
                    \end{tikzpicture}
                \end{subfigure}
                \begin{subfigure}{0.495\textwidth}
                    \begin{tikzpicture}
                        \node[anchor=south west, inner sep=0] (image) at (0,0) {
                            \includegraphics[width=\textwidth]{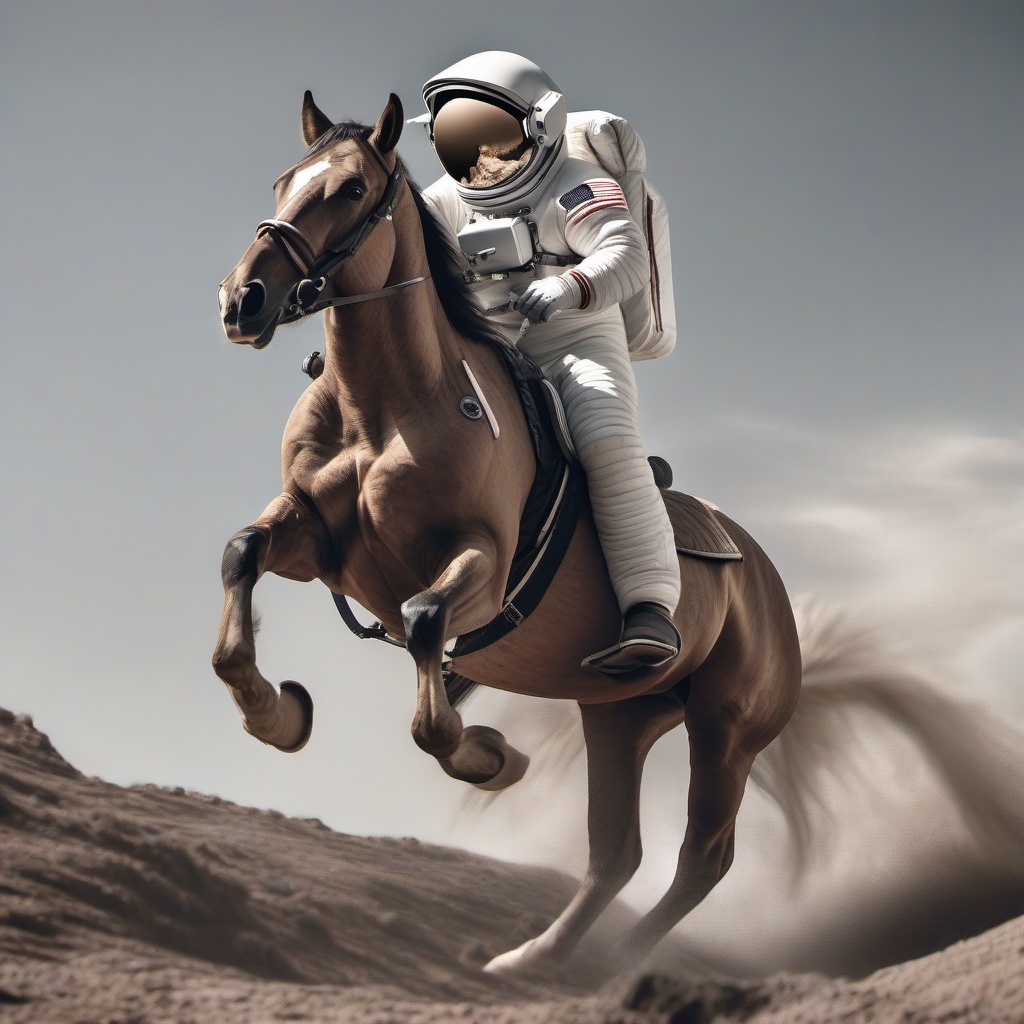}
                        };
                        \node[rotate=0, align=center, font=\normalsize, anchor=base, yshift=4pt] at (image.north) {\gls{method} (Ours) \strut};
                    \end{tikzpicture}
                \end{subfigure}
                \caption*{An astronaut riding a horse, highly realistic dslr photo}
            \end{subfigure}
            
            \begin{adjustbox}{valign=t}
            \begin{subfigure}{0.49\textwidth}
                \begin{subfigure}{0.495\textwidth}
                    \begin{tikzpicture}
                        \node[anchor=south west, inner sep=0] (image) at (0,0) {
                            \includegraphics[width=\textwidth]{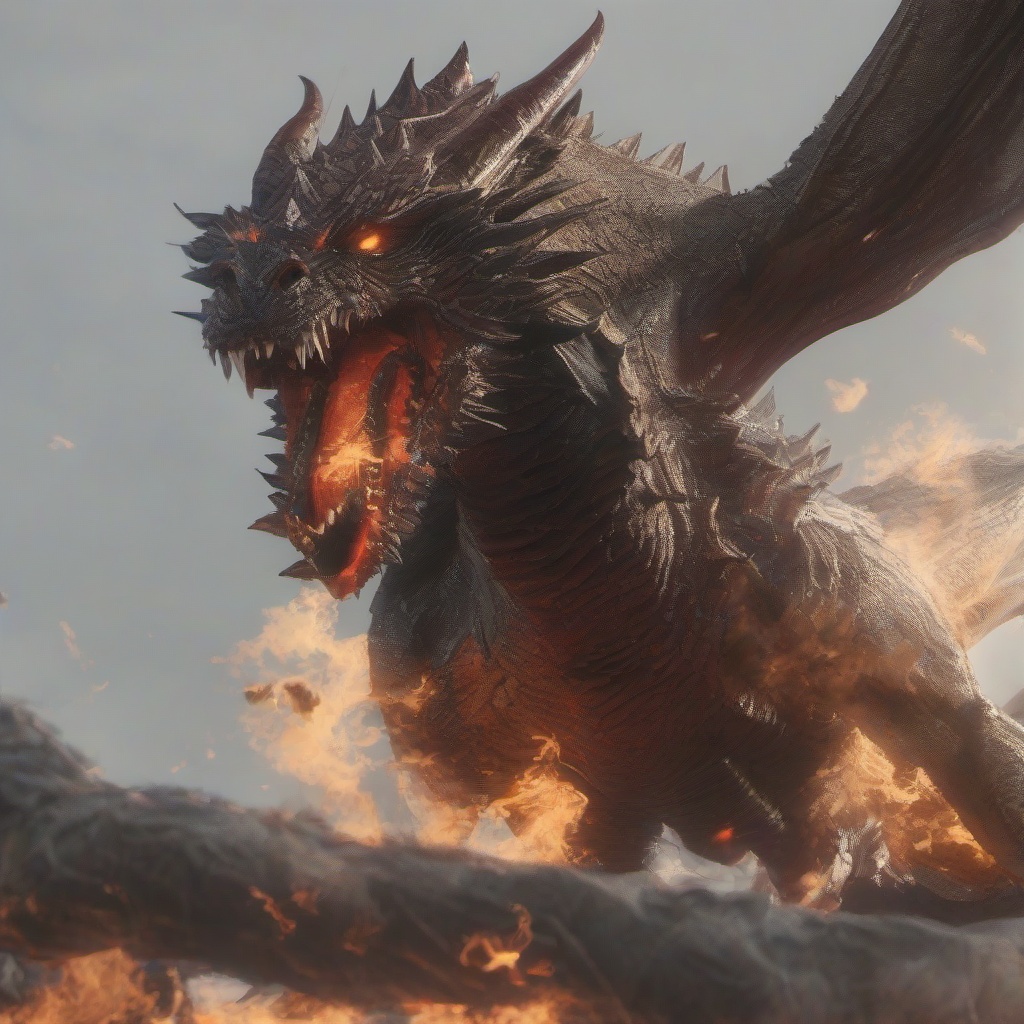}
                        };
                        \node[rotate=0, align=center, font=\normalsize, anchor=base, yshift=4pt] at (image.north) {CFG \strut};
                    \end{tikzpicture}
                \end{subfigure}
                \begin{subfigure}{0.495\textwidth}
                    \begin{tikzpicture}
                        \node[anchor=south west, inner sep=0] (image) at (0,0) {
                            \includegraphics[width=\textwidth]{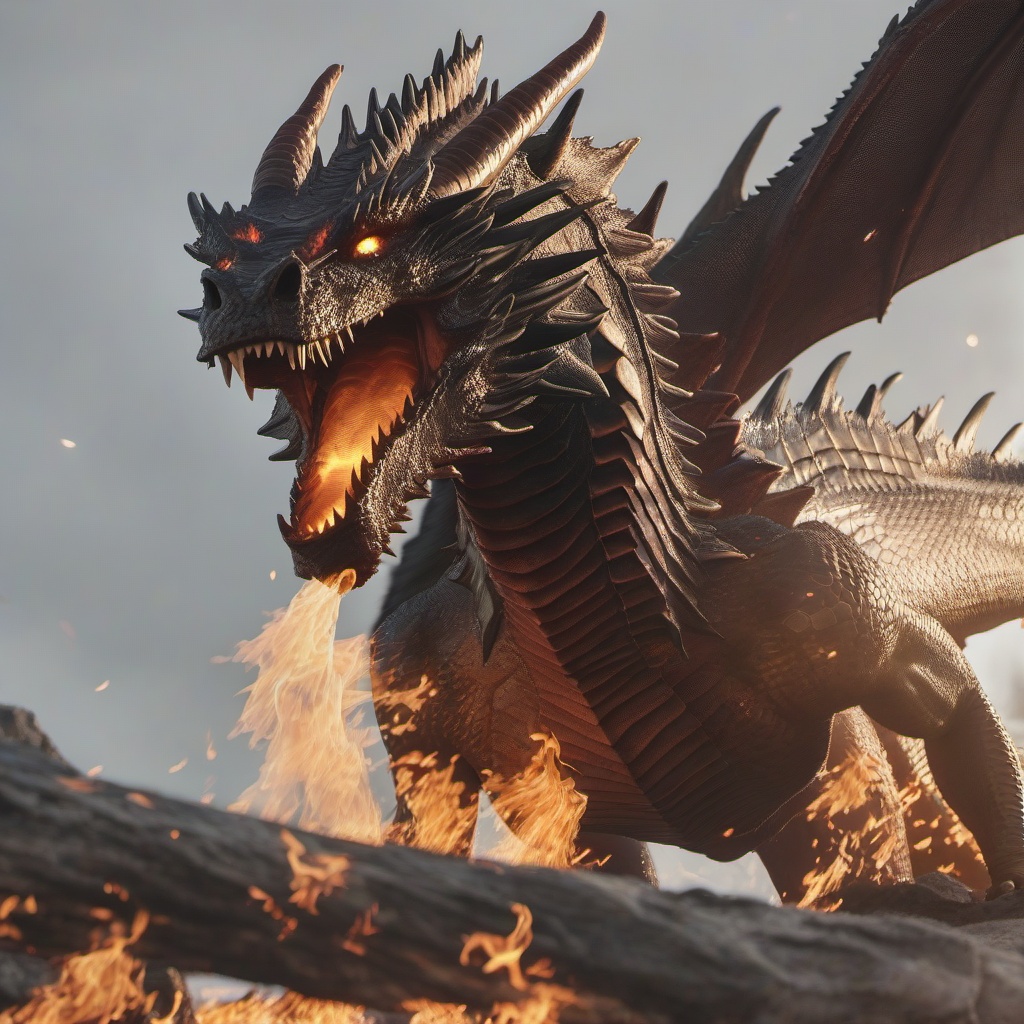}
                        };
                        \node[rotate=0, align=center, font=\normalsize, anchor=base, yshift=4pt] at (image.north) {\gls{method} (Ours) \strut};
                    \end{tikzpicture}
                \end{subfigure}
                \caption*{{a close-up of a fire spitting dragon, cinematic shot.}}
            \end{subfigure}
        \end{adjustbox}
            \begin{adjustbox}{valign=t}
                \begin{subfigure}{0.49\textwidth}
                    \begin{subfigure}{0.495\textwidth}
                        \begin{tikzpicture}
                            \node[anchor=south west, inner sep=0] (image) at (0,0) {
                                \includegraphics[width=\textwidth]{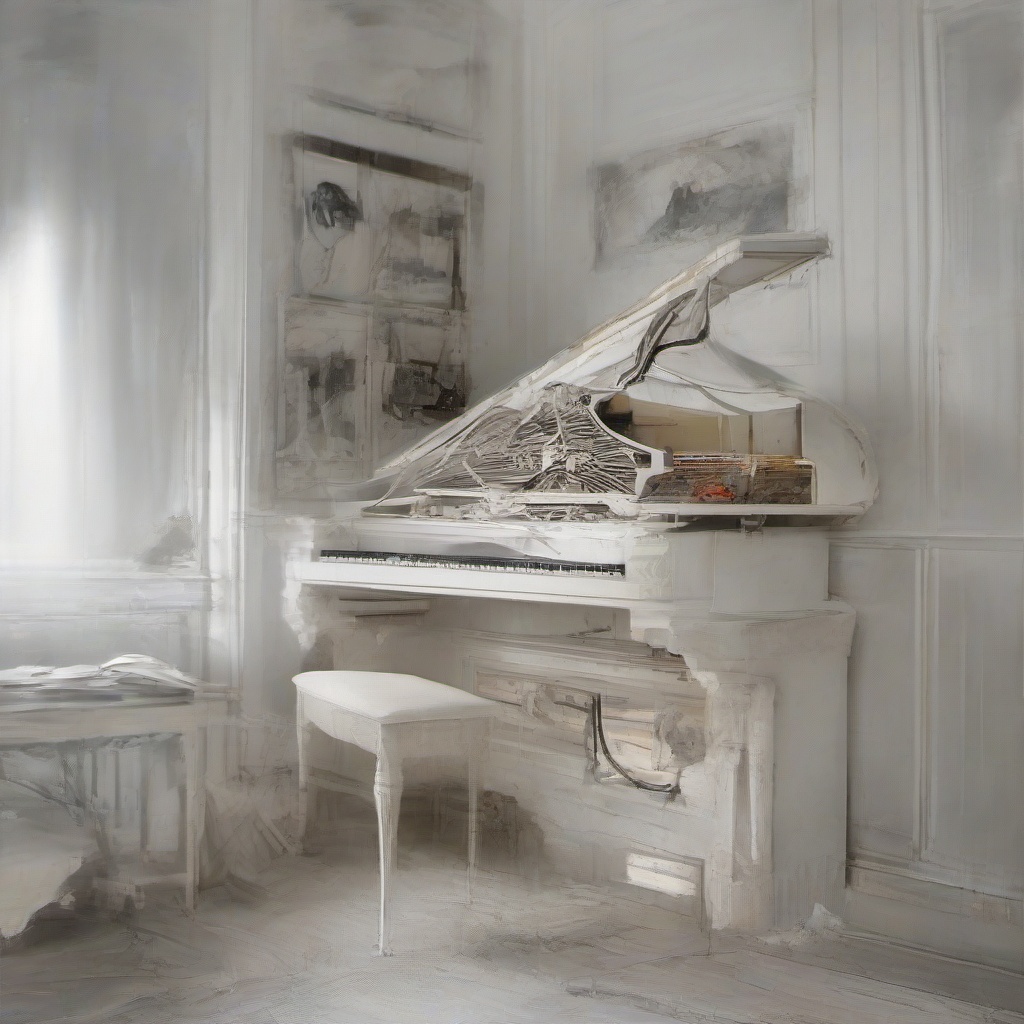}
                            };
                            \node[rotate=0, align=center, font=\normalsize, anchor=base, yshift=4pt] at (image.north) {CFG \strut};
                        \end{tikzpicture}
                    \end{subfigure}
                    \begin{subfigure}{0.495\textwidth}
                        \begin{tikzpicture}
                            \node[anchor=south west, inner sep=0] (image) at (0,0) {
                                \includegraphics[width=\textwidth]{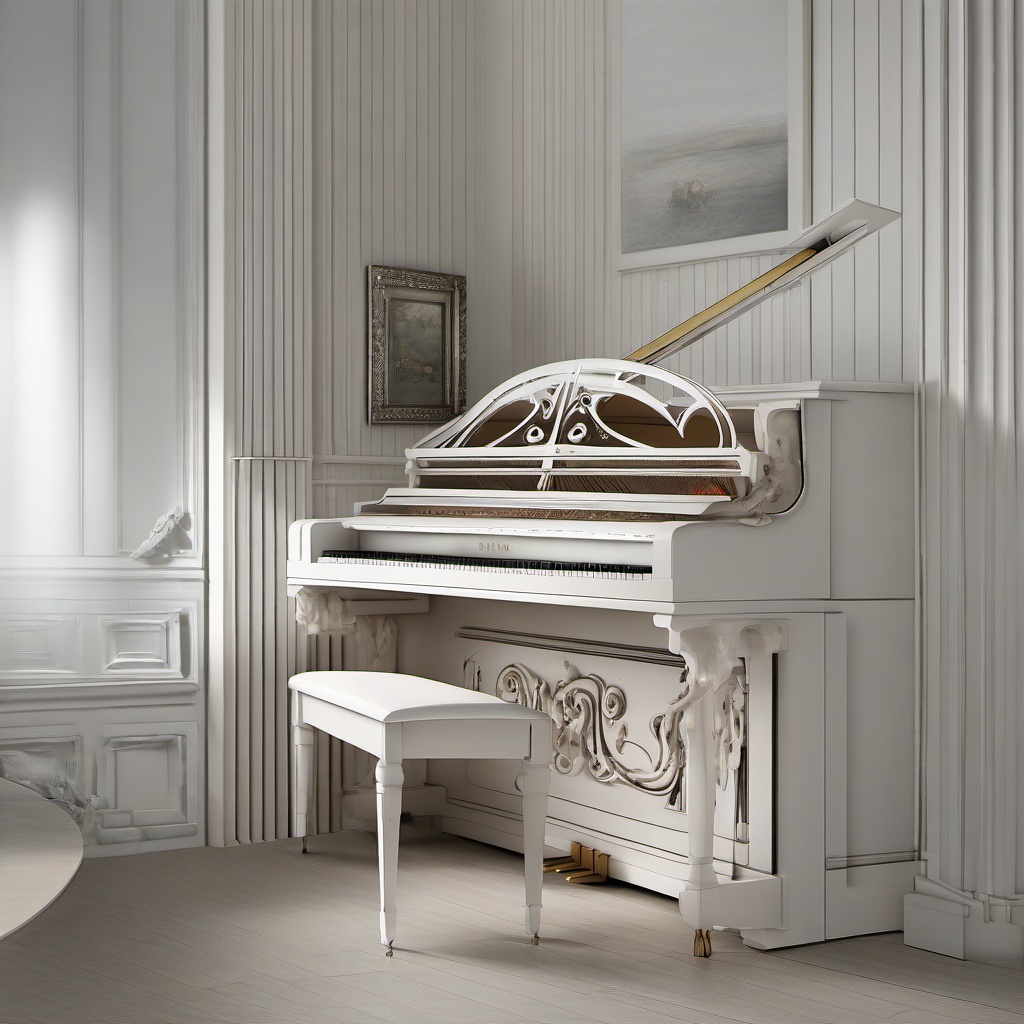}
                            };
                            \node[rotate=0, align=center, font=\normalsize, anchor=base, yshift=4pt] at (image.north) {\gls{method} (Ours) \strut};
                        \end{tikzpicture}
                    \end{subfigure}
                    \caption*{A white piano}
                \end{subfigure}
            \end{adjustbox}
        \end{minipage}
        \caption{More visual examples comparing \gls{method} with CFG using Stable Diffusion XL.}
        \label{fig:sdxl-appendix}
    \end{figure}
}

\newcommand{\figSDThree}{
    \begin{figure}[t!]
        \centering
        \begin{minipage}{\textwidth}
            \begin{subfigure}{0.49\textwidth}
                \begin{subfigure}{0.495\textwidth}
                    \begin{tikzpicture}
                        \node[anchor=south west, inner sep=0] (image) at (0,0) {
                            \includegraphics[width=\textwidth]{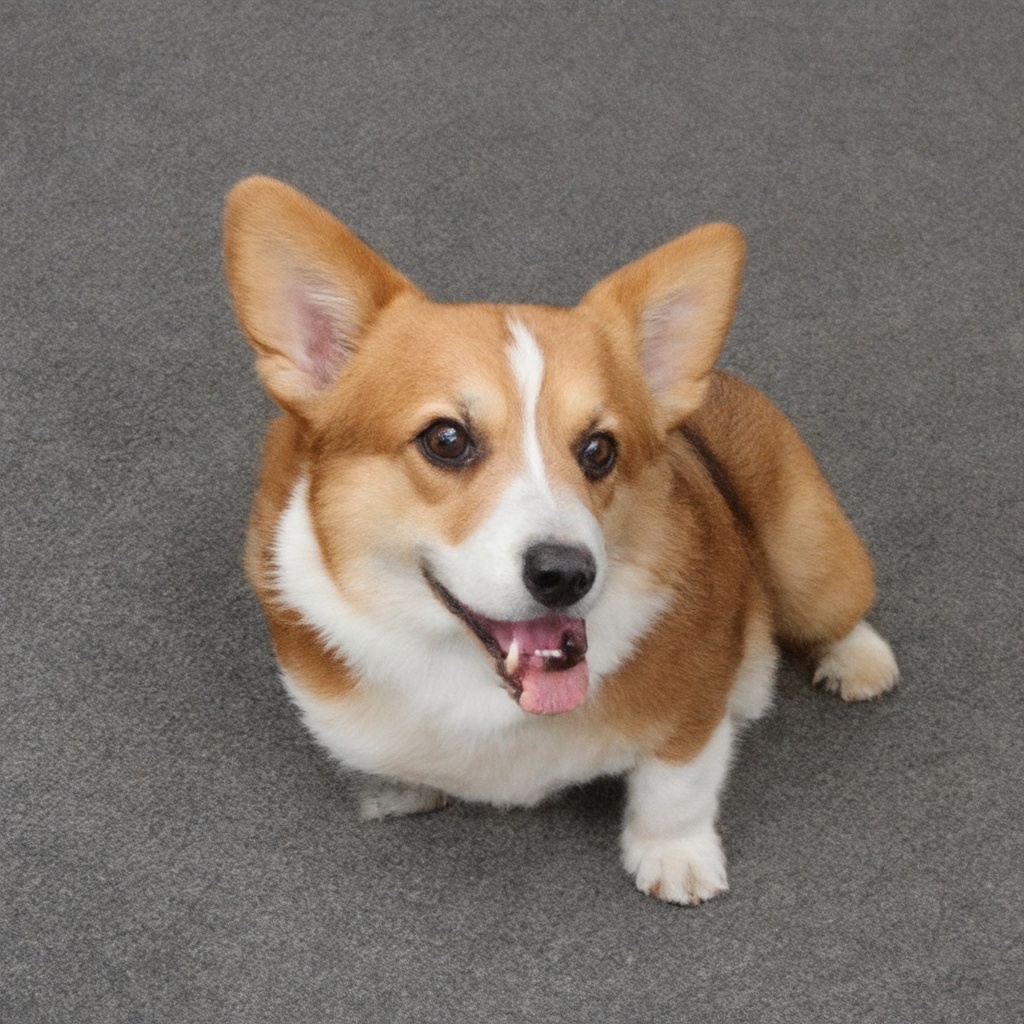}
                        };
                        \node[rotate=0, align=center, font=\normalsize, anchor=base, yshift=4pt] at (image.north) {CFG \strut};
                    \end{tikzpicture}
                \end{subfigure}
                \begin{subfigure}{0.495\textwidth}
                    \begin{tikzpicture}
                        \node[anchor=south west, inner sep=0] (image) at (0,0) {
                            \includegraphics[width=\textwidth]{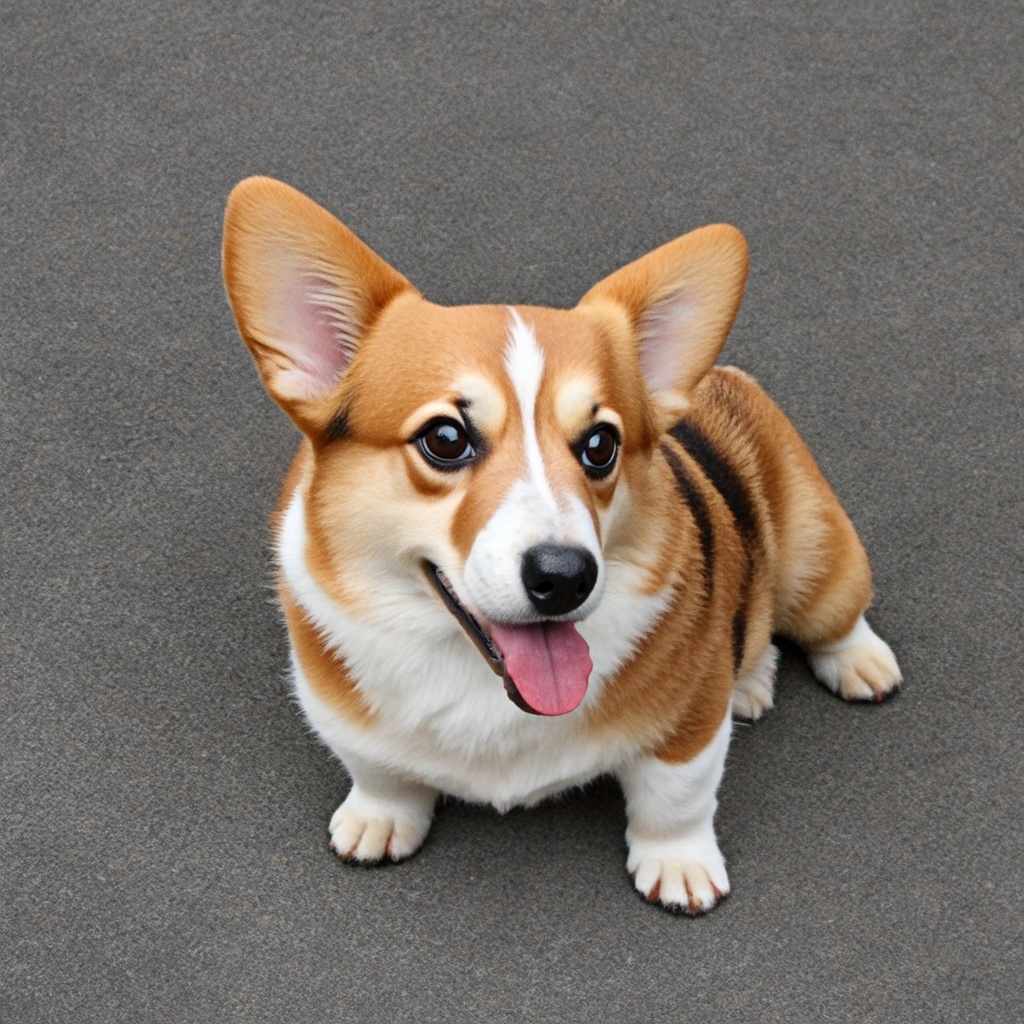}
                        };
                        \node[rotate=0, align=center, font=\normalsize, anchor=base, yshift=4pt] at (image.north) {\gls{method} (Ours) \strut};
                    \end{tikzpicture}
                \end{subfigure}
                \caption*{A photo of a corgi}
            \end{subfigure}
            \begin{subfigure}{0.49\textwidth}
                \begin{subfigure}{0.495\textwidth}
                    \begin{tikzpicture}
                        \node[anchor=south west, inner sep=0] (image) at (0,0) {
                            \includegraphics[width=\textwidth]{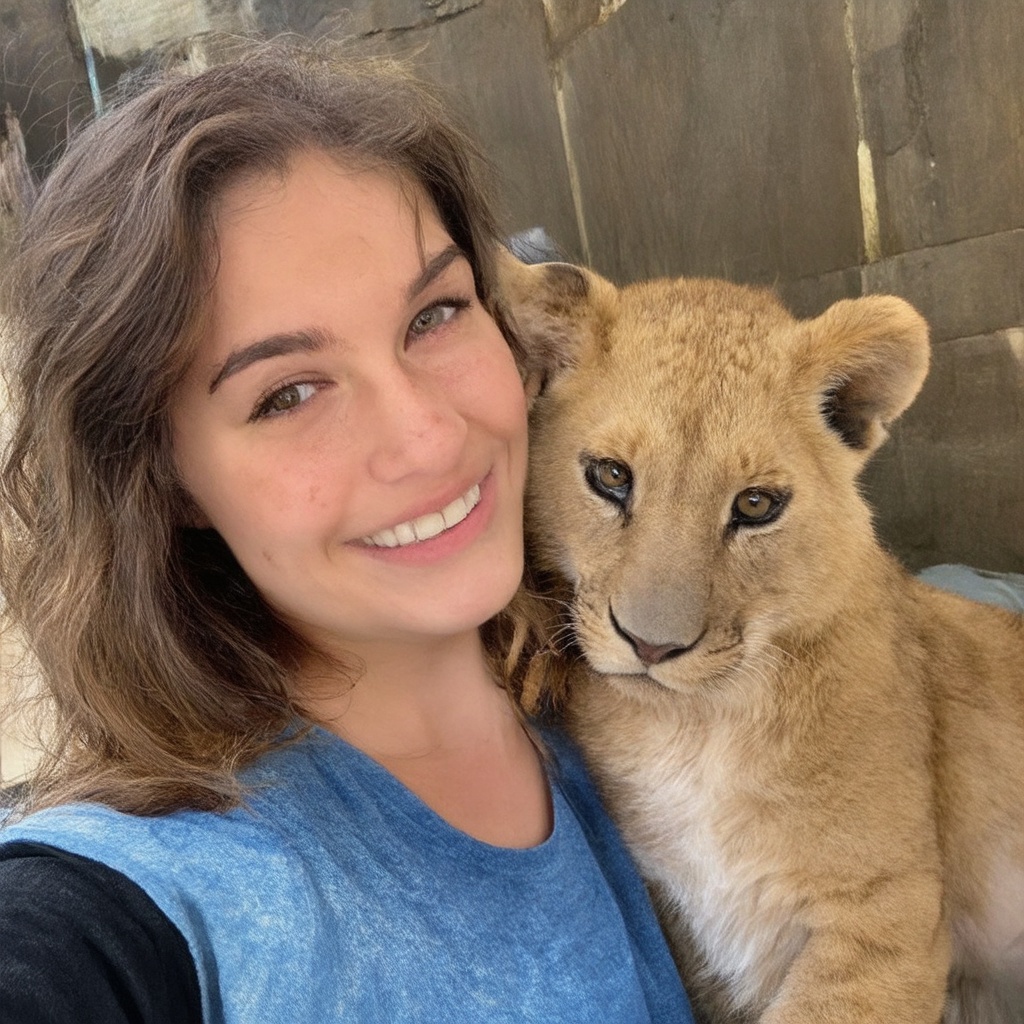}
                        };
                        \node[rotate=0, align=center, font=\normalsize, anchor=base, yshift=4pt] at (image.north) {CFG \strut};
                    \end{tikzpicture}
                \end{subfigure}
                \begin{subfigure}{0.495\textwidth}
                    \begin{tikzpicture}
                        \node[anchor=south west, inner sep=0] (image) at (0,0) {
                            \includegraphics[width=\textwidth]{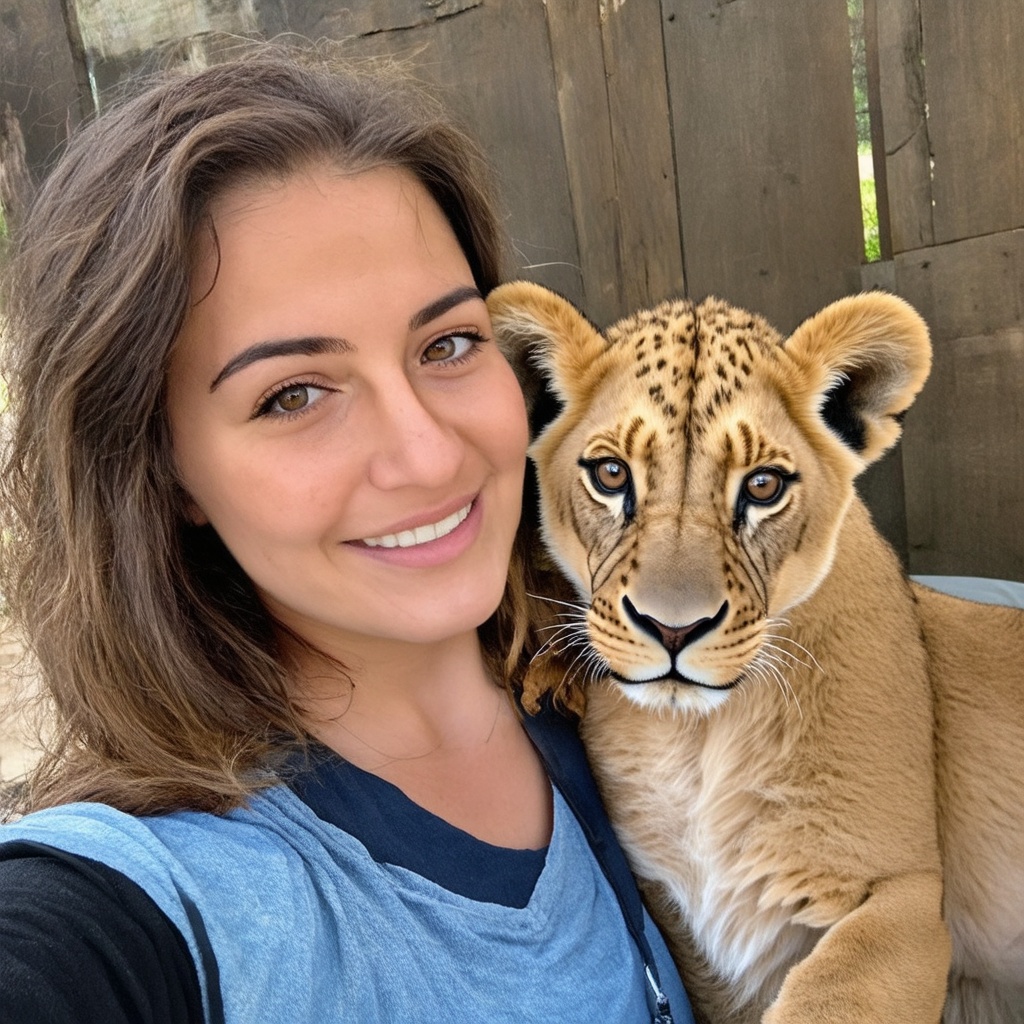}
                        };
                        \node[rotate=0, align=center, font=\normalsize, anchor=base, yshift=4pt] at (image.north) {\gls{method} (Ours) \strut};
                    \end{tikzpicture}
                \end{subfigure}
                \caption*{Selfie of a woman and her lion cub}
            \end{subfigure}
            
            \begin{adjustbox}{valign=t}
            \begin{subfigure}{0.49\textwidth}
                \begin{subfigure}{0.495\textwidth}
                    \begin{tikzpicture}
                        \node[anchor=south west, inner sep=0] (image) at (0,0) {
                            \includegraphics[width=\textwidth]{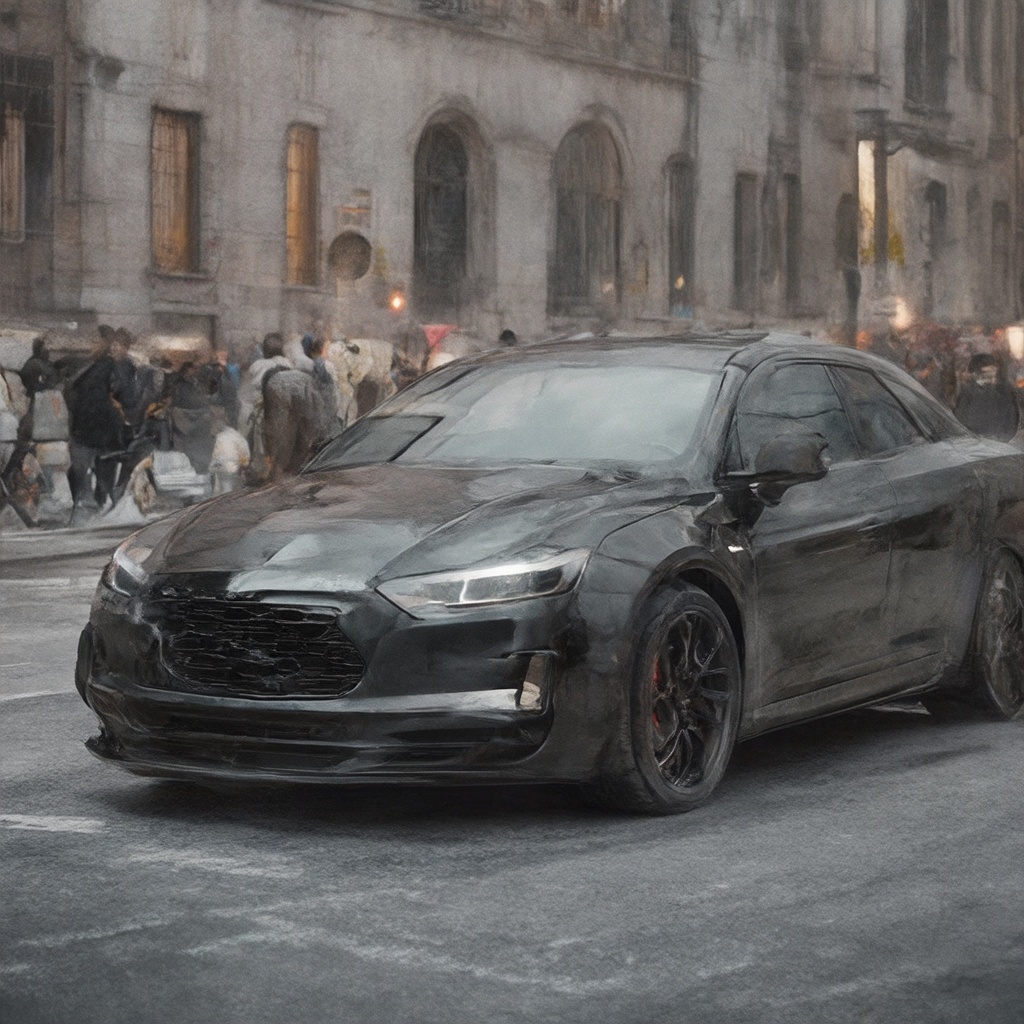}
                        };
                        \node[rotate=0, align=center, font=\normalsize, anchor=base, yshift=4pt] at (image.north) {CFG \strut};
                    \end{tikzpicture}
                \end{subfigure}
                \begin{subfigure}{0.495\textwidth}
                    \begin{tikzpicture}
                        \node[anchor=south west, inner sep=0] (image) at (0,0) {
                            \includegraphics[width=\textwidth]{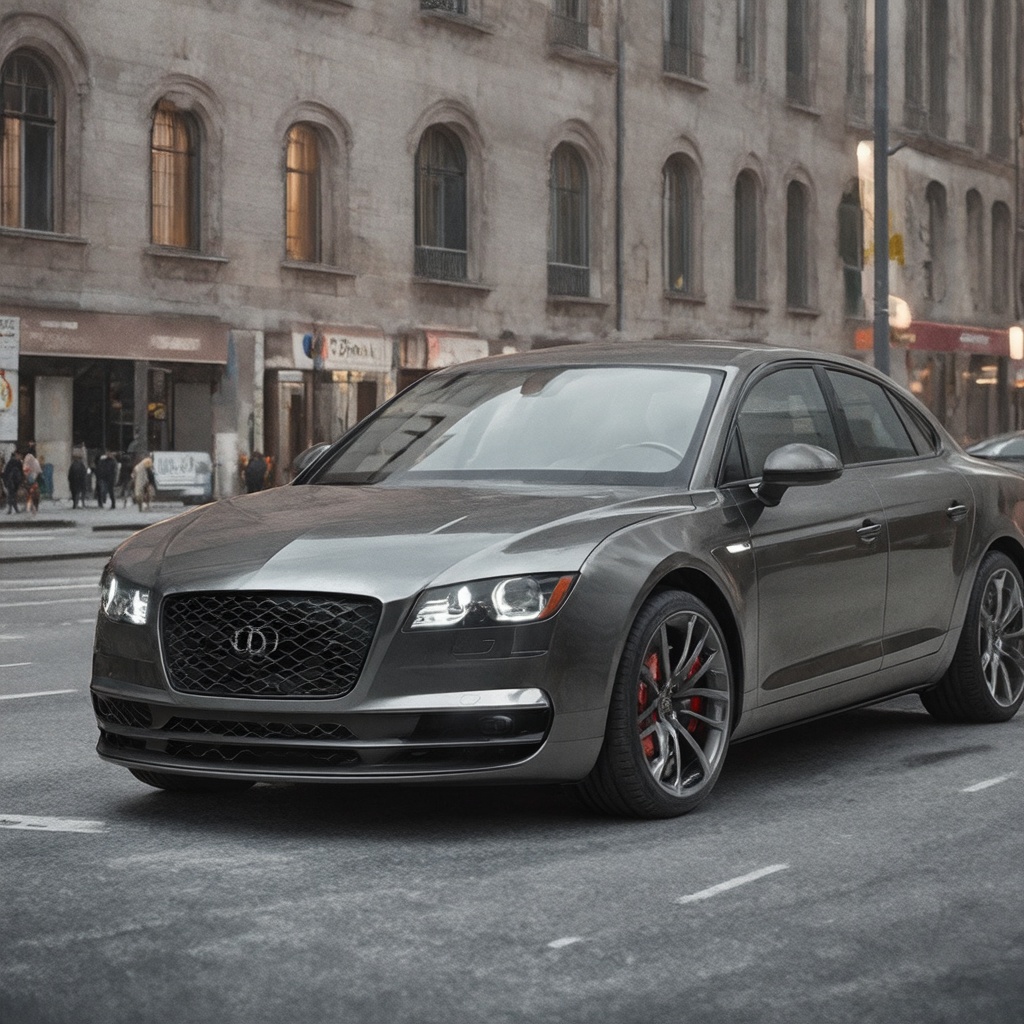}
                        };
                        \node[rotate=0, align=center, font=\normalsize, anchor=base, yshift=4pt] at (image.north) {\gls{method} (Ours) \strut};
                    \end{tikzpicture}
                \end{subfigure}
                \caption*{{Realistic image of a car}}
            \end{subfigure}
        \end{adjustbox}
            \begin{adjustbox}{valign=t}
                \begin{subfigure}{0.49\textwidth}
                    \begin{subfigure}{0.495\textwidth}
                        \begin{tikzpicture}
                            \node[anchor=south west, inner sep=0] (image) at (0,0) {
                                \includegraphics[width=\textwidth]{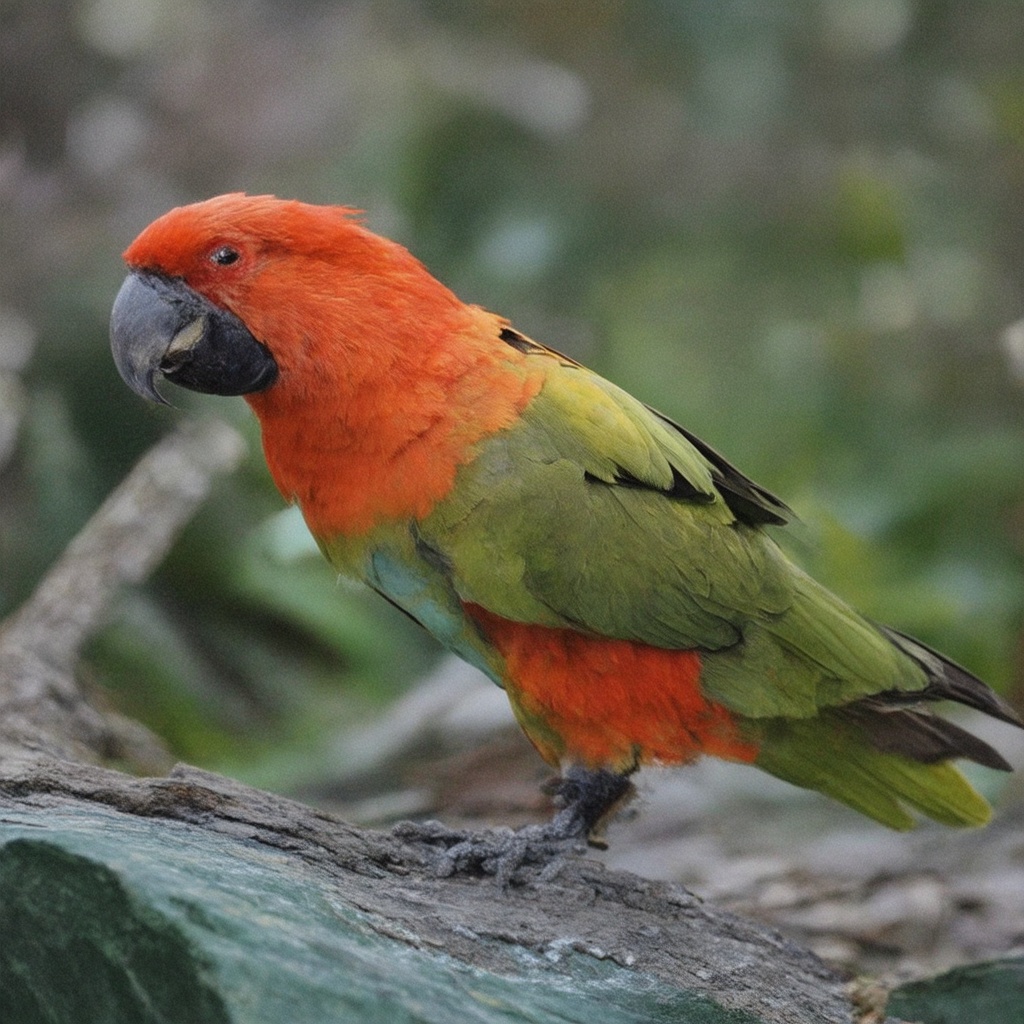}
                            };
                            \node[rotate=0, align=center, font=\normalsize, anchor=base, yshift=4pt] at (image.north) {CFG \strut};
                        \end{tikzpicture}
                    \end{subfigure}
                    \begin{subfigure}{0.495\textwidth}
                        \begin{tikzpicture}
                            \node[anchor=south west, inner sep=0] (image) at (0,0) {
                                \includegraphics[width=\textwidth]{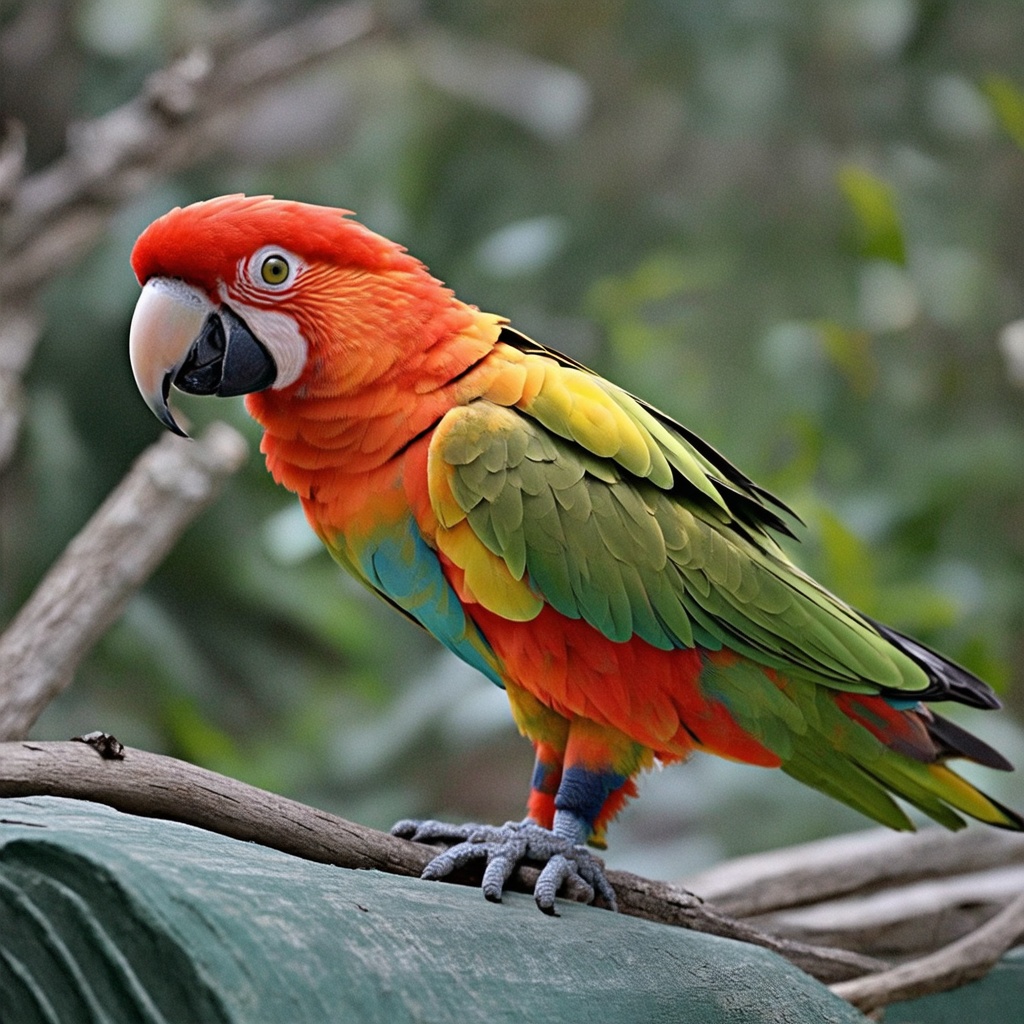}
                            };
                            \node[rotate=0, align=center, font=\normalsize, anchor=base, yshift=4pt] at (image.north) {\gls{method} (Ours) \strut};
                        \end{tikzpicture}
                    \end{subfigure}
                    \caption*{Image of a parrot}
                \end{subfigure}
            \end{adjustbox}
        \end{minipage}
        \caption{More visual examples comparing \gls{method} with CFG using Stable Diffusion 3.}
        \label{fig:sdthree-appendix}
    \end{figure}
}
\newcommand{\tabResultMain}{
    \begin{table}[t!]
        \vspace*{-0.7cm}
        \centering
        \caption{Quantitative comparison between CFG and \gls{method}. \gls{method} consistently improves FID and recall, showing high generation quality while maintaining good diversity.}
        \label{tab:main-results}
        \maxsizebox{\textwidth}{!}{
            \small
            \begin{tabular}{llccccccc}
                \toprule
                Model & Guidance & FID $\downarrow$ & Precision $\uparrow$ & Recall $\uparrow$\\
                \midrule
                
                \multirow{2}{*}{EDM2-S \citep{karras2023analyzing}} & CFG & 9.77  & \textbf{0.85} & 0.52  \\
                & \mycc \gls{method} (Ours) & \mycc \textbf{5.44} & \mycc {0.83} & \mycc \textbf{0.68}\\
                \midrule

                \multirow{2}{*}{EDM2-XXL \citep{karras2023analyzing}} & CFG & 8.65   & \textbf{0.83} & 0.56  \\
                & \mycc \gls{method} (Ours) & \mycc \textbf{4.99} & \mycc {0.82} & \mycc \textbf{0.68}\\
                \midrule

                \multirow{2}{*}{DiT-XL/2 \citep{peeblesScalableDiffusionModels2022}} & CFG & 9.31  & \textbf{0.89} & 0.54 \\
                & \mycc \gls{method} (Ours) & \mycc \textbf{5.33} & \mycc {0.84} & \mycc \textbf{0.65}\\
                \midrule

                \multirow{2}{*}{Stable Diffusion 2.1 \citep{rombachHighResolutionImageSynthesis2022}} & CFG & 24.99 & {0.68} & 0.44 \\
                & \mycc \gls{method} (Ours) & \mycc \textbf{23.33} & \mycc \textbf{0.69} & \mycc \textbf{0.49}\\
                \midrule

                \multirow{2}{*}{Stable Diffusion XL \citep{sdxl}} & CFG & 25.23  & \textbf{0.64} & 0.49 \\
                & \mycc \gls{method} (Ours) & \mycc \textbf{24.60} & \mycc {0.62} & \mycc \textbf{0.52}\\
                \midrule

                \multirow{2}{*}{Stable Diffusion 3 \citep{esser2024scaling}} & CFG & 30.32  & \textbf{0.76} & 0.37 \\
                & \mycc \gls{method} (Ours) & \mycc \textbf{27.68} & \mycc \textbf{0.76} & \mycc \textbf{0.42}\\

                \bottomrule
            \end{tabular}
        }

    \end{table}
}

\newcommand{\tabResultSD}{
    \begin{table}[t!]
    \vspace*{-0.3cm}
        \centering
        \caption{Quantitative comparison of CFG and \gls{method} across various evaluation metrics for text-to-image models. \gls{method} consistently outperforms CFG on all metrics and models.}
        \label{tab:sd-results}
        \maxsizebox{\textwidth}{!}{
            \begin{booktabs}{
        colspec = {Q[l, m]Q[l, m]Q[l, m]Q[c,m]Q[c, m]Q[c, m]Q[c, m]},
        rows = {m},
        cell{3,5,7,9,11,13,15,17,19}{3-Z}={LightGray},
            }
                \toprule
                Benchmark & Model & Guidance & ImageReward $\uparrow$ & HPSv2 $\uparrow$ & PickScore $\uparrow$ & CLIP Score $\uparrow$\\
                \midrule
                \SetCell[r=6]{m} DrawBench \citep{saharia2022photorealistic} &  \SetCell[r=2]{m} Stable Diffusion 2.1 & CFG & \textminus {0.112} & 0.273& 0.45& 30.82 \\
                & & \gls{method} (Ours) & \textbf{0.101} & \textbf{0.279} & \textbf{0.55} & \textbf{31.81} \\
                \cmidrule{2-Z}
                & \SetCell[r=2]{m} Stable Diffusion XL & CFG & 0.323  &{0.277} & 0.38 & 31.99 \\
                & & \gls{method} (Ours) & \textbf{0.595} & \textbf{0.286} & \textbf{0.62} & \textbf{32.78}\\
                \cmidrule{2-Z}
                & \SetCell[r=2]{m} Stable Diffusion 3 & CFG & 0.185  & {0.273} & 0.33 & 31.26 \\
                & & \gls{method} (Ours) & \textbf{0.803} & \textbf{0.288} & \textbf{0.67} & \textbf{33.08}\\
                \cmidrule{1-Z}
                \SetCell[r=6]{m} Parti Prompts \citep{yu2022scaling} &  \SetCell[r=2]{m} Stable Diffusion 2.1 & CFG & 0.109 & 0.270& 0.42& 31.55 \\
                & & \gls{method} (Ours) & \textbf{0.332} & \textbf{0.277} & \textbf{0.58} & \textbf{32.09} \\
                \cmidrule{2-Z}
                & \SetCell[r=2]{m} Stable Diffusion XL & CFG & 0.442  &{0.276} & 0.48 & 32.22 \\
                & & \gls{method} (Ours) & \textbf{0.663} & \textbf{0.283} & \textbf{0.52} & \textbf{32.81}\\
                \cmidrule{2-Z}
                & \SetCell[r=2]{m} Stable Diffusion 3 & CFG & 0.527  & {0.272} & 0.31 & 31.84 \\
                & & \gls{method} (Ours) & \textbf{0.988} & \textbf{0.286} & \textbf{0.69} & \textbf{32.83}\\
                \cmidrule{1-Z}
                \SetCell[r=6]{m} HPS Prompts \citep{wu2023human} &  \SetCell[r=2]{m} Stable Diffusion 2.1 & CFG & {0.035} & 0.269& 0.35& 32.73 \\
                & & \gls{method} (Ours) & \textbf{0.282} & \textbf{0.276} & \textbf{0.65} & \textbf{33.56} \\
                \cmidrule{2-Z}
                & \SetCell[r=2]{m} Stable Diffusion XL & CFG &  0.603  & {0.278} & 0.35 & 33.99 \\
                & & \gls{method} (Ours) & \textbf{0.824} & \textbf{0.286} & \textbf{0.65} & \textbf{34.77}\\
                \cmidrule{2-Z}
                & \SetCell[r=2]{m} Stable Diffusion 3 & CFG & 0.470  & {0.272} & 0.29 & 32.22 \\
                & & \gls{method} (Ours) & \textbf{1.00} & \textbf{0.288} & \textbf{0.71} & \textbf{33.73}\\

                \bottomrule
            \end{booktabs}
        }
        \vspace*{-0.3cm}
    \end{table}
}

\newcommand{\tabSamplers}{
    \begin{table}[t]
        \centering
        \caption{Impact of applying \gls{method} with popular diffusion samplers on the class-conditional ImageNet model (DiT-XL/2). \gls{method} achieves improved FID and recall across all samplers.}
        \label{table:samplers}
        {
            \small
            \begin{tabular}{lcccccc}
                \toprule
                & \multicolumn{2}{c}{{\gls{method} (Ours)}} & \multicolumn{2}{c}{{CFG}} \\
                \cmidrule(lr){2-3} \cmidrule(lr){4-5}
                Sampler & FID $\downarrow$ & Recall $\uparrow$ & FID $\downarrow$ & Recall $\uparrow$  \\
                \midrule
                DDIM \citep{songDenoisingDiffusionImplicit2022} & \mycc \textbf{4.84} & \mycc \textbf{0.69} & 6.91 & 0.60 \\
                DPM++ \citep{lu2022dpm} & \mycc \textbf{4.77} & \mycc \textbf{0.69} & 7.11 & 0.61 \\
                SDE-DPM++ \citep{lu2022dpm} &\mycc \textbf{5.06} & \mycc \textbf{0.68} & 9.10 & 0.56 \\
                PNDM \citep{liu2022pseudo} & \mycc \textbf{4.75} & \mycc \textbf{0.69} & 7.02 & 0.61 \\
                UniPC \citep{zhao2023unipc} & \mycc \textbf{4.76} & \mycc \textbf{0.69} & 7.16 & 0.60 \\
                \bottomrule
            \end{tabular}
        }
    \end{table}
}

\newcommand{\tabAblationMain}{
    \begin{table}
        \centering
        \caption{Ablation of the frequency decomposition operator $\opfreq$ based on the EDM2 model \citep{karras2023analyzing}.}
        \label{tab:ablation-main}
        \begin{subtable}[b]{0.49\textwidth}
            \centering
            \caption{The choice of the frequency decomposition function}
            \label{tab:ablation-dwt}
            \maxsizebox{\textwidth}{!}
            {
            \begin{booktabs}{
                        colspec = {Q[c, m]Q[c, m]Q[c, m]Q[c, m]},
                        row{3-Z} = {LightGray}
                    }
                    \toprule
                    $\opfreq$ & FID $\downarrow$ & Precision $\uparrow$ & Recall $\uparrow$\\
                    \midrule
                    CFG & 8.89 & \textbf{0.85} & 0.55  \\
                    Laplacian pyramid & \textbf{5.12} & {0.83} & {0.67}\\
                    Wavelet transform & {5.26} &  {0.81} & \textbf{0.71}\\

                    \bottomrule
                    \end{booktabs}
            }
        \end{subtable}
        \hfil
        \begin{subtable}[b]{0.42\textwidth}
            \centering
            \caption{multi-level vs single-level transformation}
            \label{tab:multilevel}
            \maxsizebox{\textwidth}{!}
            {
            \begin{booktabs}{
                        colspec = {Q[l, m]Q[c, m]Q[c, m]Q[c, m]},
                        row{3-Z} = {LightGray}
                    }
                    \toprule
                    Config & FID $\downarrow$ & Precision $\uparrow$ & Recall $\uparrow$\\
                    \midrule
                    CFG & 8.89 & \textbf{0.85} & 0.55  \\
                    single-level & \textbf{5.12} & {0.83} & \textbf{0.67}\\
                    multi-level & {5.25} &  {0.83} & \textbf{0.67}\\
                    \bottomrule
                    \end{booktabs}
            }
        \end{subtable} 

    \end{table}
}

\newcommand{\tabGIComp}{
    \begin{table}[t!]
        \centering
        \caption{Comparing guidance interval \citep{intervalGuidance} with \gls{method} based on Stable Diffusion XL \citep{sdxl}. \gls{method} achieves better quality metrics due to having high-frequency guidance in earlier sampling steps.}
        \begin{minipage}{\linewidth}
              \centering
                {
                \begin{booktabs}{
                    colspec = {Q[l]Q[c]Q[c]Q[c]Q[c]},
                    row{3} = {LightGray}}
                \toprule
                Method & ImageReward $\uparrow$ & HPSv2 $\uparrow$ & PickScore $\uparrow$ & CLIP Score $\uparrow$\\
                \midrule
                Guidance interval \citep{intervalGuidance} & 0.437 & 0.282 & 0.34 & 32.77 \\
                \gls{method} (Ours) & \textbf{0.595} & \textbf{0.286} & \textbf{0.66} & \textbf{32.78} \\
                \bottomrule
              \end{booktabs}
                }
              \label{tab:gi-comp}
            \end{minipage}%
    \end{table}
}

\newcommand{\tabSDXLLightning}{
    \begin{table}[t!]
        \centering
        \caption{Quantitative comparison between CFG and \gls{method} using SDXL-Lightning \citep{sdlightning} as an example of a distilled model that uses fewer sampling steps. \gls{method} outperforms sampling both with and without CFG, achieving better quality and good prompt alignment.}
        \begin{minipage}{\linewidth}
              \centering
                {
                \begin{booktabs}{
                    colspec = {Q[l]Q[c]Q[c]Q[c]},
                    row{4} = {LightGray}}
                \toprule
                Method & ImageReward $\uparrow$ & HPSv2 $\uparrow$ & CLIP Score $\uparrow$\\
                \midrule
                w/o CFG & 0.535 & 0.282 & 31.77 \\
                CFG & 0.573 & 0.286 & \textbf{32.44}\\
                \gls{method} (Ours) & \textbf{0.672} & \textbf{0.292} & \textbf{32.44} \\
                \bottomrule
              \end{booktabs}
                }
              \label{tab:sdxl-distilled}
            \end{minipage}%
    \end{table}
}

\newcommand{\tabParameters}{
\begin{table}[t!]
    \begin{minipage}{\textwidth}
    \centering
    \caption{Guidane parameters used to compare the performance of \gls{method} with CFG.}
    \label{tab:imp-detail}
    \begin{subtable}{0.495\textwidth}
        \caption{Guidance parameters used for Table 1.}
        \maxsizebox{\linewidth}{!}{
    \begin{booktabs}{lccc}
        \toprule
        Model & $w$ & $\wcfglow$ &  $\wcfghigh$  \\
        \midrule
        EDM2-S & 3 & 1 & 3  \\
        EDM2-XL & 2 & 1 & 2  \\
        DiT-XL/2 & 2 & 1 & 2\\
        Stable Diffusion 2.1 & 7 & 3 & 7 \\
        Stable Diffusion XL & 10 & 5 & 10\\
        Stable Diffusion 3 & 7 & 3 & 7 \\
        \bottomrule
        \end{booktabs}
        }
    \end{subtable}
    \begin{subtable}{0.495\textwidth}
        \caption{Guidance parameters used for Table 2.}
        \maxsizebox{\linewidth}{!}{
    \begin{booktabs}{lccc}
        \toprule
        Model & $w$ & $\wcfglow$ &  $\wcfghigh$  \\
        \midrule
        Stable Diffusion 2.1 & 3 & 3 & 12 \\
        Stable Diffusion XL & 3 & 3 & 12  \\
        Stable Diffusion 3 & 1.5 & 1.5 & 12 \\
        \bottomrule
        \end{booktabs}
        }
    \end{subtable}
    \end{minipage}
\end{table}
}

\newcommand{\tabCADS}{
    \begin{table}[t]
    \centering
    \begin{minipage}{\linewidth}
        \centering
        \caption{Effectiveness of CADS on \gls{method} using DiT-XL/2 at a guidance scale of 5. Combining \gls{method} with CADS yields the best FID, outperforming each method used in isolation.}
    \label{table:cads}
    \maxsizebox{\linewidth}{!}{
        \begin{tabular}{lccc}
          \toprule
          Guidance & FID $\downarrow$ & Precision $\uparrow$ & Recall $\uparrow$ \\
          \midrule
          CFG & 21.48 & \textbf{0.92} & 0.30 \\
          \gls{method} (Ours) & 13.90 & {0.90} & 0.48 \\
          CFG + CADS & {14.53} & 0.88 & {0.48} \\
          \gls{method} + CADS (Ours) & \textbf{8.98} & 0.81 & \textbf{0.61} \\
          \bottomrule
        \end{tabular}
        }
    \end{minipage}
\end{table}
}
\newcommand{\EDMCFGHighScalesPlot}{
    \begin{figure}[t!]
        \vspace*{-0.75cm}
        \centering
        \begin{minipage}{\textwidth}
            \resizebox{\textwidth}{!}{
                \begin{tikzpicture}
                    \begin{groupplot}[
                            group style={
                                    group size=4 by 1,
                                    horizontal sep=1.25cm,
                                },
                            xlabel={$w$},
                            ymajorgrids=true,
                            xmajorgrids=true,
                            grid style=dashed,
                            major grid style = {lightgray},
                            tick label style={font=\large},
                            label style={font=\huge},
                            title style={font=\huge},
                            legend pos={north west}, legend cell align={left},
                            legend style={font=\Large},
                            scale only axis,
                        ]
                        \nextgroupplot[title={FID}]
                        \addplot [C0, mark=*, very thick,dashed, mark options={solid}] table [x index=0, y index=1, col sep=space] {data/edm_cfg.dat};
                        \addplot [C2, mark=*, very thick, dotted, mark options={solid}] table [x index=0, y index=1, col sep=space] {data/edm_cfg_low.dat};
                        \addplot [C1, mark=*, very thick, mark options={solid}] table [x index=0, y index=1, col sep=space] {data/edm_cfg_high.dat};

                        \nextgroupplot[title={Precision}]
                        \addplot [C0, mark=*, very thick, dashed, mark options={solid}] table [x index=0, y index=3, col sep=space] {data/edm_cfg.dat};
                        \addplot [C2, mark=*, very thick, dotted, mark options={solid}] table [x index=0, y index=3, col sep=space] {data/edm_cfg_low.dat};
                        \addplot [C1, mark=*, very thick, mark options={solid}] table [x index=0, y index=3, col sep=space] {data/edm_cfg_high.dat};

                        \nextgroupplot[title={Recall}]
                        \addplot [C0, mark=*, very thick, dashed, mark options={solid}] table [x index=0, y index=4, col sep=space] {data/edm_cfg.dat};
                        \addplot [C2, mark=*, very thick, dotted, mark options={solid}] table [x index=0, y index=4, col sep=space] {data/edm_cfg_low.dat};
                        \addplot [C1, mark=*, very thick, mark options={solid}] table [x index=0, y index=4, col sep=space] {data/edm_cfg_high.dat};

                        \nextgroupplot[title={Saturation}]
                        \addplot [C0, mark=*, very thick, dashed, mark options={solid}] table [x index=0, y index=5, col sep=space] {data/edm_cfg.dat};
                        \addplot [C2, mark=*, very thick, dotted, mark options={solid}] table [x index=0, y index=5, col sep=space] {data/edm_cfg_low.dat};
                        \addplot [C1, mark=*, very thick, mark options={solid}] table [x index=0, y index=5, col sep=space] {data/edm_cfg_high.dat};

                        \legend{$\wcfglow=\wcfghigh=w$, {$\wcfglow=w$, $\wcfghigh=1$}, {$\wcfglow=1$, $\wcfghigh=w$}}
                    \end{groupplot}
                \end{tikzpicture}
            }
        \end{minipage}
            
        \caption{Illustrating the effect of different frequency components on CFG behavior. Although CFG improves quality, it rapidly restricts diversity and increases saturation, leading to higher FID and lower recall. Note that the low-frequency component is mainly responsible for these adverse effects, whereas the high-frequency component enhances quality while preserving diversity and color composition, resulting in better FID and recall.}
        \label{fig:scales}
        \vspace*{-0.5cm}
    \end{figure}
}

\newcommand{\NFEPlot}{
    \begin{figure}[t!]
        \centering
        \begin{minipage}{\textwidth}
            \resizebox{\textwidth}{!}{
                \begin{tikzpicture}
                    \begin{groupplot}[
                            group style={
                                    group size=4 by 1,
                                    horizontal sep=1.25cm,
                                },
                            xlabel={\# Steps},
                            ymajorgrids=true,
                            xmajorgrids=true,
                            grid style=dashed,
                            major grid style = {lightgray},
                            tick label style={font=\normalsize},
                            label style={font=\Large},
                            title style={font=\Large},
                            legend cell align={left},
                            legend style={at={(0.95,0.525)},anchor=east, font=\Large},
                            scale only axis,
                        ]
                        \nextgroupplot[title={FID}]
                        \addplot [C0, mark=*, very thick, mark options={solid}] table [x index=0, y index=1, col sep=space] {data/edm_nfe_cfg.dat};
                        \addplot [C1, mark=*, very thick, mark options={solid}] table [x index=0, y index=1, col sep=space] {data/edm_nfe_ours.dat};
                        \nextgroupplot[title={Precision}]
                        \addplot [C0, mark=*, very thick, mark options={solid}] table [x index=0, y index=3, col sep=space] {data/edm_nfe_cfg.dat};
                        \addplot [C1, mark=*, very thick, mark options={solid}] table [x index=0, y index=3, col sep=space] {data/edm_nfe_ours.dat};
                        \nextgroupplot[title={Recall}]
                        \addplot [C0, mark=*, very thick, mark options={solid}] table [x index=0, y index=4, col sep=space] {data/edm_nfe_cfg.dat};
                        \addplot [C1, mark=*, very thick, mark options={solid}] table [x index=0, y index=4, col sep=space] {data/edm_nfe_ours.dat};
                        \legend{CFG, \gls{method}}
                    \end{groupplot}
                \end{tikzpicture}
            }
        \end{minipage}
            
        \caption{Comparison between \gls{method} and CFG across different numbers of sampling steps. \gls{method} consistently outperforms CFG, maintaining a clear FID improvement at all sampling budgets.}
        \label{fig:nfe}
    \end{figure}
}

\newcommand{\EDMNormsPlot}{
    \begin{figure}[t!]
        \vspace*{-0.75cm}
        \begin{minipage}{\textwidth}
            \centering
            \begin{subfigure}{0.495\textwidth}
                \centering
                \maxsizebox{\textwidth}{!}{
                \begin{tikzpicture}
                    \begin{groupplot}[
                            group style={
                                    group size=4 by 1,
                                    horizontal sep=1.25cm,
                                },
                            xlabel={$\sigma(t)$},
                            ymajorgrids=true,
                            xmajorgrids=true,
                            grid style=dashed,
                            major grid style = {lightgray},
                            tick label style={font=\normalsize},
                            label style={font=\normalsize},
                            title style={font=\normalsize},
                            legend pos={north east}, legend cell align={left},
                            legend style={font=\small},
                            scale only axis,
                            height=0.525\textwidth,
                            width=\textwidth,
                        ]
                        \nextgroupplot[title={$\norm{\Delta D_t}$ for CFG}]
                        \addplot [C0, very thick,mark=., mark options={solid}, dashed] table [x index=0, y index=2, col sep=space] {data/edm_norms_cfg.dat};
                        \addplot [C1, very thick, mark options={solid}] table [x index=0, y index=1, col sep=space] {data/edm_norms_cfg.dat};
                        \fill[pattern=north west lines, pattern color=gray] (axis cs:0.6,0) rectangle (axis cs:5,10);
                    \end{groupplot}
                \end{tikzpicture}
                    }
            \end{subfigure}
            \begin{subfigure}{0.495\textwidth}
                \centering
                \maxsizebox{\textwidth}{!}{
                \begin{tikzpicture}
                    \begin{groupplot}[
                            group style={
                                    group size=4 by 1,
                                    horizontal sep=1.25cm,
                                },
                            xlabel={$\sigma(t)$},
                            grid=both,
                            ymajorgrids=true,
                            xmajorgrids=true,
                            grid style=dashed,
                            major grid style = {lightgray},
                            tick label style={font=\normalsize},
                            label style={font=\normalsize},
                            title style={font=\normalsize},
                            legend pos={north east}, legend cell align={left},
                            legend style={font=\small},
                            scale only axis,
                            height=0.525\textwidth,
                            width=\textwidth,
                        ]
                        \nextgroupplot[title={$\norm{\Delta D_t}$ for Autoguidance}]
                        \addplot [C0, very thick, mark=., mark options={solid}, dashed] table [x index=0, y index=2, col sep=space] {data/edm_norms_autog.dat};
                        \addplot [C1, very thick, mark options={solid}] table [x index=0, y index=1, col sep=space] {data/edm_norms_autog.dat};
                        \legend{Low frequency, High frequency}
                    \end{groupplot}
                \end{tikzpicture}
                }
            \end{subfigure}
                
        \end{minipage}
        \vspace*{-0.1cm}
        \caption{Illustration of how the norm of the guidance signal $\Delta D_t$ behaves in the frequency domain for CFG and Autoguidance. For CFG, low-frequency components are dominant during most early steps (high $\sigma(t)$), which can be harmful. Guidance interval \citep{intervalGuidance} improves this by limiting CFG to the shaded region (found by grid search), i.e., the steps where high-frequency components are dominant. In contrast, Autoguidance maintains strong high-frequency components throughout the sampling process, making the signal useful at all steps as observed by \citet{karras2024guiding}.}
        \label{fig:norm}
        \vspace*{-0.3cm}
    \end{figure}
}

\newcommand{\algMethod}{

    \begin{figure}[t!]
        \centering
        \begin{minipage}{\textwidth}
            \begin{algorithm}[H]
                \caption{Guided sampling with \gls{method}}
                \label{algo:sampling}
                \begin{algorithmic}[1]
                    \Require Frequency decomposition operators $\opfreq[\cdot]$ and $\opfreqinv[\cdot]$ (e.g., Laplacian pyramid)
                    \Require Guidance weights $\wcfglow$ (low-frequency), $\wcfghigh$ (high-frequency)
                    \Require Conditioning input $\vy$
            
                    \State Initialize: $\vz_{1} \sim \mathcal{N}(\vzero, \sigma_{\mathrm{max}}^2 \mI)$
                    \For{$t = \Set{1, 1 - \delta t, \dotsc, 0}$}
                        \vspace{0.1cm}
                        \State Compute the frequency decomposition of the conditional and unconditional predictions:
                        \[
                        \opfreq\brk{\predcond} = \Set{\oplow\brk{\predcond}, \ophigh\brk{\predcond}}
                        \]
                        \[
                            \opfreq\brk{\preduncond} = \Set{\oplow\brk{\preduncond}, \ophigh\brk{\preduncond}}
                        \]
                        
                        \State Compute the low- and high-frequency components of \gls{method}
                        \[
                            \predguidedlowfdg = \oplow\brk{\preduncond} + \wcfglow \left( \oplow\brk{\predcond} - \oplow\brk{\preduncond} \right)
                        \]
                        \[
                            \predguidedhighfdg = \ophigh\brk{\preduncond} + \wcfghigh \left( \ophigh\brk{\predcond} - \ophigh\brk{\preduncond} \right)
                        \]

                        \State Convert the guided prediction to the data space using the inverse transform:
                        \[
                        \predguidedours = \opfreqinv[\Set{\predguidedlowfdg, \predguidedhighfdg}]
                        \]
                    
                        \State Perform one sampling step (e.g., one step of DDIM):
                        \[
                            \vz_{t-1} = \text{diffusion\_reverse}(\hat{D}_{\textnormal{\gls{method}}}, \vz_{t}, t)
                        \]
                    \EndFor
            
                    \State \textbf{return} $\vz_{0}$
                \end{algorithmic}
            \end{algorithm}
        \end{minipage}
    \end{figure}
}

\newcommand{\AligmentPlot}{
    \setlength{\columnsep}{16pt}
    \begin{wrapfigure}[9]{r}{0.4\textwidth}
        \vspace*{-1.5cm}
        \begin{center}
            \resizebox{\linewidth}{!}{
    \begin{tikzpicture}
      \begin{axis}[
          title={CLIP Score},
          xlabel={$w$},
          ymajorgrids=true,
          xmajorgrids=true,
          grid style=dashed,
          major grid style={lightgray},
          tick label style={font=\large},
          label style={font=\huge},
          title style={font=\huge},
          legend pos=south east,
          legend cell align=left,
          legend style={font=\Large},
          height=0.575\textwidth,
          width=\textwidth,
          scale only axis,
        ]
        \addplot [C0, mark=*, very thick, mark options={solid}]
          table [x index=0, y index=1, col sep=space] {data/alignment.dat};
        \addlegendentry{$\wcfglow=\wcfghigh=w$}

        \addplot [C1, mark=*, dashed, very thick, mark options={solid}]
          table [x index=0, y index=2, col sep=space] {data/alignment.dat};
        \addlegendentry{$\wcfglow=w,\;\wcfghigh=1$}

        \addplot [C2, mark=*, dotted, very thick, mark options={solid}]
          table [x index=0, y index=3, col sep=space] {data/alignment.dat};
        \addlegendentry{$\wcfglow=1,\;\wcfghigh=w$}

        \draw[darkgray,dashed,very thick]
  (axis cs:\pgfkeysvalueof{/pgfplots/xmin},33.28)
  -- (axis cs:\pgfkeysvalueof{/pgfplots/xmax},33.28) node [pos=0.12, below, font=\Large] {\gls{method} (Ours)};; 

        \addlegendentry{FDG (ours)}
      \end{axis}
    \end{tikzpicture}
  }
\end{center}
\vspace*{-0.25cm}
            \caption{Effect of frequency components of CFG on prompt alignment.}
            \label{fig:alignment}
    \end{wrapfigure}
}

\title{Guidance in the Frequency Domain Enables High-Fidelity Sampling at Low CFG Scales}

\author{Seyedmorteza Sadat\textsuperscript{1}, Tobias  Vontobel\textsuperscript{1}, Farnood Salehi\textsuperscript{2}, Romann M.\ Weber\textsuperscript{2} \\
\textsuperscript{1}ETH Z\"urich, \textsuperscript{2}DisneyResearch\textbar{}Studios\\
\texttt{\{votobias, ssadat\}@ethz.ch} \\
\texttt{\{farnood.salehi, romann.weber\}@disneyresearch.com}
}

\begin{document}

\maketitle
\begin{abstract}
  Classifier-free guidance (CFG) has become an essential component of modern conditional diffusion models. Although highly effective in practice, the underlying mechanisms by which CFG enhances quality, detail, and prompt alignment are not fully understood. We present a novel perspective on CFG by analyzing its effects in the frequency domain, showing that low and high frequencies have distinct impacts on generation quality. Specifically, low-frequency guidance governs global structure and condition alignment, while high-frequency guidance mainly enhances visual fidelity. However, applying a uniform scale across all frequencies---as is done in standard CFG---leads to oversaturation and reduced diversity at high scales and degraded visual quality at low scales. Based on these insights, we propose \gls{method}, an effective approach that decomposes CFG into low- and high-frequency components and applies separate guidance strengths to each component. \gls{method} improves image quality at low guidance scales and avoids the drawbacks of high CFG scales by design. Through extensive experiments across multiple datasets and models, we demonstrate that \gls{method} consistently enhances sample fidelity while preserving diversity, leading to improved FID and recall compared to CFG, establishing our method as a plug-and-play alternative to standard classifier-free guidance.
\end{abstract}
\figTeaser

\section{Introduction}
Diffusion models \citep{sohl2015deep,hoDenoisingDiffusionProbabilistic2020,score-sde} are a class of generative models that learn the data distribution by reversing a forward process that gradually corrupts data with Gaussian noise until it resembles pure noise. While the theory suggests that simulating this reverse process should yield high-quality samples, in practice, unguided sampling often produces low-quality images that poorly match the input condition. To address this, \gls{cfg} \citep{hoClassifierFreeDiffusionGuidance2022} has become a standard technique in modern diffusion models for improving both output quality and alignment with the conditioning signal—though often at the cost of reduced diversity \citep{hoClassifierFreeDiffusionGuidance2022,sadat2024cads} and excessive oversaturation \citep{sadat2025eliminating}.

Typically, diffusion models rely on high guidance scales to achieve better image quality and prompt alignment. However, high guidance scales degrade sample diversity and introduce color saturation artifacts \citep{hoClassifierFreeDiffusionGuidance2022}. Conversely, low CFG scales tend to produce more diverse samples with natural color compositions but often suffer from poor global structure and lower visual fidelity. To address these trade-offs, several empirical strategies have been proposed to balance diversity and quality of CFG \citep{intervalGuidance,sadat2024cads,sadat2025eliminating,wang2024analysis}. Despite this progress, a systematic understanding of how CFG improves image quality and prompt alignment remains limited. Existing works do not fully explain the internal mechanisms of CFG or explore how to improve generation quality at low guidance scales.

In this paper, our objective is to advance the understanding of how CFG works and improve image quality at low CFG scales, thereby avoiding the detrimental effects associated with high guidance scales. We begin by analyzing the CFG update rule in the frequency domain and show that CFG enhances image quality and prompt alignment through distinct frequency components. Specifically, we find that the low-frequency components primarily govern the global structure and condition alignment, while the high-frequency components mainly contribute to the visual quality and details, with minimal impact on the overall composition of the image. We also observe that excessive guidance in the low-frequency domain leads to reduced diversity and oversaturation, whereas high-frequency components usually benefit from higher guidance scales. Since standard CFG applies a uniform scale across all frequencies, this explains why low CFG scales degrade quality, and high CFG scales boost detail at the cost of diversity and oversaturation.

Building on this insight, we propose a new classifier-free guidance scheme in the frequency domain, termed \acrfull{method}, which disentangles the guidance scales applied to the low- and high-frequency components of the CFG signal. We argue that low-frequency components should be guided more conservatively to avoid  low-diversity and oversaturated generations, while high-frequency components can benefit from stronger guidance to enhance image quality. By assigning separate guidance strengths to these components, \gls{method} improves image quality while retaining the diversity typically associated with low CFG scales. \gls{method} is among the first approaches to systematically enhance the quality of low CFG scales, thereby avoiding the drawbacks of high CFG scales by design.

\gls{method} introduces practically no additional sampling cost and can be applied to any pretrained diffusion model without extra training or fine-tuning. Through extensive experiments, we show that \gls{method} consistently improves image quality and maintains diversity across a range of datasets, models, and metrics. Moreover, \gls{method} offers a deeper understanding of how CFG enhances image quality and prompt alignment by separately analyzing the roles of low- and high-frequency components in the CFG update rule. Accordingly, we consider \gls{method} as a superior plug-and-play alternative to standard classifier-free guidance for conditional diffusion models.
\section{Related work}
Score-based diffusion models \citep{DBLP:conf/nips/SongE19,score-sde,sohl2015deep,hoDenoisingDiffusionProbabilistic2020} learn the data distribution by reversing a forward diffusion process that progressively corrupts the data with Gaussian noise. These approaches have rapidly surpassed previous generative modeling techniques in terms of fidelity and diversity \citep{nichol2021improved,dhariwalDiffusionModelsBeat2021}, achieving state-of-the-art performance across a wide range of applications, including unconditional image synthesis \citep{dhariwalDiffusionModelsBeat2021,karras2022elucidating}, text-to-image generation \citep{dalle2,saharia2022photorealistic,balaji2022ediffi,rombachHighResolutionImageSynthesis2022,sdxl,yu2022scaling,esser2024scaling}, video synthesis \citep{blattmann2023align,stableVideoDiffusion,gupta2023photorealistic}, image-to-image translation \citep{saharia2022palette,liu20232i2sb}, motion synthesis \citep{tevet2022human,Tseng_2023_CVPR}, and audio synthesis \citep{WaveGrad,DiffWave,huang2023noise2music}.

Recent studies have introduced numerous enhancements to the original DDPM framework \citep{hoDenoisingDiffusionProbabilistic2020}, including improved network architectures \citep{hoogeboom2023simple,karras2023analyzing,peeblesScalableDiffusionModels2022,dhariwalDiffusionModelsBeat2021}, novel sampling strategies \citep{songDenoisingDiffusionImplicit2022,karras2022elucidating,plms,dpm_solver,salimansProgressiveDistillationFast2022}, and better training techniques \citep{nichol2021improved,karras2022elucidating,score-sde,salimansProgressiveDistillationFast2022,rombachHighResolutionImageSynthesis2022}. Despite these advancements, guidance methods—such as classifier-free guidance \citep{hoClassifierFreeDiffusionGuidance2022}—remain essential for enhancing sample quality and improving alignment between conditioning information and generated outputs \citep{glide}.

In modern diffusion models, high guidance scales are essential for improving image quality and enhancing alignment between the generated output and conditioning inputs. However, this comes at the cost of reduced diversity \citep{hoClassifierFreeDiffusionGuidance2022,sadat2024cads} and undesirable effects such as oversaturation \citep{sadat2025eliminating}. Our work addresses this issue by disentangling the benefits of high CFG from its detrimental effects. Leveraging frequency decomposition, we propose a novel approach that improves image quality at lower CFG scales, inherently avoiding the trade-offs associated with high guidance.

Several recent works have aimed to improve diffusion model training for unguided generation \citep{karras2023analyzing,yu2025representation,leng2025repa,SimpleDiffusionV2}. While unguided generation is theoretically expected to produce high-quality, diverse samples without the drawbacks of CFG, in practice, strong CFG guidance is still often necessary for high-quality generation even with such improved models. In contrast, our method enhances sample quality at low CFG scales without retraining, making it readily applicable to any pretrained model.

Frequency decomposition techniques, such as Laplacian pyramids \citep{burt1987laplacian} and wavelet transforms \citep{waveletIntro}, have recently been employed in generative models to boost both quality and efficiency \citep{denton2015deep,SwaGAN,atzmon2024edify,sadat2024litevae,agarwal2025cosmos,xiao2024frequency}. However, their use in improving the sampling characteristics of diffusion models remains underexplored. We demonstrate that applying the CFG update rule in the frequency domain significantly enhances generation quality while avoiding the common issues of high CFG scales.

\section{Background}

\paragraph{Diffusion models} Let $\vx \sim \pdata(\vx)$ denote a data sample, and let $t \in [0, 1]$ represent a continuous time variable. The forward diffusion process adds noise to the data as $\vz_t = \vx + \sigma(t)\mepsilon$, where $\sigma(t)$ is a time-dependent noise schedule. This schedule controls the degree of corruption, with $\sigma(0) = 0$ (no noise) and $\sigma(1) = \sigma_{\textnormal{max}}$ (maximum noise). As shown by \citet{karras2022elucidating}, this forward process corresponds to the following ODE:
\begin{equation}\label{eq:diffusion-ode}
    \dd\vz = - \dot{\sigma}(t) \sigma(t) \grad_{\vz_t} \log p_t(\vz_t) \dd t,
\end{equation}
where $p_t(\vz_t)$ is the distribution over noisy samples at time $t$, with $p_0 = \pdata$ and $p_1 = \normal{\zero}{\sigma_{\textnormal{max}}^2 \mI}$. Given access to the time-dependent score function $\grad_{\vz_t} \log p_t(\vz_t)$, one can sample from the original data distribution by integrating the ODE in reverse from $t=1$ to $t=0$. Since this score function is unknown, it is approximated by a neural denoiser $D_{\mtheta}(\vz_t, t)$, which is trained to recover the clean data $\vx$ from its noisy counterpart $\vz_t$. Conditional generation is enabled by augmenting the denoiser with an auxiliary input $\vy$, such as class labels or text prompts, yielding $D_{\mtheta}(\vz_t, t, \vy)$.

\paragraph{\textbf{Classifier-free guidance}} 
{Classifier-free guidance (CFG)} is an inference technique designed to enhance the quality of generated samples by interpolating between conditional and unconditional model predictions \citep{hoClassifierFreeDiffusionGuidance2022}. Let $\vy_{\textnormal{null}} = \varnothing$ denote a null condition representing the unconditional case. CFG modifies the denoiser output at each sampling step as follows:
\begin{equation}\label{eq:cfg}
    \predcfg =   \prednull + \wcfg \prn{\pred - \prednull},
\end{equation}
where $\wcfg = 1$ corresponds to the unguided case. The unconditional model $\prednull$ is trained by randomly replacing the condition $\vy$ with $\vy_{\textnormal{null}}$ during training. Alternatively, the unconditional score can be learned using a separate model as in \citet{karras2023analyzing}. Similar to the truncation trick used in GANs \citep{brockLargeScaleGAN2019}, CFG enhances image quality, but often at the cost of sample diversity \citep{pml2Book}.

\paragraph{Frequency decompositions} Multi-level frequency decompositions, such as Laplacian pyramids and wavelet transforms, are commonly used to separate an image into different frequency bands. These techniques decompose an image into coarse, low-frequency structures and fine, high-frequency details. The low-frequency components capture the global characteristics of the image, including object placement, overall geometry, and color distribution, while the high-frequency components represent localized information such as edges, textures, and fine structural details.

\section{Classifier-free guidance in the frequency domain}\label{sec:method}
We now describe how guidance in the frequency domain can enhance the characteristics of CFG. Our first goal is to understand how different frequency components in the CFG prediction $\predcfg$ influence the final generation. Let $\opfreq[\cdot]$ denote a linear and invertible frequency transformation, such as a Laplacian pyramid or wavelet transform, which decomposes each input $\vx$ into low- and high-frequency components, denoted by $\oplow[\vx]$ and $\ophigh[\vx]$, respectively.
For notational convenience, we define $\predcond \doteq \pred$, $\preduncond \doteq \prednull$, and $\predguided \doteq \predcfg$ to represent the conditional, unconditional, and CFG outputs, respectively. As a consequence of the assumed properties of $\opfreq$, the \gls{cfg} update rule can be expressed as
\begin{align} 
\predguided &= \opfreqinv[\opfreq[\predguided]]\\
 &= \opfreqinv \left[\opfreq\brk{\preduncond} + \wcfg \left( \opfreq\brk{\predcond} - \opfreq\brk{\preduncond} \right)\right]. 
\end{align} 
This implies that, in the frequency domain, the CFG update affects both the low- and high-frequency components of $\opfreq[\predguided] = \Set{\predguidedlow, \predguidedhigh}$ as follows: 
\begin{align} 
     \predguidedlow &= \oplow\brk{\preduncond} + \wcfg \left(\oplow\brk{\predcond} - \oplow\brk{\preduncond} \right) ,\\ 
\predguidedhigh &= \ophigh\brk{\preduncond} + \wcfg \left( \ophigh\brk{\predcond} - \ophigh\brk{\preduncond} \right). 
\end{align}
Thus, in standard CFG, both low- and high-frequency components are guided using the same scale $\wcfg$ throughout the sampling process. However, we argue that this approach is suboptimal, as the low- and high-frequency components exhibit different behaviors and influence distinct aspects of the generated image. We find that strong guidance on the low-frequency component $\predguidedlow$ leads to oversaturation and reduced diversity, whereas high guidance on the high-frequency component $\predguidedhigh$ primarily enhances image quality. On the other hand, low scales for $\wcfglow$ keeps diversity and realistic color composition while low values for $\wcfghigh$ degrade the visual details of the image and result in reduced sample quality. Motivated by this, we propose a generalized CFG scheme, called \acrfull{method}, that employs separate guidance scales—$\wcfglow$ for low-frequency and $\wcfghigh$ for high-frequency components.

\CFGDecompFig
\Cref{fig:method} illustrates the effect of low- and high-frequency components on generated images. For CFG ($\wcfglow = \wcfghigh$), low guidance scales lead to poor global structure and visual degradation, while high guidance reduces diversity. We observe that strong guidance on the low-frequency signal {(i.e., a large $\wcfglow$)} primarily causes diversity issues, whereas increasing $\wcfghigh$ enhances quality without adverse effects on diversity. These findings highlight the limitations of using a single scale $\wcfg$ in standard CFG: low $\wcfg$ produces blurry or incoherent outputs, while high $\wcfg$ reduces diversity and causes oversaturation. Our results therefore advocate for asymmetric guidance, where we set $\wcfglow < \wcfghigh$.

\paragraph{Implementation details} We use Laplacian pyramids \citep{burt1987laplacian} as the frequency transform $\opfreq$ in our experiments. The complete algorithm for applying \gls{method} as well as the pseudocode of our method is provided in the appendix. Notably, \gls{method} requires only minor modifications to the standard CFG sampling procedure, introduces no significant computational overhead, and is readily compatible with all pretrained diffusion models.
\section{Experiments}
\figMain
\paragraph{Setup} We primarily conduct experiments on text-to-image generation using Stable Diffusion models \citep{rombachHighResolutionImageSynthesis2022,sdxl,esser2024scaling}, and class-conditional ImageNet \citep{imagenet} generation using EDM2 \citep{karras2023analyzing} and DiT-XL/2 \citep{peeblesScalableDiffusionModels2022}. In all cases, we adopt the default diffusion samplers provided by each model (e.g., Euler scheduler for Stable Diffusion XL), and utilize the official pretrained checkpoints and codebases to maintain consistency with the original implementations. More details on the experimental setup and the hyperparameters used for each experiment are given in the appendix.

\paragraph{Evaluation metrics} For class-conditional models, we adopt Fréchet Inception Distance (FID) \citep{fid} as the main metric to assess both image quality and diversity, given its strong correlation with human perception. To account for FID's sensitivity to implementation details, we evaluate all models under a consistent setup. Additionally, we report precision \citep{improvedPR} as a supplementary quality metric and recall \citep{improvedPR} to capture diversity. For text-to-image tasks, we further use ImageReward \citep{xu2023imagereward}, HPSv2 \citep{wu2023human}, PickScore \citep{kirstain2023pickapic}, and CLIP Score \citep{Hessel2021CLIPScore} to evaluate both image quality and prompt alignment.

\subsection{Main results}

\paragraph{Qualitative results}
\Cref{fig:edm-main,fig:sd-main} qualitatively compare the generations of \gls{method} and CFG for the EDM2 and Stable Diffusion XL models at low CFG scales. \gls{method} enhances generation quality while preserving the overall structure and color palette of the CFG output. Thus, \gls{method} improves the quality at low CFG scales while avoiding the issues caused by high CFG values.

\tabResultMain
\tabResultSD
\paragraph{Quantitative results}
\Cref{tab:main-results} provides the metrics for \gls{method} and \gls{cfg} across several models at guidance values typically used in practice. We observe that \gls{method} consistently improves FID and recall while largely maintaining the precision of the CFG outputs. We therefore conclude that \gls{method} enhances quality while preserving diversity, leading to significantly better FID. Additionally, we evaluate generation quality at lower CFG values in \Cref{tab:sd-results} using metrics specifically designed to reflect human preferences for text-to-image models. As \gls{method} boosts quality without introducing the detrimental effects of high CFG scales, it significantly outperforms CFG across all quality metrics for the Stable Diffusion models.

\subsection{Effect of different frequency components of CFG on the generated distribution}
To directly measure the effect of different frequency components on CFG outputs, we compared CFG with two sampling variants that use only $\predguidedlow$ or $\predguidedhigh$. This was achieved by setting $\wcfghigh = 1$ or $\wcfglow = 1$ in \gls{method}. As shown in \Cref{fig:scales}, increasing $\wcfglow$ is the main cause of the reduced diversity in CFG, as evidenced by the higher FID and lower recall values. In contrast, increasing $\wcfghigh$ improves precision while maintaining recall, leading to significantly better FID across most guidance scales. These results suggest that excessive low-frequency guidance is the main contributor to the adverse effects of high CFG, whereas increasing high-frequency guidance generally enhances generation quality. Additionally, we observed that the low-frequency component is the primary cause of oversaturation, explaining why high guidance scales result in color artifacts. Therefore, our method adopts low values for $\wcfglow$ and high values for $\wcfghigh$.
\EDMCFGHighScalesPlot

\subsection{Frequency analysis of prompt alignment in CFG}
\AligmentPlot
We next demonstrate how different frequency components of the CFG signal influence the alignment between generated images and the input condition. \Cref{fig:alignment} shows that although both low- and high-frequency components improve alignment as the guidance scale increases, the low-frequency component is the primary driver of this effect. Therefore, \gls{method} can achieve comparable or superior prompt alignment to CFG by appropriately combining low- and high-frequency components, as shown in \Cref{tab:sd-results}.

\subsection{{Additional experiments}}
\paragraph{Relation to variable guidance scale} Several works have explored the use of time-dependent guidance scales to balance diversity and quality in classifier-free guidance \citep{sadat2024cads,intervalGuidance,wang2024analysis}. For example, guidance interval (GI) \citep{intervalGuidance} applies the guidance scale only during a limited range of sampling steps identified via grid search. We argue that the improvement in diversity offered by these methods is closely related to the frequency decomposition of the guidance signal. To support this, we provide the norms of the low- and high-frequency components of the guidance signal across sampling steps in \Cref{fig:norm}. Note that the norm of the high-frequency component increases over sampling, while the norm of the low-frequency component decreases. We observe that GI starts applying guidance when the norms of the two components become roughly equal. This suggests that applying guidance in the mid-stages, as GI does, implicitly increases the effective guidance on the high-frequency component relative to the low-frequency component. This analysis also provides a principled way to select an interval for applying GI, avoiding costly grid searches. Additionally, GI may still lead to quality degradation due to the absence of guidance at the beginning of sampling. As shown in \Cref{tab:gi-comp}, \gls{method} outperforms GI in terms of image quality for Stable Diffusion XL by maintaining high-frequency guidance during the early sampling steps.
\EDMNormsPlot

\paragraph{Frequency analysis of Autoguidance} Autoguidance \citep{karras2024guiding} proposed a modified version of CFG that replaces the unconditional prediction with a degraded version of the main diffusion model. \Cref{fig:norm} presents a frequency analysis of the update provided by Autoguidance, showing that both low- and high-frequency components remain strong throughout the sampling process, unlike CFG where low-frequency response dominates. This likely explains why the authors found Autoguidance effective at all steps and why it provides a better guidance direction compared to CFG.
\tabGIComp

\tabSDXLLightning
\paragraph{Compatibility with distilled models}
CFG is often detrimental to distilled models that use a small number of sampling steps. In contrast, we show that \gls{method} can be effectively applied to distilled models, such as SDXL-Lightning \citep{sdlightning}, without quality degradation. \Cref{tab:sdxl-distilled} provides a quantitative comparison between \gls{method} and CFG. Compared to both baselines, \gls{method} achieves higher quality metrics with good prompt alignment.

\figSDthreeMain
\paragraph{Improving text rendering in diffusion models}
We next demonstrate that \gls{method} can enhance the quality of generated text in Stable Diffusion 3 \citep{esser2024scaling}. Generating high-quality text requires substantial details, which poses a challenge for standard CFG, since high guidance scales often result in unrealistic samples. In contrast, \Cref{fig:sd3-text} shows that \gls{method} achieves realistic generation and correct spelling of text by separately controlling $\wcfglow$ and $\wcfghigh$.

\subsection{Ablation studies}
\paragraph{Effect of the frequency decomposition operator} We next test the performance of \gls{method} for two choices of $\opfreq$ in \Cref{tab:ablation-dwt}. We note that \gls{method} is not sensitive to this design choice as long as the operation provides an informative low- and high-frequency component. We chose the Laplacian pyramid, as it slightly outperformed DWT in our experiments.

\paragraph{Using multi-level frequency decomposition} We also experiment with a multi-level Laplacian pyramid for frequency decomposition, applying different guidance scales to each frequency level. \Cref{tab:multilevel} shows both multi-level and single-level approaches outperform CFG. For simplicity, our main experiments used a single-level pyramid, though multi-level decomposition can help control separate high-frequency bands and remains viable for applications needing such control.
\tabAblationMain
\section{Conclusion}\label{sec:conclusion}

In this work, we have taken a principled look at classifier-free guidance in the frequency domain and have shown that its beneficial effects on structural fidelity and fine details stem from strong guidance applied to the high-frequency components of the CFG signal, while its detrimental impact on diversity and oversaturation arises from excessive guidance on the low-frequency components. Building on this insight, we proposed \acrfull{method} to disentangle guidance strength across frequency bands, applying conservative scaling to low frequencies while exploiting stronger scaling at high frequencies. This approach preserves the diversity and color composition of low guidance scales while enhancing details akin to high guidance scales. As a result, \acrshort{method} improves the quality of low CFG values while avoiding the adverse effects of high CFG scales by design. Importantly, \gls{method} introduces practically no additional training or sampling cost and can be seamlessly integrated as a \emph{plug-and-play} enhancement to any pretrained diffusion model using CFG. As with CFG itself, challenges remain in accelerating sampling and improving generation quality in extreme out-of-distribution domains, which we identify as promising directions for future research.


{
\clearpage
\bibliographystyle{plainnat}
\bibliography{ref}
}

\appendix
\clearpage
\section{Broader impact statement}\label{sec:impact-statemetn}
Our approach has the potential to enhance the realism and quality of outputs generated by diffusion models without the need for costly retraining. As such, it offers practical benefits for visual content creation. However, with the continued advancement of generative modeling, the generation and dissemination of fabricated or inaccurate data become increasingly accessible. While advancements in AI-generated content have the potential to enhance productivity and creativity, it remains essential to critically assess the accompanying risks and ethical considerations. For a comprehensive discussion on ethics and creativity within computer vision, we direct readers to \cite{rostamzadeh2021ethics}.

\section{Background on frequency decompositions}
\paragraph{Laplacian Pyramids}
The Laplacian pyramid \citep{burt1987laplacian} is a multi-scale representation of an image based on successive band-pass filtering. Starting from an input image \(\vx\), a Gaussian pyramid \(\{G_0, G_1, \ldots, G_N\}\) is constructed, where \(G_0 = \vx\), and each level is a progressively downsampled (and low-pass filtered) version of the previous:
\begin{equation}
    G_{i+1} = \texttt{Downsample}(\texttt{GaussianBlur}(G_i)).
\end{equation}
The Laplacian pyramid \(\{L_0, L_1, \ldots, L_{N-1}\}\) is then formed by subtracting the upsampled version of each Gaussian level from its corresponding higher-resolution level:
\begin{equation}
    L_i = G_i - \texttt{Upsample}(G_{i+1}).
\end{equation}
The top level of the pyramid is typically the final low-resolution image \(G_N\). The decomposition can be inverted to reconstruct the original signal by sequentially upsampling and summing the Laplacian levels:
\begin{equation}
    G_i = L_i + \texttt{Upsample}(G_{i+1}).
\end{equation}
This approach enables localized manipulation of image details at different spatial scales and is commonly used in image compression, enhancement, and blending \citep{szeliski2022computer}.

\paragraph{Wavelet transforms}
Wavelet transforms \citep{waveletIntro} are widely used in signal processing to extract spatial-frequency characteristics from input data. They rely on a pair of filters: a low-pass filter $L$ and a high-pass filter $H$. In the case of 2D signals, four filters are derived as $L\trp{L}$, $L\trp{H}$, $H\trp{L}$, and $H\trp{H}$. When applied to an image $\vx$, the 2D wavelet transform decomposes it into a low-frequency sub-band $\vx_{L}$ and three high-frequency sub-bands $\{\vx_{H}, \vx_{V}, \vx_{D}\}$, which capture horizontal, vertical, and diagonal details, respectively. For an input image of size $H \times W$, each sub-band has a spatial size of $H/2 \times W/2$. Multi-resolution analysis can be performed by recursively applying the wavelet transform to $\vx_{L}$. The transform is also invertible, allowing the reconstruction of the original image $\vx$ from the set $\{\vx_{L}, \vx_{H}, \vx_{V}, \vx_{D}\}$ using the inverse wavelet transform. Moreover, fast wavelet transform (FWT) \citep{mallat1989theory} makes it possible to compute wavelet sub-bands with linear complexity relative to the number of pixels in $\vx$.

\section{Additional experiments}
We present additional experiments in this section to further demonstrate the effectiveness of \gls{method} across various scenarios. 

\figEffectFreq
\subsection{Effect of low- and high-frequency components on generations}
To further demonstrate that low-frequency components govern global structure and high-frequency components contribute to details, we conducted an experiment in which we explicitly set either the high- or low-frequency portion of the CFG signal to zero. The results in \Cref{fig:freq-effect} show that when high frequencies are removed, the final generation retains the overall structure of the base image. Conversely, when low frequencies are removed, the generated image roughly shows the high-frequency details of the base image such as edges. These findings suggest that low-frequency components in the CFG signal control global structure, while high-frequency components determine more localized details.

\subsection{Relation to APG} \Gls{apg} \citep{sadat2025eliminating} is a method designed to reduce oversaturation and artifacts caused by high guidance scales. \Gls{apg} complements our methods, as its update rule can be integrated with the frequency decomposition in \gls{method} to get better guidance directions. We demonstrate this combined approach in \Cref{fig:apg} by incorporating the orthogonal projection from \gls{apg} into \gls{method}. The results indicate that this projection continues to effectively produce more realistic color compositions and fewer artifacts in generations.
\figAPG

\subsection{Combining \gls{method} with CADS}
CADS \citep{sadat2024cads} is an inference method designed to increase the diversity of diffusion models at high guidance scales by perturbing the conditional embedding with Gaussian noise. In this section, we show that CADS is compatible with our method, and that their combination outperforms either approach used in isolation. \Cref{table:cads} supports this finding using the DiT-XL/2 model as a benchmark. We therefore conclude that the benefits of \gls{method} are complementary to those of CADS.
\tabCADS

\subsection{Using different diffusion samplers} 
We also show that the effectiveness of \gls{method} is not limited to a specific diffusion sampler. \Cref{table:samplers} compares the metrics of \gls{method} and CFG across several popular diffusion samplers for DiT-XL/2, demonstrating that \gls{method} consistently outperforms CFG across all setups.
\tabSamplers

\subsection{Changing the number of sampling steps} 
We compared the performance of \gls{method} and CFG across different numbers of sampling steps. \Cref{fig:nfe} shows that \gls{method} maintains a consistent advantage over CFG across various sampling steps, leading to improved FID and recall while preserving a similar level of precision. Therefore, we conclude that the observed improvements in \gls{method} hold across different sampling budgets.
\NFEPlot

\section{Implementation details}
The sampling algorithm of \gls{method} is provided in \Cref{algo:sampling}, and the corresponding PyTorch implementation is given in \Cref{alg:code}. Compared to CFG, \gls{method} only adds a few extra lines of code and does not incur any noticeable computational overhead. As stated in the code, we convert the model’s output to the denoised estimate $\pred$ (also known as the $x_0$ prediction), apply the guidance step, and then convert it back to the original output format at each sampling step.

For evaluation, we mainly rely on the ADM evaluation suite \citep{dhariwalDiffusionModelsBeat2021} to calculate FID, precision, and recall. For class-conditional ImageNet models, FID is computed using $\num{10000}$ generated images along with the complete training dataset. In the case of text-to-image models, FID is measured using the validation split of MS COCO 2017 \citep{lin2014microsoft}. To evaluate text-to-image quality metrics such as ImageReward \citep{xu2023imagereward}, we follow the official implementations and use the authors' provided test datasets. For PickScore \citep{kirstain2023pickapic}, we calculate the win probability and report a win if one image outperforms the other by a margin greater than 0.1; otherwise, the result is reported as a tie. Details of the hyperparameters used in our main experiments are listed in \Cref{tab:imp-detail}.
\tabParameters

\algMethod
\begin{algorithm}[t!]
    \caption{PyTorch implementation of \acrshort{method}.}
        \label{alg:code}
        \centering
        \begin{minted}
    [
    % frame=lines,
    framesep=2mm,
    baselinestretch=1.2,
    bgcolor=LG,
    fontsize=\footnotesize,
    % linenos
    ]
    {python}
    
import torch
from kornia.geometry.transform import build_laplacian_pyramid


def project(
    v0: torch.Tensor, # [B, C, H, W] 
    v1: torch.Tensor, # [B, C, H, W]
):
    dtype = v0.dtype
    v0, v1 = v0.double(), v1.double()
    v1 = torch.nn.functional.normalize(v1, dim=[-1, -2, -3])
    v0_parallel = (v0 * v1).sum(dim=[-1, -2, -3], keepdim=True) * v1
    v0_orthogonal = v0 - v0_parallel
    return v0_parallel.to(dtype), v0_orthogonal.to(dtype)


def build_image_from_pyramid(pyramid):
    img = pyramid[-1]
    for i in range(len(pyramid) - 2, -1, -1):
        img = kornia.geometry.pyrup(img) + pyramid[i]
    return img


# We assume all model predictions are converted to "x_0" prediction.
def laplacian_guidance(
    pred_cond: torch.Tensor,   # [B, C, H, W]
    pred_uncond: torch.Tensor, # [B, C, H, W]
    guidance_scale=[1.0, 1.0], # Guidance scales from high- to low-frequency
    parallel_weights=None,     # Optional weights for projection
):
    levels = len(guidance_scale)
    if parallel_weights = None:
        parallel_weights = [1.0] * levels

    pred_cond_pyramid = build_laplacian_pyramid(pred_cond, levels)
    pred_uncond_pyramid = build_laplacian_pyramid(pred_uncond, levels)

    pred_guided_pyramid = []
    parameters = zip(
        pred_cond_pyramid, pred_uncond_pyramid, guidance_scale, parallel_weights
        )
    for idx, (p_cond, p_uncond, scale, par_weight) in enumerate(parameters):
        diff = p_cond - p_uncond
        diff_parallel, diff_orthogonal = project(diff, p_cond)
        diff = par_weight * diff_parallel + diff_orthogonal
        p_guided = p_cond + (scale - 1) * diff
        pred_guided_pyramid.append(p_guided)
    pred_guided = build_image_from_pyramid(pred_guided_pyramid)
    return pred_guided.to(pred_cond.dtype)
    \end{minted}
\end{algorithm}

\section{More visual examples}
This section provides additional samples comparing the performance of \gls{method} with CFG. \Cref{fig:imagenet-appendix} presents further results for class-conditional generation using DiT-XL/2 and EDM2. For both models, we observe that \gls{method} preserves the structure of the base image while significantly enhancing the details. Additional results for text-to-image models are shown in \Cref{fig:sdxl-appendix,fig:sdthree-appendix}.

\figDiTAppendix
\figSDXLAppendix
\figSDThree

\end{document}